\documentclass{article}

    \PassOptionsToPackage{numbers, compress, sort}{natbib}
 \usepackage[main, final]{neurips_2026}

\usepackage[utf8]{inputenc} 
\usepackage[T1]{fontenc}    
\usepackage{hyperref}       
\usepackage{url}            
\usepackage{booktabs}       
\usepackage{amsfonts}       
\usepackage{amssymb}        
\usepackage{nicefrac}       
\usepackage{microtype}      
\usepackage{xcolor}         
\usepackage{colortbl}
\usepackage{array}

\newif\ifshowcomments
\showcommentsfalse   

\ifshowcomments
  \newcommand{\rr}[1]{\textcolor{blue}{(\textbf{rr:} #1)}}
  \newcommand{\ck}[1]{\textcolor{orange}{(\textbf{CK:} #1)}}
  \newcommand{\vigk}[1]{\textcolor{purple}{(\textbf{VK:} #1)}}

  \newcommand{\mf}[1]{\textcolor{red}{(\textbf{Matthias:} #1)}}
  \newcommand{\jure}[1]{\textcolor{red}{(\textbf{Jure:} #1)}}

  \newcommand{\cc}[1]{\textcolor{orange}{(\textbf{cc:} #1)}}
  
  \providecommand{\claude}[1]{{\color{orange}{/* claude: #1 */}}}
\else
  \newcommand{\rr}[1]{}
  \newcommand{\ck}[1]{}
  \newcommand{\vigk}[1]{}

  \newcommand{\mf}[1]{}
  \newcommand{\jure}[1]{}

  \newcommand{\cc}[1]{}
  
  \providecommand{\claude}[1]{}
\fi

\definecolor{rtmaroon}{HTML}{8C1515}  
\definecolor{paloaltogreen}{HTML}{175E54}  
\definecolor{coolgrey}{HTML}{53565A}  
\newcommand{\theJoin}{\textcolor{paloaltogreen}{\textbf{\textsc{the Join}}}}
\newcommand{\TheJoin}{\textcolor{paloaltogreen}{\textbf{\textsc{The Join}}}}
\newcommand{\rtj}{\mbox{\textcolor{rtmaroon}{\textbf{RT-J}}}}
\newcommand{\rtp}{RT-PluRel}

\usepackage{amsmath}
\usepackage{cleveref}
\crefname{figure}{Figure}{Figures}
\Crefname{figure}{Figure}{Figures}
\crefname{section}{Section}{Sections}
\Crefname{section}{Section}{Sections}
\crefname{subsection}{Section}{Sections}
\Crefname{subsection}{Section}{Sections}
\crefname{appendix}{Appendix}{Appendices}
\Crefname{appendix}{Appendix}{Appendices}
\crefname{subappendix}{Appendix}{Appendices}
\Crefname{subappendix}{Appendix}{Appendices}
\crefname{subsubappendix}{Appendix}{Appendices}
\Crefname{subsubappendix}{Appendix}{Appendices}
\crefname{table}{Table}{Tables}
\Crefname{table}{Table}{Tables}
\usepackage{algorithm}
\usepackage{algpseudocode}
\usepackage{graphicx}
\usepackage{subcaption}

\definecolor{subsampledgray}{gray}{0.88}

\definecolor{provisionalpink}{rgb}{1.0,0.85,0.90}

\usepackage{wrapfig}
\usepackage{multirow}
\usepackage{listings}
\usepackage{longtable, array}
\usepackage{tikz}

\title{\rtj: Large-Scale Pretraining of Relational Transformers for Context-Efficient Predictions}

\author{%
  \bfseries Rishabh Ranjan$^{\bowtie}$ \quad Vignesh Kothapalli$^{\bowtie}$ \quad Harshvardhan Agarwal$^{\bowtie}$ \\ \bfseries Charilaos Kanatsoulis$^{\bowtie}$ \quad Roshan Upendra$^{\pi}$ \quad Tom Palczewski$^{\pi}$ \\
  \bfseries Carlos Guestrin$^{\bowtie}$ \quad Jure Leskovec$^{\bowtie}$ \\[5pt]
  {\normalfont $^{\bowtie}$Stanford University \quad $^{\pi}$SAP 
  }
   \\[10pt]
  {\normalfont\small
  \href{https://star-project.stanford.edu/rt-j}{%
    \raisebox{-0.32\height}{\includegraphics[height=1.6em]{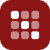}}\hspace{6pt}%
    \texttt{https://star-project.stanford.edu/rt-j}}}
}

\hypersetup{pdftitle={RT-J: Large-Scale Pretraining of Relational Transformers for Context-Efficient Predictions},pdfauthor={Rishabh Ranjan, Vignesh Kothapalli, Harshvardhan Agarwal, Charilaos Kanatsoulis, Roshan Upendra, Tom Palczewski, Carlos Guestrin, Jure Leskovec}}

\begin{document}

\maketitle
{\let\thefootnote\relax\footnotetext{
Correspondence to: {\texttt{<ranjanr,vigneshk,guestrin,jure>@cs.stanford.edu}}
}}

\begin{abstract}
Foundation models for
tables and relational databases are increasingly used for predictive tasks across various scientific and industrial domains.
However, adapting existing models to unseen datasets requires tens of thousands of examples.
Here we introduce \rtj{}, an $86$M-parameter relational foundation model (RFM) that makes accurate predictions even with hundreds of in-context examples or fewer.
To achieve this,
we first assemble \theJoin{},
a large collection of $639$ web-crawled databases
with $6{,}023$ LLM-generated forecasting tasks.
We pretrain with long contexts of up to $8{,}192$ database cells,
using in-context examples from a novel lightweight random-walk retriever.
Finally, we leverage context tuning and
ensembling to scale up test-time compute.
On in-context regression tasks from RelBench,
\rtj{} requires $10\times$ fewer examples on average
to match strong feature engineering pipelines with
LLM-agent or RDBLearn featurizers and
TabICLv2 or LightGBM predictors.
Overall, our work pushes the frontier
for context-efficient predictions on relational data and
highlights the importance
of real-world semantics, context engineering and large-scale pretraining for RFMs.
\end{abstract}

\begin{figure}[t]
  \centering
  \includegraphics[width=\linewidth]{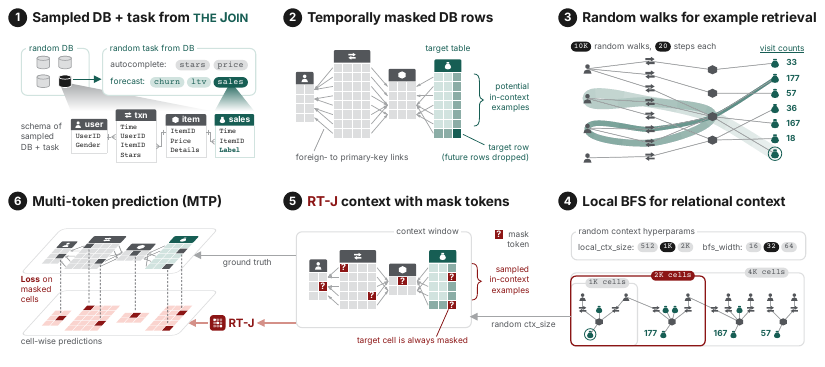}
  \caption{\rtj{} pretraining recipe.
\textbf{(1)} We sample a database and prediction task from \theJoin{}. The task specifies the \emph{target table} and the \emph{target column}.
\textbf{(2)} We sample a \emph{target row} from the target table. Rows after the target row's timestamp are dropped; kept target-table rows form the candidate \emph{in-context examples}.
\textbf{(3)} From the target row, many parallel random walks rank candidate examples by visit count, on the fly.
\textbf{(4)} We process candidate examples in descending visit-count order, using each as the root of a local bounded-width breadth-first-search (BFS) over primary--foreign key links; the context size and BFS hyperparameters are sampled randomly.
\textbf{(5)} The collected rows form the \emph{context window} for the relational transformer (RT), which treats cells as \emph{tokens}. The target table rows in the context enable \emph{in-context learning}. The \emph{target cell} is always masked; non-target cells may be masked optionally.
\textbf{(6)} \rtj{} predicts all masked cells and is trained against their ground-truth values.
}
  \label{fig:recipe}
\end{figure}

\section{Introduction}
Foundation models have  transformed how machine learning systems are adapted to new tasks~\cite{bommasani2022foundation,zhou2024comprehensive}. In language and vision, pretrained models can often solve new tasks using only a small number of examples in context~\cite{gpt3,flamingo}. 
Yet this capability remains largely out of reach for relational data, even though relational databases and data warehouses store much of the world’s structured enterprise, scientific, and operational data~\cite{vogel2022towards,rdl}. Today, strong relational predictors typically require task-specific training~\cite{relgnn,relgt,dbformer}, or substantial labeled data~\cite{kumorfm,kumorfm2,chen2016xgboost,lam2021automated,shwartz2022tabular,tabpfn3-rel,tabpfn3.5-rel}.
This limits predictive modeling precisely in the settings where it is most valuable, including cases where labeled examples are scarce,
and users need rapid ad hoc predictions for decision-making.


Tabular foundation models (TFMs) like TabPFN~\cite{tabpfn,tabpfnv2} and TabICL~\cite{tabicl,tabiclv2} 
provide in-context learning for single-table prediction, and systems such as RDBLearn~\cite{rdblearn} extend them to relational databases by aggregating relational neighborhoods into engineered features.
Such an approach can be effective with many labeled examples, but it flattens the database and removes the explicit relational structure needed to effectively reason over entities, relationships, and cross-table dependencies.
Natively relational architectures like RelGT~\cite{relgt} and RelGNN~\cite{relgnn} operate directly on multi-table databases by representing rows as nodes and primary--foreign-key links as edges, but require training for each schema and task. 
More recent schema-agnostic relational foundation models (RFMs), such as Griffin~\cite{griffin}, KumoRFM~\cite{kumorfm,kumorfm2}, and relational transformers (RTs)~\cite{rt}, pretrain across many databases and adapt to  
unseen ones
at inference time. However, these models have not yet been tailored for strong prediction with only a few labeled examples.
See \Cref{sec:related_work} for related work.

This inability is not primarily an architectural limitation, but a mismatch between existing RFM training recipes and inference with limited in-context examples.
RT~\cite{rt} provides a natural starting point because it operates directly over database cells, schema metadata, and relational links. However, the original RT recipe was not designed to turn the architecture into a strong in-context learner. There are three main gaps. First,  publicly available real world relational pretraining data remains  scarce~\citep{gu2026relbench,motl2015ctu}.
Second, existing pretraining recipes target zero-shot or fine-tuning regimes, which precludes in-context learning. Third, standard context construction either ignores relational structure or expands local neighborhoods naively, without preferentially retrieving rows relevant to the target.

\begin{wrapfigure}{r}{0.4\linewidth}
\vspace{-0.65em}
  \centering
  \includegraphics[width=\linewidth]{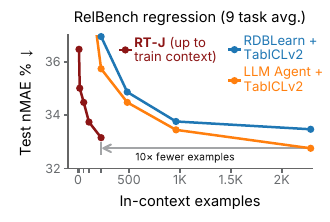}
  \caption{
  In-context regression.
  Examples are matched.
  \rtj{} at train context size of $8192$ cells fits
  around $225$ examples on average.
  At all example counts, \rtj{} is $>2$ nMAE points better
  than the best baseline;
  at all performance levels, \rtj{} needs $\gtrsim 10\times$ fewer examples.
  }
  \label{fig:results}
  \vspace{-2em}
\end{wrapfigure}
In this paper, we show that a pretrained RT can become a strong context-efficient predictor when the  recipe for training, data, context retrieval and test-time compute is designed for this regime. Our goal is not to introduce a new backbone architecture, but to identify the ingredients that turn an RT model into an effective in-context learner for relational databases. First, we develop a pretraining recipe (\Cref{fig:recipe}) built on stochastic context sizes and hyperparameters, which 
exposes the model 
to long and diverse contexts.
Second, to train with high-quality in-context examples, we propose a lightweight random-walk retriever (\Cref{fig:recipe}) that fills the context window with rows selected for both relational proximity to the target and likelihood of carrying useful predictive signal. Third, we leverage context tuning and context ensembling to scale test-time compute. With this recipe, we first obtain \rtp{}, an $86$M-parameter RT model pretrained on synthetic PluRel data~\cite{kothapalli2026plurel}.
With task-specific fine-tuning, \rtp{} ranked first among all system and model submissions on the RelArena-$\alpha$~\citep{hayler2026relarena} leaderboard (\Cref{app:leaderboards}). Finally, to improve generalization to real data, we assemble \theJoin{}, the largest open pretraining corpus of real-world relational data to date, comprising $6$K LLM-generated forecasting tasks and $11$K autocomplete tasks
across $639$ diverse web-crawled databases.
We refer to the model obtained by continued pretraining (CPT) of \rtp{} on \theJoin{} with our recipe as \rtj{}.
Compared to the original RT pretraining~\cite{rt}, our pretraining used $11\times$ more total tokens with up to $8\times$ larger contexts and a $4\times$ larger model, amounting to roughly $70\times$ more pretraining FLOPs.

Together, these ingredients make \rtj{} far more context-efficient than strong feature-engineering pipelines (\Cref{fig:results}). We compare \rtj{} against pipelines that compose an LLM agent or RDBLearn featurizer with a TabICLv2 or LightGBM predictor, each given the same examples as in \rtj{}'s context.
Other RFMs such as KumoRFM~\citep{kumorfm,kumorfm2}, Griffin~\citep{griffin} and RDB-PFN~\citep{wang2026rdbpfn} do not support this regime,
hence could not be compared against (\Cref{app:other-rfms}).
On RelBench regression, 
\rtj{} fits $225$ examples on average
at its max training context size of $8192$ database cells,
and achieves a normalized mean absolute error (nMAE)
which
the strongest baseline matches only with roughly $10\times$ as many examples. Moreover, with only $15$ examples, \rtj{} already beats the best error any pipeline reaches with $225$ examples.
At matched example counts, \rtj{} leads at every context size by $2.3$--$5.0$ nMAE points.
On classification, \rtj{} leads by $3.1$ AUROC points when examples are scarcest, and the strongest pipeline catches up only at the largest context size.
We attribute this efficiency to \rtj{} processing the raw relational context natively, which feature-engineering pipelines flatten into a feature table via aggregation techniques.

Our results demonstrate that context-efficient relational prediction can be attained
with a pretraining recipe matched to this regime and scaling up along the axes of data, context and test-time compute.
Overall, this is a step toward RFMs that work for databases the way LLMs work for text: pretrained once, reused across domains, and adapted to new tasks with only a  few examples. 

\section{Background: Relational Transformers for Predictive Tasks on Databases}
\label{sec:background}


\paragraph{Relational databases and relational structure.} Let $\mathcal{D}$ denote a relational database with schema graph
$\mathcal{G} = (\mathcal{T}, \mathcal{E})$, where each node
$T \in \mathcal{T}$ is a table and each directed edge
$(\widetilde{T}, T) \in \mathcal{E}$ indicates that table $T$ contains a
foreign key column referencing the rows of the parent table $\widetilde{T}$. Each table $T$
contains rows indexed by primary key(s) and columns consisting of numerical, categorical, textual, or temporal values, and optional foreign keys. Thus, the relational structure of $\mathcal{D}$ is characterized by the schema graph along with row-level bipartite connectivity via primary--foreign key links~\citep{rdl,relbench,kothapalli2026plurel}.

\paragraph{Prediction tasks on relational data.} 
A prediction task specifies a target table $T^\star$, a target column
$c^\star$, and a set of prediction rows $\mathcal{I}$ in $T^\star$. For each
row $r \in \mathcal{I}$, the target value is 
$y_r = x_{r,c^\star}$.
The goal of a foundation model $f_\theta$ is to predict $y_r$ from the observed relational context around row $r$. 
For boolean or categorical targets, $f_\theta$ outputs a class distribution.
For numerical targets, it outputs a real-valued prediction.
This formulation covers both (1) \textit{autocompletion} and (2) \textit{forecasting} tasks. In an
autocompletion task, $T^\star$ is an existing database table and $c^\star$ is
one of its columns (such as imputing the price of a transaction). In a forecasting task,
$T^\star$ may be a curated prediction table whose rows define future
prediction queries (such as forecasting revenue for an entity in the next
quarter). When either task is temporal, the context is restricted to rows available before the prediction time.

\paragraph{In-context learning for prediction tasks.}
In a classical in-context learning approach, $f_\theta$ predicts the target column $c^\star$
for a query row $r$ by conditioning on a support set $\mathcal{S}_r$
of labeled rows from the target table $T^\star$.
Each support row
$i \in \mathcal{S}_r$ has an observed target value
$y_i = x_{i,c^\star}$. A local relational context $\mathcal{C}_i$ is then constructed
around each support row and a context $\mathcal{C}_r$ around the
query row $r$ (with its target value masked). 
The model $f_\theta$ predicts $y_r$ using
the support and query context as:
\begin{align}
\label{eq:icl_pred}
    \hat{y}_r
    =
    f_\theta\!\left(
        \{(\mathcal{C}_i, y_i) : i \in \mathcal{S}_r\}, \mathcal{C}_r
    \right).
\end{align}
Here, in-context examples are sampled explicitly as labeled rows, and each
example is represented by its own relational context. However, support rows
often retrieve overlapping cells through primary--foreign key links. For
example, labeled rows for people who purchased the same shoe after seeing the
same advertisement may include the same cells from the shoe table in their
contexts. This duplication consumes the context budget and limits the number of examples in context.


\paragraph{Context window of relational transformers.} RT~\citep{rt} instantiates the context preparation idea differently.
Rather than separately sampling support rows and constructing an independent
context around each one, RT samples a single query-centered context
$\mathcal{C}_r$ around the query row $r$ (see stage (4) in \Cref{fig:recipe}). This context is obtained by expanding
from $r$ along primary--foreign key links while keeping $y_r$ masked. During
this expansion, the context may naturally include other rows from the target
table whose target values are observed. Thus, the support examples are not
sampled separately. They are induced implicitly by the relational context
around the query row. Therefore, the prediction of an RT model is given by $\hat{y}_r = f_\theta(\mathcal{C}_r)$.
The benefit is that overlapping relational neighborhoods are represented only
once inside a single query-centered context, rather than being duplicated
across independently sampled support examples.

\paragraph{Masked token prediction with relational transformers.}  RT treats each cell in $\mathcal{C}_r$
as a token and
predicts the masked target cell with a masked token prediction
objective. Thus, heterogeneous prediction tasks are reduced to a common
`cell prediction' problem over any database in a schema-agnostic manner. RT computes the input token embedding by combining the cell value, datatype-specific information, and frozen text embeddings of the corresponding table and column names. During the forward pass, each transformer block of RT applies three attention layers~\cite{kothapalli2026plurel} in sequence: (1) \emph{column attention}, where tokens attend within the same column, (2) \emph{feature attention}, where tokens attend within the same row, and (3) \emph{neighbor attention}, 
where tokens attend across rows connected by primary--foreign key links. This attention structure
enables information sharing across rows, columns, and tables while preserving
the relational structure of the database. See \Cref{sec:rt_architecture} for details.

\section{\TheJoin{}: A Large-Scale Pretraining Corpus of Relational Data}
\label{sec:corpus}


Existing collections of relational datasets, such as RelBench~\citep{relbench, gu2026relbench}, CTU~\citep{motl2015ctu}, BIRD~\citep{li2023can}, and Spider~\citep{yu2018spider, lei2025spider}
comprise diverse databases but fail to capture the structural and distributional variety of predictive tasks needed for training RFMs.
To address this issue, we introduce \theJoin{}, a corpus comprising 639 relational databases across diverse domains (e-commerce, sports, media, finance, healthcare, etc.) and 6K forecasting tasks spanning various levels of difficulty, including rare events and cold-start entities.
We follow a four-stage approach to curate the corpus: (1) collection of raw databases from the web, (2) standardization of data types in tables, (3) task generation using LLMs, and (4) task filtering using an XGBoost model.
See \Cref{app:add_details_dataset} for details.


\subsection{Collection}
\label{sec:collection}

We collected the raw data from five disjoint source classes: (1) curated relational repositories (CTU, OGB,
OpenAlex), (2) public APIs (SEC EDGAR, CFPB, World
Bank, EIA), (3) domain specific portals (Lahman, ATP, StatsBomb, Cricsheet,
OpenFlights), (4) community dumps (Stack Exchange), and (5) benchmarks (Kaggle,
Spider/BIRD).
Every contributing source is either a publicly downloadable archive or a public API. Licenses vary by source, from public domain and permissive licenses to research-only, non-commercial and provider-specific terms, and these per-source restrictions apply to the released data (\Cref{sec:appendix-licenses}).
The classes carry qualitatively different relational
structures: CTU schemas are small, clean, and follow third-normal-form,
whereas Stack Exchange dumps are wide and denormalized.
As a preliminary quality filter, we only collect databases that have at least two tables joined by a foreign key, and at least one column that
can be parsed as a timestamp. \Cref{tab:domains} lists the domains included in the corpus. To avoid contamination, we exclude
databases that  overlap with  ones in
RelBench~\citep{relbench}, RelBench v2~\citep{gu2026relbench} and 4DBInfer~\citep{wang20244dbinfer} using source identity
rather than schema similarity.



\subsection{Standardization}
\label{sec:standardization}
The databases collected in their raw form are usually messy, with inconsistencies across column data types, timestamps, and primary-foreign key links being the major issues. For example: natively boolean columns are represented as strings \texttt{"true"}/\texttt{"false"}, dates are represented as ISO strings, and foreign keys exist only as a column-naming convention with no explicit declaration. To address such issues, we apply five idempotent detector-and-transformer fixes that
recover the canonical relational structure: (1) \emph{boolean coercion}, rewriting string and
\texttt{uint8} columns whose values fall within a canonical
truth-set as booleans, (2) \emph{numeric coercion}, casting string-encoded
numerics to integers or floats with an identifier blacklist
guarding zip codes and ISBNs, (3) \emph{foreign-key inference} by name match plus an
$\geq\!80\%$ value-containment test against parent primary keys, (4) \emph{temporal normalization}, electing a single
time column per table from regex, name-hint, and integer-year
candidates;
and (5) \emph{entity-axis subsampling} of oversized databases, retaining all activity rows for sampled entities to preserve their full activity histories.
\subsection{Task generation}
\label{sec:tasks}

While autocomplete tasks are naturally defined by simply masking a cell/column in an existing database, in forecasting tasks the target cell/column (e.g., revenue next month) does not appear in the database and has to be explicitly generated. To obtain forecasting tasks we proceed as follows.

\paragraph{Audit and eligibility.} Each database is first parsed
into a typed schema graph with two named roles: an \emph{entity
table} (whose primary key is referenced by a foreign key (FK) from another table,
identifying a long-lived object such as a customer, player, or
hospital) and an \emph{activity table} (with both a timestamp column
and an FK to an entity table, so each row is a timestamped event
attributable to an entity). A database is marked as eligible for task generation if it has at least
one entity table, at least one activity table, $\geq\!5$ unique
timestamps in that activity table, and $\geq\!100$ rows. The first
two conditions ensure the schema can express a per-entity forecasting task. The timestamp and row count conditions allow us to perform XGBoost-based filtering of the generated tasks for quality control. 


\paragraph{Propose and generate.}
Proposing forecasting tasks for a database requires subject-matter expertise and a scalable way to verify correctness. We use an LLM (Claude Opus 4.6) for the former and a lightweight deterministic checker for the latter. For every eligible database, the LLM is given the schema description in text and is asked to write a set of task proposals in JSON. 
Each proposal selects one of $25$ registered template families, grouped into eight reasoning regimes. These include \emph{aggregation}, \emph{threshold/existence}, \emph{trend/change}, \emph{behavior/churn}, \emph{ranking}, \emph{adoption/novelty}, \emph{streak}, and \emph{tail-event}  tasks.
The checker then iterates through each proposal and drops it if: (1) a referenced column does not exist, (2) the entity column is not a foreign key, or (3) the requested pattern is not one of the registered templates. The LLM proposer is run multiple times per database, where each pass sees the tasks already accepted and is asked specifically for templates with low coverage. 
\subsection{Task filtering}
\label{sec:quality}

Finally, we score each task using an XGBoost model over window-aggregation features. We retain tasks that are either \emph{learnable}, where XGBoost beats the lightweight historical baselines,
or  
where the signal is too weak for these baselines but which still act as harder learning objectives for stronger models.
Overall, this process gives us $6{,}023$ forecasting tasks, where each task is defined by its description and a table with timestamped labels (e.g., future revenue for each quarter)\footnote{Our code is publicly available at: \href{https://github.com/stanford-star/agentic-task-gen}{\texttt{https://github.com/stanford-star/agentic-task-gen}}}. See \Cref{app:add_details_dataset} for case studies and a manual audit of the tasks for their practicality and realism.


\section{Pretraining Recipe}
\label{sec:training-methodology}

Here we describe our key improvements to the pretraining recipe
of the original RT paper~\citep{rt},
which enable large-scale pretraining and context-efficient predictions
(\Cref{fig:recipe}).

\subsection{Long-context training}
The context size $L$ is the number of cells presented to RT
and is a key scaling axis to improve prediction quality. 
However, training at large $L$ is computationally expensive.
Further, the same pretrained model should be able to handle
both small and large contexts at test-time,
so that inference compute can be traded off against quality.
Hence, we train at a mix of context sizes.
For each training batch,
we sample $L \sim \mathrm{Unif}(\mathcal{L})$
from a fixed set $\mathcal{L}$ of context sizes.
We keep batch size $B$ constant irrespective of $L$
for stable optimization.
As GPU memory scales as $L \times B$,
we use gradient accumulation adaptively
to prevent out-of-memory (OOM) errors at large $L$. 

\subsection{Retrieval}
The original RT recipe~\citep{rt} samples the relational context
$\mathcal{C}_r$ with a bounded-width breadth-first search (BFS)
over primary and foreign key links
starting from the target row $r$. As the context length $L$ grows, BFS can
exhaust the local neighborhood, retrieve less relevant rows, and miss useful
support rows from the target table. Retrieval is especially important for relational data because contexts are
heterogeneous.  
For example, when predicting whether a user will click on a shoe ad, plain BFS
may spend the context budget on generic user history or similar product
cells.  
We use a two-stage sampling  approach based on random walks to first retrieve similar rows and then expand the local neighborhood of each such row to prepare the context. Our approach can hence prioritize nearby target rows repeatedly
reached through shared users, shoes, and ads tables for high quality contexts.

\paragraph{Stage 1: Random-walks for global retrieval.}
The first stage uses random walks inspired by Pixie~\citep{pixie} and
PinSAGE~\citep{pinsage}.
We run $W$ random walks of length $K$ from the
target row $r$ and record visit counts for rows in the target table $T^\star$.
We rank these rows by visit count, optionally using timestamps when the recency
bias parameter $\rho$ is set to $1$.
The walk parameters $W$ and $K$
approximate the random-walk ordering efficiently and are tuned for latency.
Unlike vector-similarity retrieval, this sampler requires no pre-computation
and can run on the fly across large databases in a highly parallelizable manner.
For example, we use $W=10{,}000$ random walks, each of $K=20$ steps, with zero data loading overhead as the context sampler running on the CPU is ready with the next batch before the forward and backward passes have completed on the GPU.
While random walks add a small latency overhead at test-time,
our sensitivity analysis in \Cref{app:rw-context} shows that lower values of
$W$ and $K$ reduce this overhead significantly while largely retaining accuracy.

\paragraph{Stage 2: BFS for local retrieval.}
For the second stage, we use the ranked rows as seeds for
local BFS with context size $\ell$ and width bound $b$ until the total context
size reaches $L$. The full procedure is given in
\Cref{alg:rw-context} (\Cref{app:rw-context}).
Along with $L$, the retrieval parameters $(\ell, b, \rho)$ can be tuned at test
time.
To allow for this, we need to expose the model to diverse contexts during training.
Hence, we randomly sample
$(\ell, b, \rho) \sim
\mathrm{Unif}(\mathcal{L}_\ell \times \mathcal{B} \times \mathcal{R})$
using the values listed in \Cref{sec:results}.

\section{Pretraining Results}
\label{sec:results}

\paragraph{Pretraining details.}
We use an $86$M-parameter RT model with 12 layers, 512 hidden size and 8 attention heads, incorporating the improvements of \citep{kothapalli2026plurel} (such as QK-Norm) over the original architecture of \citep{rt}. During pretraining, we sample context size $L$ from $\mathcal{L} = \{512, \dots, 8192\}$, the local context size $\ell$ from $\mathcal{L}_\ell = \{128, \dots, 8192\}$, the BFS width-bound $b$ from $\mathcal{B} = \{8, \dots, 256\}$ (all powers of two) and the recency parameter $\rho$ from $\mathcal{R} = \{0, 1\}$, and mask only the target cell (see \Cref{app:pretraining-ablations} for experiments with multi-cell masking).

We pretrain in two phases, first on the synthetic PluRel corpus~\citep{kothapalli2026plurel} ($86{,}211$ tasks over $1{,}900$ databases)\footnote{The dataset is publicly available at: \href{https://huggingface.co/datasets/stanford-star/plurel}{\texttt{https://huggingface.co/datasets/stanford-star/plurel}}} and then on $13{,}243$ tasks from a $523$-database subset of \theJoin{} (excluding the largest DBs due to memory  bottlenecks). We optimize with Muon~\citep{jordan2024muon} at a constant learning rate after warmup and select the best checkpoint after stochastic weight averaging (SWA)~\citep{izmailov2018averaging} as recommended by~\citet{ranjan2024post}. The same checkpoint is used for all classification and regression tasks. Each phase takes about a day on a single node with $8\times$A100 GPUs. See \Cref{app:pretraining-details} for details.

\paragraph{Evaluation.}
We evaluate on RelBench~\citep{relbench},
which comprises $21$ forecasting tasks across $7$ databases
($9$ regression and $12$ binary classification tasks).
These tasks span rare-event, cold-start, heavy-tailed and long-horizon regimes (see \Cref{sec:appendix-relbench-v1-tasks} for task descriptions).
All databases used at evaluation are held out from pretraining.
Following the official RelBench leaderboard, we use Z-score normalized mean absolute error (nMAE, lower is better) for regression
and AUROC (higher is better) for classification tasks and report the average metrics. 

\subsection{Baseline comparisons}
\label{sec:baseline-comparisons}

\paragraph{Baselines.}
We compare \rtj{} against the strongest publicly available in-context learning (ICL)
pipelines for relational prediction. A pipeline consists of a featurizer followed by a predictor.
We consider two featurizers: (1) \textbf{RDBLearn}~\citep{rdblearn},
which aggregates relational neighborhoods into a flat feature table 
and (2) an \textbf{LLM Agent} (Claude Code with Opus 4.6),
which explores the database and writes SQL queries
to perform feature engineering
before applying a tabular predictor.
We use two predictors: (1)
\textbf{TabICLv2}, an in-context tabular foundation model~\citep{tabicl,tabiclv2},
and (2) \textbf{LightGBM}~\citep{ke2017lightgbm}, gradient-boosted decision trees
at the library's default hyperparameters.
Other RFMs such as KumoRFM~\citep{kumorfm,kumorfm2}, Griffin~\citep{griffin}, RDB-PFN~\citep{wang2026rdbpfn} and TabPFN-Rel~\citep{tabpfn3-rel,tabpfn3.5-rel} could not be used as baselines for reasons detailed in \Cref{app:other-rfms}.
We compare against \rtp{} in \Cref{sec:pretraining-phases}.

\paragraph{Setup.}


For every task, we sweep the test-time context size
$L \in \{256, 512, 1024, 2048, 4096, 8192\}$
cells for \rtj{},
holding all other context hyperparameters fixed,
ensuring that the $256$-cell context is a subset of the $512$-cell context
and so on.
For a fair label-matched evaluation of the baselines,
we extract the task rows from the context prepared for \rtj{},
but do not further constrain the feature extraction or processing
from the source database.
Any checkpoint selection or hyperparameter tuning is done on the validation split,
and all numbers are reported on the full test splits.


\begin{figure}[t]
  \centering
  \includegraphics[width=\linewidth]{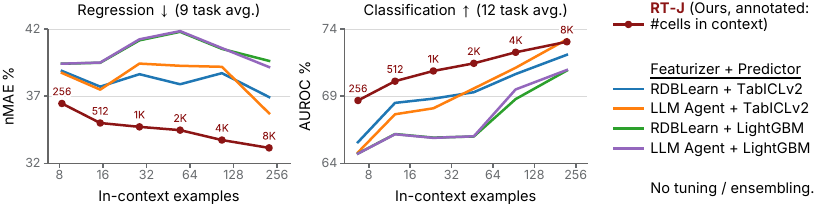}
  \caption{
  \rtj{} vs baselines.
    On regression (left), \rtj{} dominates
    baselines at every example count, by up to $5$ nMAE points
    over the strongest one. On classification (right), \rtj{} leads by up to $3$ AUROC points, especially when examples are scarce.
    Per-task plots are in \Cref{app:additional-results}.
    }
  \label{fig:baselines}
\end{figure}

On regression, \rtj{} beats every baseline at every context size, by $2.3$--$5.0$ nMAE points (see \Cref{fig:baselines}). It is also far more example-efficient: with $15$ in-context examples it already beats the best error any baseline reaches with $225$, and the strongest pipeline needs around $2300$ examples, $10\times$ more, to surpass it. Since all methods see the same examples, we hypothesize that the gap comes from \rtj{} better exploiting the relational context around them, which baselines flatten into a set of features. On classification, \rtj{} leads by $3.1$ AUROC points when examples are scarcest, but the baselines improve steeply with more examples and the strongest catches up at the rightmost end. Among the baselines, the predictor matters more than the featurizer: TabICLv2 beats LightGBM at every context size on both metrics, while LLM Agent and RDBLearn features perform similarly.

\subsection{Impact of synthetic and real pretraining phases}
\label{sec:pretraining-phases}

\paragraph{Setup.}
We isolate the contribution of each pretraining phase
by comparing three RT models with the same architecture:
randomly initialized RT,
\rtp{} (PluRel phase only)
and \rtj{} (PluRel followed by \theJoin{}),
on both fine-tuning and in-context learning regimes.
These pipelines include test-time scaling (\Cref{sec:test-time-compute}) and reflect our submissions to the official RelBench leaderboard. \Cref{app:leaderboards} details both pipelines and reports per-task results.

\paragraph{Observations.} \Cref{tab:leaderboard-summary} shows the results. Synthetic pretraining on PluRel improves fine-tuning over a randomly initialized RT by $1.7$ nMAE and $0.8$ AUROC points, and continued pretraining on \theJoin{} adds $0.5$ nMAE and $0.5$ AUROC points on top. The real-data phase matters most in-context, where \rtj{} improves over \rtp{} by $0.6$ nMAE and $1.5$ AUROC points.

\begin{table}[t]
\centering
\caption{Regression nMAE and classification AUROC averaged over all RelBench tasks for RT models with no pretraining (RT), synthetic-only pretraining (\rtp) and synthetic + real pretraining (\rtj), for both fine-tuning and in-context learning settings. Per-task results are in \Cref{app:leaderboards}.}
\label{tab:leaderboard-summary}
\small
\begin{tabular}{llcccc}
\toprule
 & & \multicolumn{2}{c}{Fine-tuned} & \multicolumn{2}{c}{In-context} \\
\cmidrule(lr){3-4}\cmidrule(lr){5-6}
Model & Pretraining data & nMAE \% $\downarrow$ & AUROC \% $\uparrow$ & nMAE \% $\downarrow$ & AUROC \% $\uparrow$ \\
\midrule
RT & none (random init) & $29.26$ & $77.76$ & -- & -- \\
\rtp{} & PluRel & $27.57$ & $78.53$ & $32.95$ & $72.98$ \\
\rtj{} & PluRel $\to$ \theJoin{} & $\mathbf{27.11}$ & $\mathbf{79.02}$ & $\mathbf{32.35}$ & $\mathbf{74.45}$ \\
\bottomrule
\end{tabular}

\end{table}

\section{Test-Time Compute Scaling}
\label{sec:test-time-compute}



Considering the fixed context window of \rtj{}, choosing informative input cells for a predictive task becomes crucial.
As the optimal context hyperparameters can be task dependent, and sampling cells with a particular configuration is inherently random, we employ (1) context tuning and (2) context ensembling as two complementary ways to improve predictions of \rtj{}. The compute, time and label overheads of tuning and ensembling approaches are compared in \Cref{app:ttc-cost}. Here we use a subsampled test set for computational feasibility.

\subsection{Context hyperparameter tuning}
\label{sec:ttc-hparam-tune}


Intuitively, during inference with \rtj{}, different tasks might prioritize
width vs.\ depth of BFS, local vs.\ global neighborhoods, and recent vs.\
long-term patterns differently. We therefore tune
$(L, \ell, b, \rho)$ separately for each task on its validation split and apply the best configuration on the test split.
The full protocol and the selected per-task configurations are given in \Cref{tab:val_vs_test_ctx8k} (\Cref{app:additional-results}).


\paragraph{Observations.} From \Cref{fig:ensemble}, we notice that context tuning is primarily useful for regression.
With a single context seed, it reduces mean nMAE from $33.4$ to $32.5$, and improves eight of the nine regression tasks. The largest gains occur on \texttt{rel-trial/site-success} ($91.0$ $\to 88.6$) and on \texttt{rel-hm/item-sales} ($14.0$ $\to 11.8$), the latter by selecting a smaller $512$-cell context.
For classification, tuning \emph{hurts} a single seed on average ($72.9$ $\to 72.4$ AUROC), as the validation choice transfers to test on only five of twelve tasks (\Cref{tab:val_vs_test_ctx8k} in \Cref{app:additional-results}).
Moreover, among the selected configurations, three of nine regression tasks (\texttt{rel-amazon/item-ltv}, \texttt{rel-amazon/user-ltv}, \texttt{rel-hm/item-sales}) and four of twelve classification tasks (\texttt{rel-avito/user-visits}, \texttt{rel-event/user-ignore}, \texttt{rel-event/user-repeat}, \texttt{rel-stack/user-engagement}) prefer a context with $L<8192$ cells.

\subsection{Context ensembling}
\label{sec:ttc-context-ensembling}

Both the random-walk retrieval and the bounded-width local BFS
stages of our context sampler are stochastic, with different random seeds yielding different yet equally viable contexts.
Context ensembling simply averages the predictions made from
$n_\mathrm{ens}$ independently sampled contexts. Ensembling reduces prediction noise due to retrieval by averaging over contexts and can be computed in parallel across $n_\mathrm{ens}$.
We sweep over $n_\mathrm{ens} \in \{1, 2, 4, 8, 16\}$ using different sampler seeds with a default configuration of $L = 8192$, $\ell = 256$, $b = 32$, $\rho = 1$ and showcase the benefits.

\paragraph{Observations.}

\begin{figure}[t]
  \centering
  \begin{minipage}[t]{0.49\linewidth}
    \centering
    \includegraphics[width=\linewidth]{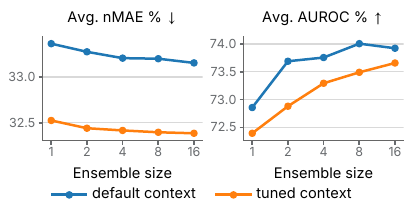}
    \caption{Context ensembling at test time, with
    default and per-task tuned context hyperparameters.
      Ensembling
      improves both metrics for both configurations.
      Per-task plots are in \Cref{app:additional-results}.
      }
    \label{fig:ensemble}
  \end{minipage}\hfill
  \begin{minipage}[t]{0.49\linewidth}
    \centering
    \includegraphics[width=\linewidth]{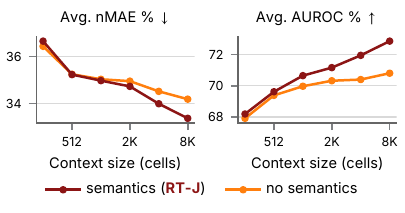}
    \caption{Schema-semantics ablation. Shuffling table and column
      names at evaluation degrades results
      increasingly with context.
      Per-task plots are in \Cref{app:additional-results}.
      }
    \label{fig:semantics}
  \end{minipage}
\end{figure}

\Cref{fig:ensemble} shows the results. With the per-task-tuned configuration, increasing $n_\mathrm{ens}$ from $1$ to $16$ yields a $+1.3$ AUROC point aggregate gain for classification tasks, while at the default context it is slightly lower at $+1.1$ AUROC points.
Regression tasks show smaller but consistent improvements ($0.2$ nMAE points on average at the default context). Based on the per-task curves in \Cref{app:additional-results}, the largest default-context gains accrue to \texttt{rel-trial/study-outcome} with $+5.3$, \texttt{rel-event/user-repeat} with $+2.6$ and \texttt{rel-hm/user-churn} with an improvement of $+2.4$ AUROC points from
$n_\mathrm{ens}{=}1$ to $16$, while \texttt{rel-avito/user-clicks} degrades by $3.0$ AUROC points. On regression tasks the
gains top out at $0.7$ nMAE points on \texttt{rel-event/user-attendance}.

\section{Ablation Studies}
\label{sec:ablations}

Here we isolate the impact of the key components of our recipe.
The schema-semantics (\Cref{sec:abl-semantics}) and context-retrieval
(\Cref{sec:abl-retrieval}) ablations swap a single component of the full
\rtj{} pipeline at inference time, evaluated on a fixed $8192$-row subsampled test set
per task.
\Cref{tab:attribution} summarizes the gain from various components. More pretraining ablations are in \Cref{app:pretraining-ablations}.

\subsection{Schema semantics}
\label{sec:abl-semantics}

\rtj{} consumes language model (LM) embeddings of table/column names
in addition to cell values (see also the cell embedding design in  \Cref{sec:rt_architecture}).
\Cref{fig:semantics} shows a comparison with and without such semantics,
where semantics are removed by shuffling table/column names at
evaluation time.

\paragraph{Observations.}
Shuffling table/column names degrades classification at every point of
the sweep, by up to $2.1$ AUROC points, and regression by up to $0.8$
nMAE points.
Notably, the gap widens with more in-context examples, which suggests that schema text provides prior information that the examples do not replace.
These observations highlight that \rtj{} has evidently
absorbed world knowledge attached to real table and column names during
pretraining on \theJoin.

\begin{figure}[t]
  \centering
  \includegraphics[width=\linewidth]{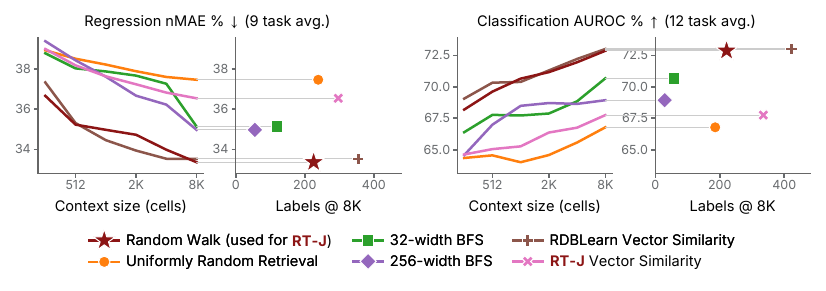}
  \caption{Context-retrieval ablation with \rtj{} at test-time.
    Left sub-panels plot the metric against context size $L$ and right
    sub-panels plot it against the average labels delivered at $L = 8192$ (non-label cells provide relational context).
    Random-walk (ours) is competitive with the best, despite being cheaper and more scalable.
    \rtj{} generalizes to other retrievers too.
    Retriever quality is not strongly correlated with the label count; relational context is important too.
    Per-task plots are in \Cref{app:additional-results}.
    }
  \label{fig:retriever}
\end{figure}

\subsection{Alternatives to random-walk retrieval}
\label{sec:abl-retrieval}

We compare our random-walk retrieval approach
(\Cref{sec:training-methodology}) against five alternatives at
inference:
uniformly random task-row selection, BFS of width $\{32, 256\}$ rooted at the target (the original scheme
of \citep{rt}), and
two vector-similarity retrievers that score task rows by
RDBLearn features or \rtj{} embeddings. In particular, the \rtj{} embedding of a task row is the final hidden state at its masked target cell,  computed over a $256$-cell BFS context around the row. Both similarity retrievers rank labeled task rows dated before the target by cosine similarity and use them, in that order, as seeds for the same local BFS as random-walk retrieval.

\paragraph{Observations.} 
\Cref{fig:retriever} shows that 
our random walk approach and similarity search over RDBLearn's precomputed features are well ahead of the other retrievers and effectively tie at $L=8192$ ($33.4$ versus $33.5$ nMAE and $72.9$ versus $73.0$ AUROC). Random walk leads regression at the two smallest contexts ($L=256$, by $0.7$ nMAE, and $L=512$) and again at $L=8192$, whereas the similarity retriever is ahead by $0.4$--$0.8$ nMAE at $L=1024$--$4096$ and by at most $0.9$ AUROC.


The scatter panels in \Cref{fig:retriever} separate label count from label relevance. RDBLearn-feature similarity packs the most labels into the context ($394$) and reaches parity with random walk performance (that uses $223$ labels at $L=8192$).
However, uniform random selection retrieves a similar count ($208$) and trails by $4.1$ nMAE and $6.1$ AUROC. \rtj{}-embedding similarity stays $2.3$--$3.2$ nMAE and $3.6$--$5.4$ AUROC behind random walk at every $L$, yet similarity over RDBLearn's engineered features keeps improving with $L$, highlighting the importance of the embedding quality for ranking relevant task rows.
The two BFS variants are restricted to their fixed local neighborhoods and trail random walk by $1.8$ and $1.6$ nMAE with $2.7\times$ and $5.6\times$ fewer labels, respectively. Overall, our random walk retriever sits at the label-efficient end of this trade-off as it does not need per-task feature pipelines that RDBLearn requires, while providing higher-quality contexts for predictions.

\begin{table}[t]
\centering
\caption{Summary of gains, in nMAE (regression) and AUROC (classification) points.}
\label{tab:attribution}
\small
\setlength{\tabcolsep}{1.5pt}
\begin{tabular*}{\linewidth}{@{\extracolsep{\fill}}lrrlrr@{}}
\toprule
Comparison & Reg. & Clf. & Comparison & Reg. & Clf. \\
\cmidrule(r){1-3}\cmidrule(l){4-6}
\multicolumn{3}{@{}l}{\textit{Pretraining data}} & \multicolumn{3}{l}{\textit{Context retrieval}} \\
\hspace{0.5em}\rtj{} vs.\ \rtp{} & \cellcolor[HTML]{EEF3F2}$+0.60$ & \cellcolor[HTML]{E3EBEA}$+1.47$ & \hspace{0.5em}random walk vs.\ width-256 BFS & \cellcolor[HTML]{D2DFDC}$+1.60$ & \cellcolor[HTML]{B4CAC6}$+3.93$ \\
\hspace{0.5em}\theJoin{}-only vs.\ PluRel-only & \cellcolor[HTML]{CCDBD8}$+1.79$ & \cellcolor[HTML]{C9D9D6}$+2.83$ & \hspace{0.5em}random walk vs.\ uniform random & \cellcolor[HTML]{8BADA6}$+4.09$ & \cellcolor[HTML]{8BADA6}$+6.09$ \\
\midrule
\multicolumn{6}{@{}l}{\textit{Test-time compute}} \\
\hspace{0.5em}tuned vs.\ default context & \cellcolor[HTML]{E7EEED}$+0.84$ & $-0.47$ & \hspace{0.5em}16 context ensemble vs.\ 1 context & \cellcolor[HTML]{F9FBFA}$+0.21$ & \cellcolor[HTML]{EBF1F0}$+1.06$ \\
\bottomrule
\end{tabular*}

\end{table}

\section{Discussion and Conclusion}
\label{sec:discussion}

Our results show that \rtj{} can make accurate predictions on new datasets with as few as a hundred examples, in a single forward pass.
This makes it practical to solve a wide range of predictive tasks without building dedicated feature engineering pipelines,
or relying on extensive labeled data.
Examples include flagging likely churn or fraud
in a small-scale enterprise without a large customer base;
predicting outcomes of novel clinical trials with small cohorts;
and answering one-off predictive questions from an analyst.
\rtj{} could be exposed directly in the database system, allowing users to request quick ad-hoc forecasts
without an expensive labeling step.
It could also be offered as a tool for LLM agents to call when a question requires predicting an unknown value,
unlike transforming existing ones with SQL.
\TheJoin{} is a first step toward capturing
the immense diversity of
real-world applications of
relational foundation models (RFMs) like \rtj{}.
Beyond synthetic pretraining corpora prevalent for structured data foundation models,
\theJoin{} offers hundreds of real databases across science, healthcare, finance, sports, etc.,
with semantically meaningful predictive tasks spanning rare-event, cold-start, heavy-tailed and long-horizon regimes.

\paragraph{Conclusion.}
We developed a pretraining recipe for relational transformers (RTs)~\citep{rt} with long-context training, random-walk retrieval of high-quality in-context examples, and test-time compute scaling via context tuning and ensembling. With this recipe, we obtain \rtj{} by first pretraining on PluRel~\citep{kothapalli2026plurel} synthetic data and then on \theJoin{}, our large-scale real data corpus.
\rtj{} matches strong feature engineering pipelines
with as much as $10\times$ fewer examples.
Experiments highlight the role of real data pretraining, schema semantics and context engineering.
Our work moves RFMs closer to practical systems that can solve novel predictive tasks over databases on the fly.

\paragraph{Limitations and future work.}
\rtj{} has been trained up to $8{,}192$ context cells;
future work can explore efficient longer context training,
which would allow both for more in-context examples,
and larger relational contexts for them.
\rtj{} is unable to solve link prediction and recommendation tasks, sharing this limitation with RT~\citep{rt}.
Our evaluation is limited to RelBench,
with only $21$ tasks on $7$ databases---inspired by \theJoin{},
future work can expand the scope of real-world relational benchmarks.
We release
(1) the RT-PluRel and \rtj{} models,
(2) \theJoin{} data,
(3) the LLM agent task curation pipeline, and
(4) our pretraining, fine-tuning and inference code
at \href{https://star-project.stanford.edu/rt-j}{\texttt{star-project.stanford.edu/rt-j}},
hoping to accelerate future research and adoption of RFMs.

\begin{ack}

We thank
Anvita Gupta,
Joe Meyer,
Johannes Hoffart,
Jonas De Schouwer,
Lata Nair,
Mahmoud Mohammadi,
Marcel Roed,
Mark Znidar,
Matthias Fey,
Michael Bereket,
Rok Sosic,
Ryan Welch,
Valter Hudovernik,
Vijay Prakash Dwivedi,
Viswa Ganapathy,
Yanay Rosen, and
Yangyi Shen
for discussions, feedback and support.
We thank the anonymous reviewers for helping improve the paper.
We also gratefully acknowledge the support of NSF under Nos. CCF-1918940 (Expeditions), DMS-2327709 (IHBEM), IIS-2403318 (III); NIH under No. 1U24NS146314-01, Gates Foundation, Stanford Data Applications Initiative, Wu Tsai Neurosciences Institute, Stanford Institute for Human-Centered AI, Chan Zuckerberg Initiative, Amazon, Genentech, SAP, and SCBX. RR is supported by the Amazon Core AI Fellowship. The content is solely the responsibility of the authors and does not necessarily represent the views of the funding entities.

\end{ack}

\bibliography{references}
\bibliographystyle{unsrtnat}


\appendix
\crefalias{section}{appendix}
\crefalias{subsection}{subappendix}
\crefalias{subsubsection}{subsubappendix}
\newpage

\section{Leaderboard Results}
\label{app:leaderboards}

\begin{figure}[h]
  \centering
  \includegraphics[width=0.6\linewidth]{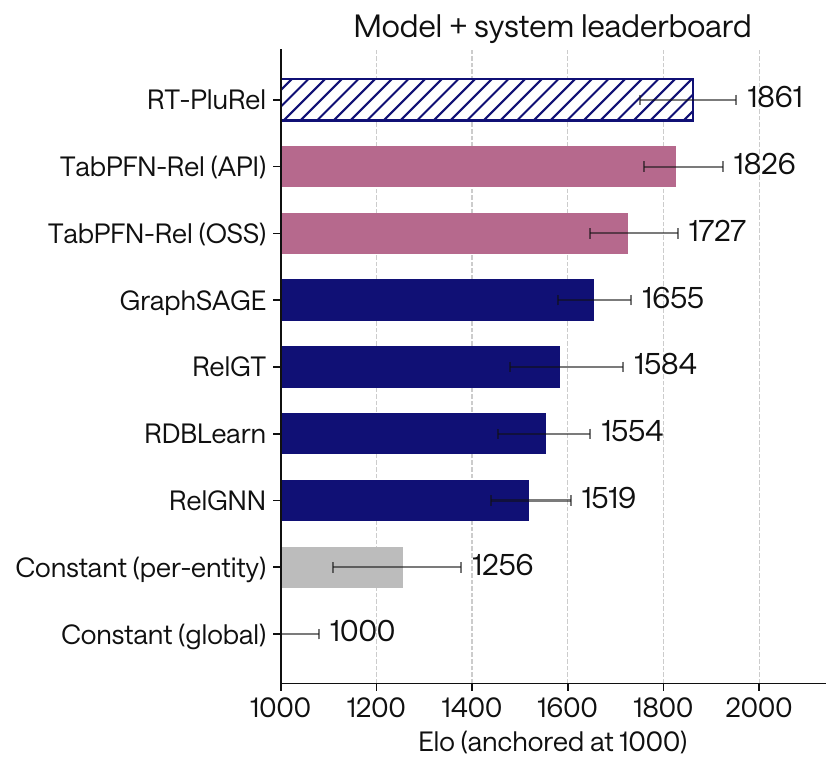}
  \caption{Combined model + system leaderboard of RelArena-$\alpha$, reproduced from \citet{hayler2026relarena}: Elo over the $21$ RelBench v1 forecasting tasks (anchored at $1000$ for the global constant predictor). \rtp{} (hatched) is a system submission with its own tuning protocol; TabPFN-Rel variants in pink, other baselines in blue, constant predictors in grey.}
  \label{fig:relarena}
\end{figure}

\paragraph{RelArena-$\alpha$.} RelArena-$\alpha$~\citep{hayler2026relarena} re-runs every submission on the $21$ entity-level forecasting tasks of RelBench v1 under a single API, data state and evaluation protocol, and aggregates per-task results into Elo ratings. It separates \emph{models}, tuned under a standardized, compute-matched tuning regime, from \emph{systems}, which follow the same data states and evaluation but bring their own tuning protocol. \Cref{fig:relarena} reproduces the combined model + system leaderboard from \citet{hayler2026relarena}. \rtp{}, submitted as a system with its own sequential tuning regime, ranks first with an Elo of $1861$, ahead of TabPFN-Rel (API) at $1826$ and of all relational deep learning baselines (GraphSAGE, RelGT, RelGNN) and RDBLearn. \rtj{} was not part of this release.

\paragraph{RelBench leaderboard.} \Cref{tab:leaderboard} lists the per-task and mean test metrics of the RT family on the RelBench leaderboard, both fine-tuned on each task and in-context. In-context, \rtj{} improves over \rtp{} on both regression ($32.35$ vs.\ $32.95$ nMAE) and classification ($74.45$ vs.\ $72.98$ AUROC). The gains from \theJoin{} carry over to fine-tuning, where \rtj{} again has the best means of the three RT models ($27.11$ nMAE and $79.02$ AUROC, against $27.57$ and $78.53$ for \rtp{} and $29.26$ and $77.76$ for RT). We detail the pipelines used for our submissions below.

\paragraph{Fine-tuning pipeline.} Each model is fine-tuned separately on each of the $21$ tasks in four stages. (1) A selection run trains on the train split for up to $50$K steps, evaluating every $100$ steps on $1024$ validation rows with a $4$-seed context ensemble and stopping after $10$K steps without improvement. (2) A context search scores $96$ context configurations $(L, \ell, b, \rho)$ with $L \le 1024$ on $4096$ validation rows. (3) A refit trains on train and validation together for the selected number of steps, scaled by the ratio of train-plus-validation to train rows. (4) The SWA weights of the refit predict the full test split under the selected context hyperparameters with an $8$-seed context ensemble. \rtp{} and \rtj{} train an additive delta on their frozen pretrained weights, whereas RT trains all weights from random initialization. All runs use Muon with learning rate $5\times10^{-4}$ and weight decay $0.1$, batches of $256$ contexts of up to $1024$ cells, binary cross-entropy for classification and L1 loss for regression.

\paragraph{In-context pipeline.} The pretrained checkpoint is frozen, and no RelBench database is seen in pretraining. Per task, we score $120$ context configurations $(L, \ell, b, \rho)$ with $L \le 8192$ on $4096$ validation rows with a $4$-seed context ensemble, then run each of the four best configurations with four context seeds on the full test split and average the $16$ predictions per row.

\begin{table}[h]
\centering
\caption{\textbf{RT models on the RelBench leaderboard.} Per-task test metrics of RT~\citep{rt}, \rtp{} and \rtj{}, fine-tuned on each task or used in-context, as listed on the RelBench leaderboard. All metrics in percent. Best per row within each group in bold; lower is better for nMAE, higher is better for AUROC.}
\label{tab:leaderboard}
\setlength{\tabcolsep}{3pt}
\begin{tabular}{llrrrrr}
\toprule
 & & \multicolumn{3}{c}{Fine-tuned} & \multicolumn{2}{c}{In-context} \\
\cmidrule(lr){3-5}\cmidrule(lr){6-7}
DB & Task & RT (rand init) & \rtp{} & \rtj{} & \rtp{} & \rtj{} \\
\midrule
\multicolumn{7}{l}{\textit{Regression (nMAE \%, lower is better)}} \\
\midrule
\texttt{rel-amazon} & \texttt{item-ltv} & $7.50$ & $7.26$ & $\mathbf{7.19}$ & $8.65$ & $\mathbf{7.81}$ \\
\texttt{rel-amazon} & \texttt{user-ltv} & $\mathbf{24.15}$ & $24.58$ & $24.29$ & $29.60$ & $\mathbf{29.14}$ \\
\texttt{rel-avito} & \texttt{ad-ctr} & $38.53$ & $37.86$ & $\mathbf{36.03}$ & $\mathbf{41.71}$ & $42.01$ \\
\texttt{rel-event} & \texttt{user-attendance} & $32.52$ & $31.64$ & $\mathbf{31.63}$ & $43.78$ & $\mathbf{42.69}$ \\
\texttt{rel-f1} & \texttt{driver-position} & $48.75$ & $39.72$ & $\mathbf{37.67}$ & $42.41$ & $\mathbf{38.75}$ \\
\texttt{rel-hm} & \texttt{item-sales} & $7.75$ & $\mathbf{7.53}$ & $7.57$ & $11.44$ & $\mathbf{10.11}$ \\
\texttt{rel-stack} & \texttt{post-votes} & $13.77$ & $13.62$ & $\mathbf{12.79}$ & $\mathbf{14.70}$ & $14.74$ \\
\texttt{rel-trial} & \texttt{site-success} & $78.36$ & $\mathbf{75.17}$ & $76.54$ & $\mathbf{87.65}$ & $89.62$ \\
\texttt{rel-trial} & \texttt{study-adverse} & $12.06$ & $10.74$ & $\mathbf{10.28}$ & $16.57$ & $\mathbf{16.33}$ \\
\midrule
\textbf{Mean} & & $29.26$ & $27.57$ & $\mathbf{27.11}$ & $32.95$ & $\mathbf{32.35}$ \\
\midrule
\multicolumn{7}{l}{\textit{Classification (AUROC \%, higher is better)}} \\
\midrule
\texttt{rel-amazon} & \texttt{item-churn} & $82.96$ & $\mathbf{83.19}$ & $83.16$ & $74.40$ & $\mathbf{78.61}$ \\
\texttt{rel-amazon} & \texttt{user-churn} & $71.30$ & $71.34$ & $\mathbf{71.46}$ & $66.24$ & $\mathbf{67.20}$ \\
\texttt{rel-avito} & \texttt{user-clicks} & $\mathbf{68.59}$ & $65.89$ & $66.81$ & $46.20$ & $\mathbf{51.39}$ \\
\texttt{rel-avito} & \texttt{user-visits} & $66.21$ & $66.75$ & $\mathbf{66.89}$ & $61.44$ & $\mathbf{64.09}$ \\
\texttt{rel-event} & \texttt{user-ignore} & $81.41$ & $80.85$ & $\mathbf{85.61}$ & $76.78$ & $\mathbf{80.97}$ \\
\texttt{rel-event} & \texttt{user-repeat} & $78.97$ & $\mathbf{80.29}$ & $79.84$ & $77.65$ & $\mathbf{79.12}$ \\
\texttt{rel-f1} & \texttt{driver-dnf} & $79.93$ & $82.09$ & $\mathbf{82.77}$ & $81.89$ & $\mathbf{83.08}$ \\
\texttt{rel-f1} & \texttt{driver-top3} & $83.84$ & $90.01$ & $\mathbf{91.71}$ & $89.69$ & $\mathbf{91.00}$ \\
\texttt{rel-hm} & \texttt{user-churn} & $70.24$ & $\mathbf{70.61}$ & $70.49$ & $\mathbf{65.77}$ & $65.66$ \\
\texttt{rel-stack} & \texttt{user-badge} & $\mathbf{89.33}$ & $89.25$ & $89.21$ & $\mathbf{83.78}$ & $81.11$ \\
\texttt{rel-stack} & \texttt{user-engagement} & $90.74$ & $\mathbf{90.91}$ & $90.65$ & $\mathbf{87.86}$ & $86.76$ \\
\texttt{rel-trial} & \texttt{study-outcome} & $69.55$ & $\mathbf{71.21}$ & $69.57$ & $64.08$ & $\mathbf{64.43}$ \\
\midrule
\textbf{Mean} & & $77.76$ & $78.53$ & $\mathbf{79.02}$ & $72.98$ & $\mathbf{74.45}$ \\
\bottomrule
\end{tabular}

\end{table}



\section{Other Relational Foundation Models}
\label{app:other-rfms}
Beyond the pipelines of \Cref{sec:results}, we looked into comparing \rtj{} with recent relational foundation models (RFMs), but none could be evaluated under our protocol, in which every method sees the same in-context examples.

\textbf{KumoRFM}~\citep{kumorfm}: closed weights, and API access does not allow controlling the in-context examples, precluding a fair comparison.

\textbf{KumoRFM-2}~\citep{kumorfm2}: closed weights. API access was released only a few weeks before our submission deadline, and did not work during rebuttal due to the ongoing migration to NVIDIA.

\textbf{Griffin}~\citep{griffin}: the released checkpoints are incapable of in-context learning, and pretraining Griffin with our recipe is not feasible, as the recipe is engineered around transformer-based architectures, which Griffin is not.

\textbf{RDB-PFN}~\citep{wang2026rdbpfn}: supports only classification tasks, whereas our evaluation covers both regression and classification.

\textbf{TabPFN-Rel} with TabPFN-3 and TabPFN-3.5~\citep{tabpfn3-rel,tabpfn3.5-rel}: released after our submission deadline. It is compared against \rtp{} on RelArena-$\alpha$ in \citet{hayler2026relarena} (\Cref{app:leaderboards}), where \rtp{} ranks first (Elo $1861$ vs.\ $1826$ for TabPFN-Rel (API-only)); evaluating it under our matched-example protocol is left to future work. See the latest RelArena-$\alpha$ leaderboard, with \rtj{}, TabPFN-3.5 and other newer submissions, at \href{https://github.com/PriorLabs/relarena\#-leaderboards}{\texttt{https://github.com/PriorLabs/relarena\#-leaderboards}}.

\section{Related Work}
\label{sec:related_work}

\paragraph{Relational deep learning and benchmarks.}

Relational deep learning (RDL) is a machine learning paradigm that operates directly on relational databases, avoiding the need to manually flatten a database into a single feature table~\citep{rdl}. A common formulation represents a database as a heterogeneous relational entity graph: rows are nodes, tables define node types, and primary--foreign-key links define typed edges~\citep{relbench}. Graph neural networks can then propagate information across related rows and tables, enabling end-to-end prediction over multi-table data~\citep{relbench,relgnn}. Recent work has improved the efficiency and expressiveness of relational message passing~\citep{relgnn}, and transformer-based relational architectures have also been proposed~\citep{dbformer,relgt}. \citet{kanatsoulis2025learning} introduce an efficient positional encoding framework based on lightweight message-passing operations, providing structural features that have been shown to improve relational deep learning models. Broader surveys and position papers frame these models as part of a larger agenda for relational deep learning~\citep{dwivedi2025relational}.

Benchmarking has been central to this progress. RelBench~\citep{relbench} introduced a standardized suite of relational prediction tasks, while 4DBInfer~\citep{wang20244dbinfer} and RelBench v2~\citep{gu2026relbench} expand the scale and diversity of relational benchmarks, including broader task families and database domains. These benchmarks make it possible to compare relational models under realistic multi-table and temporal constraints. 

\paragraph{Relational foundation models.}

Recent work has begun to develop foundation models specifically for relational data. These models seek to transfer across schemas and tasks rather than training a separate predictor for every database. KumoRFM and KumoRFM-2 study in-context learning for relational data at scale~\citep{kumorfm,kumorfm2}. Griffin develops a graph-centric relational foundation model for transfer across relational databases~\citep{griffin}. The Relational Transformer (RT) takes a complementary cell-level approach: database cells are treated as tokens, schema semantics are included in the input, and relational attention masks encode primary--foreign-key structure~\citep{rt}. PluRel~\citep{kothapalli2026plurel} enables RT pretraining with synthetic data. It synthesizes multi-table databases from scratch, including schemas, primary--foreign-key links, and feature distributions, enabling controlled studies of synthetic data scaling for relational foundation models.

Our work builds directly on RT because its cell-level formulation is well-suited to context-efficient relational prediction. Labeled task rows, unlabeled rows, schema metadata, and linked evidence can all appear in the same context window, making masked cell prediction a natural interface for both pretraining and inference. However, the original RT work primarily establishes a schema-agnostic architecture; it does not develop a scalable pretraining and inference recipe.
We therefore use RT as the backbone and study how to make it effective when only a small number of in-context examples are available.

\paragraph{Tabular foundation models and relational tabularization.}

Tabular foundation models aim to bring pretraining and in-context learning to single-table prediction. TabPFN and its successors show that transformers can perform strong
in-context prediction when labeled rows from a tabular task are provided in the context~\citep{tabpfn,tabpfnv2}. More recent models, including TabICLv2 and other large-scale tabular foundation models, extend this paradigm to larger datasets, richer feature spaces, and broader tabular distributions~\citep{tabicl,tabiclv2,portal,carte}. These methods provide strong evidence that structured prediction can benefit from foundation-model-style training.

Relational databases, however, contain structure that is not present in ordinary single tables. Predictive evidence may lie several joins away from the target row, and the relevant signal may involve temporal histories, linked entities, or multi-hop dependencies. One way to use tabular foundation models in this setting is to first transform the database into a single feature table. RDBLearn follows this strategy by aggregating relational neighborhoods into engineered features and then applying a tabular in-context learner~\citep{rdblearn}. This approach is effective and provides a strong baseline, but the flattening step changes the input representation: the model no longer reasons directly over cells, schemas, timestamps, primary--foreign-key links, or multi-table structure.

\paragraph{Large language models for relational data.}

Another line of work applies large language models to structured and relational data by serializing tables or database neighborhoods into text~\citep{snowflake2024relational}. 
This has the advantage of using powerful pretrained language models without designing a specialized architecture. 
However, serialization can be inefficient and brittle. 
Relational structure must be expressed as text, long database contexts can exceed the model context window, and the resulting input format may differ substantially from the data distribution seen during language-model pretraining. 
Hybrid approaches that combine graph or relational encoders with LLMs can partially address these issues, but they often still depend on task-specific components or limited context construction~\citep{relllm}.

\paragraph{Graph foundation models.}
Several papers have paved the way toward graph foundation models. Mole-BERT~\citep{xia2023mole} pretrains GNNs with masked atom prediction and triplet-based contrastive learning for node- and graph-level drug-discovery tasks. \citet{beaini2024towards} further show that molecular pretraining benefits from scale, using large multi-task datasets with billions of labels and demonstrating that combining quantum and biological supervision improves low-resource performance.

Recent work has extended pretraining graph models beyond molecules. PRODIGY~\citep{huang2023prodigy} introduces prompt-based graph representations for in-context learning. GraphAny~\citep{zhaofully} studies zero-shot node classification across graphs with disjoint feature and label spaces, using LinearGNN ensembles and inductive attention inspired by linear least-squares estimation. ULTRA~\citep{galkintowards} learns transferable relational representations for zero-shot reasoning over unseen knowledge graphs. GraphGPT~\citep{zhaographgpt} converts graphs into reversible token sequences using Eulerian paths, enabling scalable generative Transformer pretraining for graph-structured data. Most recently, \citet{bechler2026billion} scale the graph foundation model paradigm to heterogeneous billion-scale graphs, proposing a billion-parameter Transformer architecture and a recipe for link prediction pretraining.

\section{Additional Details on \theJoin}
\label{app:add_details_dataset}


\subsection{Agentic execution}
\label{sec:appendix-agentic}

Constructing \theJoin~corpus manually would require months of human engineering, particularly for web collection as each source presents different download mechanisms, file formats, schema conventions, parsing failures, and data-quality issues. We therefore used Claude Code agents as the execution layer for corpus construction. The agents identify and inspect candidate sources, implement and debug domain-specific collectors, adapt standardization routines to newly observed edge cases, run the pipeline, inspect logs and intermediate artifacts, and iterate until each stage reaches its completion criteria. The agents were used for engineering and orchestration, while task validation, scoring, filtering, baseline computation, loadability checks, and provenance tracking were performed by deterministic scripts based on human-defined criteria.
\subsection{Phase 1: Collection}
\label{sec:appendix-collect}

\noindent\textbf{Collecting from heterogeneous sources.}
The first phase builds a raw relational snapshot from a broad set of public data sources. This is difficult to automate with a single scraper because different sources expose data in very different ways: as SQL dumps, SQLite databases, CSV bundles, parquet or JSON exports, archive mirrors, or paginated APIs. We therefore use source-specific collectors. Each collector downloads the data, parses it in its native format, recovers table and key metadata when available, and converts the result into a common parquet-based representation. We also exclude
databases that exactly overlap with established benchmark sources in RelBench, RelBench v2 and 4DBInfer, using source identity
rather than schema similarity as the criterion.

\noindent\textbf{Respecting upstream constraints.}
The collectors are designed to be reproducible while remaining respectful of the sources. API-based collectors follow the documented rate limits and include the required identifying information. Bulk-download sources are handled at the archive level, and for sources that publish periodic snapshots, the refresh cadence of our corpus is limited by the cadence of the upstream release.

\noindent\textbf{Normalizing the collected data.}
Before any database is passed to later stages, the collector performs basic normalization. For example, (1) primary and foreign keys are converted to integer indices, (2) timestamps and booleans are stored in consistent types, and (3) mixed-type columns are either normalized or rejected. We also store primary-key, foreign-key, and time-column metadata alongside each Parquet table, so that downstream stages do not rely on source-specific assumptions. \Cref{tab:domains} lists the domains included in the corpus.

{\scriptsize
\setlength{\tabcolsep}{4pt}
\renewcommand{\arraystretch}{1.0}
\begin{longtable}{@{}p{1.0in} p{1.25in} r p{1.38in} p{1.22in}@{}}
\caption{Per-domain corpus breakdown across all collected domains, grouped by italic subject-area headers and sorted by descending DB count. \emph{DBs}: raw count. The \emph{License(s)} and \emph{Ref.} cells list one (license, reference) pair per line with rows aligning across the two columns.
}
\label{tab:domains}
\\
\toprule
\textbf{Domain} & \textbf{Source} & \textbf{DBs} & \textbf{License(s)} & \textbf{Ref.} \\
\midrule
\endfirsthead
\multicolumn{5}{c}{\textit{(continued from previous page)}}\\
\toprule
\textbf{Domain} & \textbf{Source} & \textbf{DBs} & \textbf{License(s)} & \textbf{Ref.} \\
\midrule
\endhead
\midrule\multicolumn{5}{r}{\textit{continued on next page}}\\
\endfoot
\bottomrule
\endlastfoot
\multicolumn{5}{@{}l}{\textit{Commerce \& Online Platforms}} \\
\addlinespace[1pt]
\quad tech community & Stack Exchange / GH Archive / HN & 75 & {\scriptsize No explicit license \newline Open public API \newline CC BY-SA 4.0} & {\scriptsize \href{https://www.gharchive.org/}{gharchive.org} \newline \href{https://hn.algolia.com/api}{hn.algolia.com} \newline \href{https://archive.org/details/stackexchange}{archive.org}} \\
\quad recommendations & Amazon / MovieLens / IMDb / UCI & 34 & {\scriptsize MIT (Amazon ToS) \newline Non-commercial \newline Research-use \newline Research-only, no redistribution \newline Research-only, no redistribution \newline Public Domain \newline CC BY 4.0} & {\scriptsize \href{https://github.com/hyp1231/AmazonReviews2023}{github.com} \newline \href{https://data.imdb.com/non-commercial-datasets}{data.imdb.com} \newline \href{https://cseweb.ucsd.edu/~jmcauley/datasets.html#clothing_fit}{cseweb.ucsd.edu} \newline \href{https://files.grouplens.org/datasets/movielens/ml-25m-README.html}{grouplens.org} \newline \href{https://files.grouplens.org/datasets/movielens/ml-1m-README.txt}{grouplens.org} \newline \href{https://www.research.gov/}{research.gov} \newline \href{https://archive.ics.uci.edu/dataset/502/online+retail+ii}{archive.ics.uci.edu}} \\
\quad prediction market & Manifold / Polymarket & 2 & {\scriptsize ToS-restricted \newline ToS-restricted} & {\scriptsize \href{https://docs.manifold.markets/api}{manifold.markets} \newline \href{https://docs.polymarket.com/}{polymarket.com}} \\
\quad e-commerce & RecSys Challenge 2015 & 1 & {\scriptsize CC BY-NC-ND 4.0} & {\scriptsize \href{https://www.kaggle.com/datasets/chadgostopp/recsys-challenge-2015}{kaggle.com}} \\
\quad real estate & Zillow Research & 1 & {\scriptsize Research-use} & {\scriptsize \href{https://www.zillow.com/research/data/}{zillow.com}} \\
\addlinespace[3pt]
\multicolumn{5}{@{}l}{\textit{Sports, Culture \& Media}} \\
\addlinespace[1pt]
\quad motorsport & Wikipedia / Racing-Reference / MotoGP & 9 & {\scriptsize CC BY-SA 4.0 \newline CC BY-SA 4.0 \newline CC BY-SA 4.0; non-commercial \newline CC BY-SA 4.0 \newline Public unofficial API \newline Non-commercial, no redistribution \newline CC BY-SA 4.0} & {\scriptsize \href{https://en.wikipedia.org/wiki/Champ_Car}{en.wikipedia.org} \newline \href{https://en.wikipedia.org/wiki/Formula_E}{en.wikipedia.org} \newline \href{https://www.racing-reference.info/}{racing-reference.info} \newline \href{https://en.wikipedia.org/wiki/Indy_Lights}{en.wikipedia.org} \newline \href{https://www.motogp.com/}{motogp.com} \newline \href{https://www.racing-reference.info/}{racing-reference.info} \newline \href{https://en.wikipedia.org/wiki/World_Rally_Championship}{en.wikipedia.org}} \\
\quad food & SNAP / Open Food Facts / USDA / Food.com & 6 & {\scriptsize Research-use \newline Research-use \newline ODbL \newline Public Domain \newline Kaggle ToS} & {\scriptsize \href{http://snap.stanford.edu/data/web-BeerAdvocate.html}{snap.stanford.edu} \newline \href{http://snap.stanford.edu/data/web-CellarTracker.html}{snap.stanford.edu} \newline \href{https://world.openfoodfacts.org/terms-of-use}{world.openfoodfacts.org} \newline \href{https://fdc.nal.usda.gov/}{fdc.nal.usda.gov} \newline \href{https://www.kaggle.com/datasets/shuyangli94/food-com-recipes-and-user-interactions}{kaggle.com}} \\
\quad sports & ATP / Cricsheet / Lahman / StatsBomb & 4 & {\scriptsize CC BY-NC-SA 4.0 \newline CC0 \newline CC BY-SA 3.0 \newline StatsBomb user agreement} & {\scriptsize \href{https://github.com/JeffSackmann/tennis_atp}{github.com} \newline \href{https://cricsheet.org/downloads/}{cricsheet.org} \newline \href{https://sabr.org/lahman-database/}{sabr.org} \newline \href{https://github.com/hudl/open-data/blob/master/LICENSE.pdf}{github.com}} \\
\quad music & FMA / Last.fm / ListenBrainz & 3 & {\scriptsize CC BY 4.0 \newline Research-use \newline CC0} & {\scriptsize \href{https://github.com/mdeff/fma}{github.com} \newline \href{https://grouplens.org/datasets/hetrec-2011/}{grouplens.org} \newline \href{https://listenbrainz.org/data/}{listenbrainz.org}} \\
\quad museum & Met / MoMA / Tate & 3 & {\scriptsize CC0 \newline CC0 \newline CC0} & {\scriptsize \href{https://github.com/metmuseum/openaccess}{github.com} \newline \href{https://github.com/MuseumofModernArt/collection}{github.com} \newline \href{https://github.com/tategallery/collection}{github.com}} \\
\quad gaming & EVE Online / Lichess / Steam & 3 & {\scriptsize Open public API \newline CC0 \newline Research-use} & {\scriptsize \href{https://developers.eveonline.com/license-agreement}{developers.eveonline.com} \newline \href{https://database.lichess.org/}{database.lichess.org} \newline \href{https://cseweb.ucsd.edu/~wckang/}{cseweb.ucsd.edu}} \\
\quad cultural & Europeana / Wikidata & 2 & {\scriptsize CC0 \newline CC0} & {\scriptsize \href{https://www.europeana.eu/en/rights/api}{europeana.eu} \newline \href{https://www.wikidata.org/wiki/Wikidata:Licensing}{wikidata.org}} \\
\quad cultural surveys & D-PLACE / Pew & 2 & {\scriptsize CC BY 4.0 \newline Public-use data (Pew)} & {\scriptsize \href{https://d-place.org/about}{d-place.org} \newline \href{https://www.pewresearch.org/about/terms-and-conditions/}{pewresearch.org}} \\
\quad podcast & Apple Podcasts (Kaggle) & 1 & {\scriptsize CC BY-NC-SA 4.0} & {\scriptsize \href{https://www.kaggle.com/datasets/thoughtvector/podcastreviews}{kaggle.com}} \\
\quad history & Chronicling America & 1 & {\scriptsize Public Domain} & {\scriptsize \href{https://chroniclingamerica.loc.gov/about/}{chroniclingamerica.loc.gov}} \\
\addlinespace[3pt]
\multicolumn{5}{@{}l}{\textit{Government, Finance \& Energy}} \\
\addlinespace[1pt]
\quad US government & BTS / FDA / FDIC / FEC / HMDA / NHTSA & 11 & {\scriptsize Public Domain} & {\scriptsize \href{https://www.usa.gov/government-works}{usa.gov}} \\
\quad energy & BOEM / Ember / OPSD / PHMSA / Sodir / EIA / GeoNuclearData & 10 & {\scriptsize Public Domain \newline CC BY 4.0 \newline CC BY 4.0 \newline CC BY 4.0 \newline Public Domain \newline NLOD \newline Public Domain \newline ODbL} & {\scriptsize \href{https://www.data.boem.gov/}{data.boem.gov} \newline \href{https://github.com/owid/energy-data}{github.com} \newline \href{https://ember-energy.org/data/}{ember-energy.org} \newline \href{https://open-power-system-data.org/legal/}{open-power-system-data.org} \newline \href{https://www.phmsa.dot.gov/data-and-statistics}{phmsa.dot.gov} \newline \href{https://data.norge.no/nlod/en/2.0}{norge.no} \newline \href{https://www.eia.gov/about/copyrights_reuse.php}{eia.gov} \newline \href{https://github.com/cristianst85/GeoNuclearData}{github.com}} \\
\quad international & Eurostat / GDELT / UNHCR / World Bank & 5 & {\scriptsize Eurostat re-use \newline Open (attribution) \newline CC BY 4.0 \newline CC BY 4.0} & {\scriptsize \href{https://ec.europa.eu/eurostat/web/main/help/copyright-notice}{ec.europa.eu} \newline \href{https://www.gdeltproject.org/about.html}{gdeltproject.org} \newline \href{https://www.unhcr.org/refugee-statistics}{unhcr.org} \newline \href{https://datacatalog.worldbank.org/public-licenses}{datacatalog.worldbank.org}} \\
\quad finance & CFPB / SEC EDGAR / PKDD (Berka) & 3 & {\scriptsize Public Domain \newline Public Domain \newline Research-use} & {\scriptsize \href{https://www.consumerfinance.gov/data-research/consumer-complaints/}{consumerfinance.gov} \newline \href{https://www.sec.gov/os/accessing-edgar-data}{sec.gov} \newline \href{https://relational.fel.cvut.cz/dataset/Financial}{relational.fel.cvut.cz}} \\
\quad politics & Our World in Data / V-Dem & 2 & {\scriptsize CC BY 4.0 \newline CC BY 4.0} & {\scriptsize \href{https://ourworldindata.org/faqs#how-is-our-world-in-data-licensed}{ourworldindata.org} \newline \href{https://v-dem.net/data/the-v-dem-dataset/}{v-dem.net}} \\
\quad civic & NYC OpenData (311) & 1 & {\scriptsize Public Domain} & {\scriptsize \href{https://opendata.cityofnewyork.us/overview/#tou}{opendata.cityofnewyork.us}} \\
\quad crypto & CoinGecko & 1 & {\scriptsize API ToS (free tier)} & {\scriptsize \href{https://www.coingecko.com/en/api/documentation}{coingecko.com}} \\
\quad labor & ILOSTAT & 1 & {\scriptsize CC BY 4.0} & {\scriptsize \href{https://ilostat.ilo.org/about/copyright/}{ilostat.ilo.org}} \\
\quad legal & OpenSanctions & 1 & {\scriptsize CC BY-NC 4.0} & {\scriptsize \href{https://www.opensanctions.org/licensing/}{opensanctions.org}} \\
\quad patent & USPTO / PatentsView & 1 & {\scriptsize CC BY 4.0} & {\scriptsize \href{https://data.uspto.gov/support/transition-guide/patentsview}{data.uspto.gov}} \\
\quad UK public data & NHS prescriptions (OGL) & 1 & {\scriptsize OGL v3.0} & {\scriptsize \href{https://www.nationalarchives.gov.uk/doc/open-government-licence/version/3/}{nationalarchives.gov.uk}} \\
\addlinespace[3pt]
\multicolumn{5}{@{}l}{\textit{Health \& Public Services}} \\
\addlinespace[1pt]
\quad clinical & ChEMBL / openFDA & 4 & {\scriptsize CC BY-SA 3.0 \newline Public Domain} & {\scriptsize \href{https://chembl.gitbook.io/chembl-interface-documentation/about}{chembl.gitbook.io} \newline \href{https://open.fda.gov/terms/}{fda.gov}} \\
\quad biology & ClinVar / GBIF / PubChem / Reactome & 4 & {\scriptsize Public Domain \newline CC BY 4.0 \newline Public Domain \newline CC0} & {\scriptsize \href{https://www.ncbi.nlm.nih.gov/clinvar/docs/maintenance_use/}{ncbi.nlm.nih.gov} \newline \href{https://www.gbif.org/terms}{gbif.org} \newline \href{https://www.ncbi.nlm.nih.gov/home/about/policies/}{ncbi.nlm.nih.gov} \newline \href{https://reactome.org/license}{reactome.org}} \\
\quad transport & GTFS feeds / OpenFlights / Citi Bike & 3 & {\scriptsize Per-agency \newline ODbL \newline Free use} & {\scriptsize \href{https://mobilitydata.org/}{mobilitydata.org} \newline \href{https://openflights.org/data.php}{openflights.org} \newline \href{https://citibikenyc.com/data-sharing-policy}{citibikenyc.com}} \\
\quad healthcare quality & CMS Provider Data & 2 & {\scriptsize Public Domain} & {\scriptsize \href{https://data.cms.gov/provider-data}{cms.gov}} \\
\quad education & OULAD (Open University) & 1 & {\scriptsize CC BY 4.0} & {\scriptsize \href{https://research.stem.open.ac.uk/ouanalyse/}{research.stem.open.ac.uk}} \\
\quad public health & WHO GHO & 1 & {\scriptsize Open access} & {\scriptsize \href{https://www.who.int/about/policies/publishing/copyright}{who.int}} \\
\quad agriculture & USDA NASS & 1 & {\scriptsize Public Domain} & {\scriptsize \href{https://quickstats.nass.usda.gov/}{quickstats.nass.usda.gov}} \\
\quad disaster & OpenFEMA & 1 & {\scriptsize Public Domain} & {\scriptsize \href{https://www.fema.gov/about/openfema/data-sets}{fema.gov}} \\
\quad skills & O*NET & 1 & {\scriptsize CC BY 4.0} & {\scriptsize \href{https://www.onetcenter.org/license_db.html}{onetcenter.org}} \\
\addlinespace[3pt]
\multicolumn{5}{@{}l}{\textit{Scientific \& Reference}} \\
\addlinespace[1pt]
\quad text-to-SQL & Spider / BIRD / Spider 2.0 / LiveSQL & 259 & {\scriptsize CC BY-SA 4.0 \newline Public access (Oracle) \newline MIT \newline CC BY-SA 4.0} & {\scriptsize \href{https://bird-bench.github.io/}{bird-bench.github.io} \newline \href{https://livesql.oracle.com/}{livesql.oracle.com} \newline \href{https://github.com/xlang-ai/Spider2}{github.com} \newline \href{https://yale-lily.github.io/spider}{yale-lily.github.io}} \\
\quad CTU repository & CTU Prague Repository & 114 & {\scriptsize Research-use \newline MS sample DB \newline MIT \newline MySQL sample DB \newline Research-use \newline Research-use \newline CC BY-SA 4.0 \newline MS sample DB \newline MS sample DB \newline BSD \newline TPC license \newline TPC license \newline TPC license \newline TPC license} & {\scriptsize \href{https://relational.fel.cvut.cz/}{relational.fel.cvut.cz} \newline \href{https://learn.microsoft.com/en-us/sql/samples/adventureworks-install-configure}{microsoft.com} \newline \href{https://github.com/lerocha/chinook-database}{github.com} \newline \href{https://www.mysqltutorial.org/mysql-sample-database.aspx}{mysqltutorial.org} \newline \href{https://data.imdb.com/non-commercial-datasets}{data.imdb.com} \newline \href{https://files.grouplens.org/datasets/movielens/}{grouplens.org} \newline \href{https://sabr.org/lahman-database/}{sabr.org} \newline \href{https://learn.microsoft.com/en-us/dotnet/framework/data/adonet/sql/linq/downloading-sample-databases}{microsoft.com} \newline \href{https://learn.microsoft.com/en-us/sql/samples/sql-samples-where-are}{microsoft.com} \newline \href{https://dev.mysql.com/doc/sakila/en/sakila-license.html}{dev.mysql.com} \newline \href{https://www.tpc.org/tpcc/}{tpc.org} \newline \href{https://www.tpc.org/}{tpc.org} \newline \href{https://www.tpc.org/tpcds/}{tpc.org} \newline \href{https://www.tpc.org/tpch/}{tpc.org}} \\
\quad temporal network & SNAP / KONECT & 12 & {\scriptsize BSD \newline CC BY 4.0} & {\scriptsize \href{https://snap.stanford.edu/data}{snap.stanford.edu} \newline \href{https://zenodo.org/record/8267758}{zenodo.org}} \\
\quad network & SNAP / KONECT (raw) & 10 & {\scriptsize BSD} & {\scriptsize \href{https://snap.stanford.edu/data}{snap.stanford.edu}} \\
\quad academic & OpenAlex / DBLP / CORDIS / NIH / Unpaywall & 7 & {\scriptsize CC BY 4.0 \newline CC BY 4.0 \newline ODC-BY \newline Public Domain \newline CC0 \newline CC0} & {\scriptsize \href{https://data.europa.eu/euodp/en/data/dataset/cordisH2020projects}{europa.eu} \newline \href{https://data.europa.eu/euodp/en/data/dataset/cordis-HE}{europa.eu} \newline \href{https://dblp.org/faq/How+can+I+download+the+whole+dblp+dataset.html}{dblp.org} \newline \href{https://reporter.nih.gov/exporter}{reporter.nih.gov} \newline \href{https://openalex.org/terms}{openalex.org} \newline \href{https://unpaywall.org/products/snapshot}{unpaywall.org}} \\
\quad geospatial & SNAP / Foursquare / GeoNames / OSM / Overture & 7 & {\scriptsize BSD \newline Academic-use only \newline CC BY 4.0 \newline BSD \newline ODbL \newline CDLA-Permissive} & {\scriptsize \href{https://snap.stanford.edu/data/loc-brightkite.html}{snap.stanford.edu} \newline \href{https://sites.google.com/site/yangdingqi/home/foursquare-dataset}{sites.google.com} \newline \href{https://www.geonames.org/}{geonames.org} \newline \href{https://snap.stanford.edu/data/loc-gowalla.html}{snap.stanford.edu} \newline \href{https://www.openstreetmap.org/copyright}{openstreetmap.org} \newline \href{https://docs.overturemaps.org/attribution/}{overturemaps.org}} \\
\quad graph learning & OGB / FB15k / WN18RR / YAGO & 5 & {\scriptsize MS research license \newline ODC-BY \newline WordNet license \newline CC BY 3.0} & {\scriptsize \href{https://www.microsoft.com/en-us/download/details.aspx?id=52312}{microsoft.com} \newline \href{https://github.com/snap-stanford/ogb}{github.com} \newline \href{https://wordnet.princeton.edu/license-and-commercial-use}{wordnet.princeton.edu} \newline \href{https://yago-knowledge.org/}{yago-knowledge.org}} \\
\quad FiveThirtyEight & FiveThirtyEight datasets & 4 & {\scriptsize CC BY 4.0} & {\scriptsize \href{https://github.com/fivethirtyeight/data}{github.com}} \\
\quad ecology & iNaturalist / PBDB / RAM Legacy / GBIF & 4 & {\scriptsize Various \newline CC BY 4.0 \newline CC BY 4.0 \newline Mixed CC0 / CC BY / CC BY-NC} & {\scriptsize \href{https://www.inaturalist.org/pages/help#licensing}{inaturalist.org} \newline \href{https://paleobiodb.org/#/about}{paleobiodb.org} \newline \href{https://www.ramlegacy.org/}{ramlegacy.org} \newline \href{https://www.gbif.org/terms}{gbif.org}} \\
\quad linguistics & Glottolog / OpenSubtitles / WALS & 3 & {\scriptsize CC BY 4.0 \newline Restricted \newline CC BY 4.0} & {\scriptsize \href{https://glottolog.org/}{glottolog.org} \newline \href{https://opus.nlpl.eu/datasets/OpenSubtitles}{opus.nlpl.eu} \newline \href{https://wals.info/}{wals.info}} \\
\quad security & NVD / Exploit-DB / MITRE ATT\&CK & 3 & {\scriptsize Public Domain \newline GPL 2.0 \newline MITRE ATT\&CK terms} & {\scriptsize \href{https://nvd.nist.gov/general/terms-of-use}{nvd.nist.gov} \newline \href{https://www.exploit-db.com/about-exploit-db}{exploit-db.com} \newline \href{https://github.com/mitre-attack/attack-stix-data/blob/master/LICENSE.txt}{github.com}} \\
\quad astronomy & NASA Exoplanet Archive / SDSS & 2 & {\scriptsize Public Domain \newline Free use} & {\scriptsize \href{https://exoplanetarchive.ipac.caltech.edu/docs/acknowledge.html}{ipac.caltech.edu} \newline \href{https://www.sdss.org/collaboration/citing-sdss/}{sdss.org}} \\
\quad forecasting & Good Judgment Project & 1 & {\scriptsize CC0} & {\scriptsize \href{https://dataverse.harvard.edu/dataset.xhtml?persistentId=doi:10.7910/DVN/BPCDH5}{dataverse.harvard.edu}} \\
\quad hydrology & USGS Water Data & 1 & {\scriptsize Public Domain} & {\scriptsize \href{https://waterdata.usgs.gov/provisional-data-statement/}{waterdata.usgs.gov}} \\
\quad environment & EPA AQS & 1 & {\scriptsize Public Domain} & {\scriptsize \href{https://www.epa.gov/aqs}{epa.gov}} \\
\quad network infra & RIPE / RouteViews & 1 & {\scriptsize Open data} & {\scriptsize \href{https://www.ripe.net/about-us/legal/ripe-data-repository-terms-and-conditions/}{ripe.net}} \\
\quad ocean & NOAA NDBC & 1 & {\scriptsize Public Domain} & {\scriptsize \href{https://www.ndbc.noaa.gov/}{ndbc.noaa.gov}} \\
\quad storm & NOAA Storm Events & 1 & {\scriptsize Public Domain} & {\scriptsize \href{https://www.ncei.noaa.gov/access/storm-events-database/}{ncei.noaa.gov}} \\
\quad threat intel & URLhaus (abuse.ch) & 1 & {\scriptsize abuse.ch ToS} & {\scriptsize \href{https://urlhaus.abuse.ch/api/}{urlhaus.abuse.ch}} \\
\end{longtable}
}

\subsection{Phase 2: Standardization}
\label{sec:appendix-standardise}

\noindent\textbf{Creating a common representation.}
After collection, each database is standardized into a uniform relational format. The purpose of this stage is to make heterogeneous sources loadable and comparable, without altering their semantic content. Each fix is implemented as an idempotent detector--transformer pass over the collected parquet files: a pass first checks whether the target invariant already holds, and only rewrites the data when needed.

\noindent\textbf{Recovering relational structure.}
Many public datasets contain useful structure that is implicit rather than declared. We therefore recover primary keys, foreign keys, time columns, boolean fields, and numeric columns when they can be inferred reliably. These repairs are conservative: candidate keys must satisfy uniqueness checks, foreign keys must pass containment checks, and identifier-like fields such as zip codes, ISBNs, phone numbers, and barcodes are not coerced into numeric values.

\noindent\textbf{Normalizing common type issues.}
The standardization pass resolves common inconsistencies across sources, such as booleans stored as strings, dates stored in mixed textual formats, year columns stored as integers, and numeric values stored as strings. Ambiguous cases are left unchanged or flagged rather than silently modified. This keeps the downstream interface simple while avoiding brittle source-specific assumptions.

\noindent\textbf{Controlling scale.}
For databases that are too large for repeated downstream processing, we apply deterministic entity-axis subsampling: entities are sampled with a fixed seed, all associated activity rows are retained, and keys are reindexed to preserve relational consistency. 

\noindent\textbf{Preserving auditability.}
All standardization fixes are performed by mechanical scripts, and each transformation is governed by explicit rules. As a result, downstream task generation and evaluation operate on a consistent representation, while every repair remains reproducible and inspectable.



\subsection{Phase 3: Task proposal and validation}
\label{sec:appendix-propose}

\noindent\textbf{Proposing tasks from schema context.}
For each eligible database, the proposer receives a structured summary of the database, including its domain, tables, row counts, key relationships, time columns, column types, basic column statistics, and a small number of example rows from activity tables. This context is wrapped in a fixed prompt that specifies the allowed task templates, the required JSON format, and the validation rules. The proposer also receives the names of previously generated tasks for that database, which helps reduce duplicates across passes.

\noindent\textbf{Constraining the proposal space.}
Task proposals are restricted to a registered set of templates, listed in \Cref{tab:templates}. This keeps the generated tasks interpretable and mechanically checkable, while still allowing broad coverage across domains. Proposals that refer to unknown templates are rejected before further processing.

\noindent\textbf{Validating mechanically.}
The LLM proposes candidate tasks, but acceptance is decided by a deterministic validator. The validator checks that referenced tables and columns exist, entity columns are valid foreign keys, numeric fields are used where required, generated target names do not collide with existing columns, task names follow the required format, and binary thresholds avoid trivially imbalanced labels. Rejected proposals are logged with their failure reason and are not written to the released task set.

\noindent\textbf{Improving diversity across passes.}
Each database is processed through multiple proposal passes. Early passes tend to recover natural or obvious tasks for the schema, while later passes explicitly target under-represented templates and harder prediction settings, such as rare positives, cold-start behavior, heavy-tailed regression, ranking, and derived-ratio tasks. This multi-pass strategy increases task diversity while preserving the same mechanical validation standard for every accepted proposal.

\noindent\textbf{Keeping generation auditable.}
The proposal stage uses an LLM to suggest candidate tasks, but the final task set is determined by explicit templates and deterministic checks. This separation allows the pipeline to benefit from the LLM's ability to identify plausible prediction problems from heterogeneous schemas, while keeping task inclusion reproducible, inspectable, and independent of unverified free-form judgments.



\begin{figure}[!htbp]
\centering
\begin{lstlisting}[basicstyle=\scriptsize\ttfamily,
                   frame=single,framerule=0.4pt,framesep=4pt,
                   backgroundcolor=\color{black!3},
                   xleftmargin=4pt,xrightmargin=4pt,
                   aboveskip=2pt,belowskip=2pt]
Database: civic/nyc_311
Total: 4 tables, 501223 rows
Entity tables: ['agencies', 'locations', 'complaint_types']

Schema:
  agencies (15 rows pkey=agency_id):
    columns: agency(string), agency_name(string), agency_id(Int64, range=[0..14])
  locations (270 rows pkey=location_id):
    columns: borough(string), incident_zip(Int64, range=[8854..92108]), location_id(Int64, range=[0..269])
  complaints (500000 rows pkey=unique_key time_col=created_date
              fks={'complaint_type_id': 'complaint_types', 'agency_id': 'agencies', 'location_id': 'locations'}):
    columns: unique_key(Int64), complaint_type_id(Int64), agency_id(Int64), location_id(Int64),
             created_date(datetime64[us, UTC]), closed_date(datetime64[us, UTC]),
             resolution_hours(Float64, mean=49.6, range=[0..1105]), status(string),
             community_board(string), open_data_channel_type(string)
  complaint_types (938 rows pkey=complaint_type_id):
    columns: complaint_type(string), descriptor(string), complaint_type_id(Int64, range=[0..937])

Activity tables (have time_col + FK to entity):
  complaints: time_col=created_date, entity_fks={'complaint_type_id': 'complaint_types',
              'agency_id': 'agencies', 'location_id': 'locations'}, 500000 rows
    sample: [{'unique_key': 0, 'agency_id': 0, 'created_date': '2026-04-14 02:33:29+00:00',
              'resolution_hours': None, 'status': 'Open', 'community_board': '15 BROOKLYN', ...}, ...]

Existing tasks for this DB (do NOT duplicate these):
  predict-agency-avg-resolution-hours, predict-location-max-resolution-hours,
  predict-agency-slow-resolution, predict-complaint-type-active,
  predict-agency-resolution-trend, ... and 24 more

TASK: Propose up to 15 high-quality forecasting tasks that a domain expert would find meaningful.
Each task predicts a future value for an entity over a time window.

PRIORITY: Our existing task suite is heavy on aggregate_numeric, threshold_binary, existence_binary,
count_events, churn_activity. Use at least 40% of your proposals from these under-represented
patterns (where applicable): time_until_next (REG), streak_length (REG), rank_change (CLF),
rank_regression (REG), peer_outperform (CLF), new_category_adoption (CLF), will_repeat_last (CLF),
trend_reversal (CLF), volatility_numeric (REG), concentration (REG), derived_ratio (REG).

If the schema supports a `group_col` (race/week/country) -> propose rank_change + rank_regression.
If there's a natural `peer_col` (category/country/segment) -> propose peer_outperform. If there's
a `category_col` (product/type/status) -> propose new_category_adoption + will_repeat_last.

ALLOWED PATTERNS:
  - aggregate_numeric (REG): mean/sum/max of numeric col next window. Needs `source_col`,
    `aggregation` in [mean, sum, max, min].
  - threshold_binary (CLF): aggregate exceeds threshold. Needs `source_col`, `aggregation`,
    `threshold`, `compare` in [gt, gte, lt, lte].
  - existence_binary (CLF): entity appears in event_table next window. Needs `event_table`,
    `event_entity_col`, `event_time_col`.
  ... [more templates omitted]
  - heavy_tail_regression (REG, HARD TIER): log1p(agg(source_col)) per entity next window.
    Entity pool = ALL historical entities. Matches rel-amazon/item-ltv, rel-hm/item-sales.

HARD REQUIREMENTS (validator rejects violations):
  1. entity_col MUST be a foreign key in the activity_table.
  2. source_col / numerator_col / denominator_col MUST be numeric columns in the activity_table.
  3. target_col must be short and not collide with an existing column.
  4. task_name must be kebab-case starting with "predict-".
  5. For derived_ratio: pick MEANINGFUL pairs (batting average, completion rate, margins).
  6. For threshold_binary: thresholds near the median of the column.
  7. Prefer real-world interpretability over mechanical column selection.

Return ONLY a JSON array of task objects:
[ { "task_name": "predict-...", "task_type": "REGRESSION" | "BINARY_CLASSIFICATION",
    "pattern": "aggregate_numeric" | ..., "entity_table": "...", "entity_col": "...",
    "activity_table": "...", "time_col": "...", "target_col": "...", "target_description": "...",
    ... template-specific fields ..., "skip_cols": [], "tables_used": [...],
    "timedelta_multiplier": 1 } ]
\end{lstlisting}
\caption{Full proposer prompt for \texttt{civic/nyc\_311} (one
representative database). The schema, FK map, sample rows, and
existing-task list are populated per database; the
\texttt{ALLOWED~PATTERNS}, \texttt{HARD~REQUIREMENTS}, and JSON
contract blocks are constant across databases (the
\texttt{ALLOWED~PATTERNS} block in particular is abbreviated here---the full version enumerates all 25 templates with their
required fields).}
\label{fig:example-prompt}
\end{figure}




\begin{figure}[!htbp]
\centering
\begin{lstlisting}[basicstyle=\scriptsize\ttfamily,
                   frame=single,framerule=0.4pt,framesep=4pt,
                   backgroundcolor=\color{black!3},
                   xleftmargin=4pt,xrightmargin=4pt,
                   aboveskip=2pt,belowskip=2pt]
{
  "task_name": "predict-agency-avg-resolution-hours",
  "task_type": "REGRESSION",
  "pattern": "aggregate_numeric",
  "entity_table": "agencies",
  "entity_col": "agency_id",
  "activity_table": "complaints",
  "time_col": "created_date",
  "target_col": "avg_resolution_hours",
  "target_description": "Predict the mean resolution time (hours) for complaints handled by an agency in the next window.",
  "source_col": "resolution_hours",
  "aggregation": "mean",
  "tables_used": ["agencies", "complaints"],
  "timedelta_multiplier": 1
}
\end{lstlisting}
\caption{First accepted proposal for \texttt{civic/nyc\_311},
returned by the Claude Opus proposer in response to the prompt of
\Cref{fig:example-prompt}. The proposal passes the mechanical
validator (the \texttt{entity\_col} is a declared FK in the
\texttt{activity\_table}, \texttt{source\_col} is a numeric column
of the same table, \texttt{target\_col} does not collide with any
existing column, and the \texttt{pattern} is registered) and is
forwarded to the generator for instantiation.}
\label{fig:example-proposal}
\end{figure}


\begin{table}[!htbp]
\centering
\scriptsize
\setlength{\tabcolsep}{5pt}
\renewcommand{\arraystretch}{1.02}
\caption{The 25 task-template families instantiated by the corpus, organized under italic signal-type headers. Each row gives the prediction target as the model sees it, paired with one concrete domain example. \textsc{r}: regression; \textsc{c}: binary classification.}
\label{tab:templates}
\begin{tabular}{@{}l c p{0.66\linewidth}@{}}
\toprule
\textbf{Template} & \textbf{Type} & \textbf{Prediction target (example)} \\
\midrule
\multicolumn{3}{@{}l}{\textit{Aggregation---volume / intensity of activity}} \\
\quad \texttt{aggregate\_numeric} & \textsc{r} & mean / sum of a numeric column over the next window \emph{(e.g.\ a player's home runs next season)} \\
\quad \texttt{count\_events} & \textsc{r} & number of events for the entity in the next window \emph{(e.g.\ orders per customer next month)} \\
\quad \texttt{distinct\_count} & \textsc{r} & distinct values of a column the entity touches \emph{(e.g.\ distinct co-authors next year)} \\
\quad \texttt{concentration} & \textsc{r} & entity's share of total activity \emph{(e.g.\ seller's share of category sales)} \\
\quad \texttt{derived\_ratio} & \textsc{r} & ratio of two numeric columns over the next window \emph{(e.g.\ new-to-returning customer ratio)} \\
\addlinespace[2pt]
\multicolumn{3}{@{}l}{\textit{Threshold / existence---continuous signal at a decision boundary}} \\
\quad \texttt{threshold\_binary} & \textsc{c} & whether an aggregate exceeds a threshold \emph{(e.g.\ revenue $>$\$1M next quarter)} \\
\quad \texttt{existence\_binary} & \textsc{c} & whether the entity appears in the event table next window \emph{(e.g.\ user posts at least once)} \\
\quad \texttt{top\_k\_binary} & \textsc{c} & whether the entity is in the top $K\%$ on a metric \emph{(e.g.\ artist enters top 1\% of plays)} \\
\addlinespace[2pt]
\multicolumn{3}{@{}l}{\textit{Trend / change---direction and magnitude of change}} \\
\quad \texttt{trend\_binary} & \textsc{c} & whether a metric rises relative to the current window \emph{(e.g.\ weekly volume rises)} \\
\quad \texttt{trend\_reversal} & \textsc{c} & whether the direction of change flips between windows \emph{(e.g.\ sales reverse from rising to falling)} \\
\quad \texttt{change\_numeric} & \textsc{r} & numeric delta from current to next window \emph{(e.g.\ change in monthly energy consumption)} \\
\quad \texttt{volatility\_numeric} & \textsc{r} & standard deviation of a metric in the next window \emph{(e.g.\ next-week price volatility)} \\
\addlinespace[2pt]
\multicolumn{3}{@{}l}{\textit{Behavior / churn---entity lifecycle}} \\
\quad \texttt{churn\_activity} & \textsc{c} & entity has \emph{no} activity in the next window \emph{(e.g.\ subscriber cancels next month)} \\
\quad \texttt{return\_binary} & \textsc{c} & dormant entity reactivates next window \emph{(e.g.\ dormant Stack Exchange user posts again)} \\
\quad \texttt{time\_until\_next} & \textsc{r} & days until the entity's next event \emph{(e.g.\ days until a player's next ATP match)} \\
\quad \texttt{will\_repeat\_last} & \textsc{c} & next event's category equals the most-recent one \emph{(e.g.\ customer re-buys the same product)} \\
\quad \texttt{first\_event\_binary} & \textsc{c} & entity experiences a first occurrence next window \emph{(e.g.\ first purchase by a new account)} \\
\addlinespace[2pt]
\multicolumn{3}{@{}l}{\textit{Ranking---relative position within a peer group}} \\
\quad \texttt{rank\_change} & \textsc{c} & entity's rank within its group improves by $\geq K$ \emph{(e.g.\ driver climbs $\geq$5 places)} \\
\quad \texttt{rank\_regression} & \textsc{r} & numeric rank within the group ($1=$ best) \emph{(e.g.\ predict the driver's next-race finish position)} \\
\quad \texttt{peer\_outperform} & \textsc{c} & entity beats the mean of its peer group \emph{(e.g.\ artist beats peer-group mean plays)} \\
\addlinespace[2pt]
\multicolumn{3}{@{}l}{\textit{Adoption / novelty---first-time events}} \\
\quad \texttt{new\_category\_adoption} & \textsc{c} & entity touches a category value it never has before \emph{(e.g.\ first purchase from a new product category)} \\
\addlinespace[2pt]
\multicolumn{3}{@{}l}{\textit{Streak---resilience as a single metric}} \\
\quad \texttt{streak\_length} & \textsc{r} & longest run of consecutive active sub-periods \emph{(e.g.\ consecutive seasons a player remains active)} \\
\addlinespace[2pt]
\multicolumn{3}{@{}l}{\textit{Tail-event (hard-tier)---rare and extreme regimes}} \\
\quad \texttt{cold\_start\_binary} & \textsc{c} & newly-appearing entity has repeat activity in its first window \emph{(e.g.\ new user posts within 24h of joining)} \\
\quad \texttt{rare\_positive\_binary} & \textsc{c} & rare event ($\leq\!10\%$ positive rate) occurs for the entity \emph{(e.g.\ rare adverse-quality hospital event)} \\
\quad \texttt{heavy\_tail\_regression} & \textsc{r} & $\log(1+x)$ of an aggregate $x$; preserves heavy upper tail \emph{(e.g.\ $\log(1+{}$next-month spend$)$, large spenders preserved)} \\
\bottomrule
\end{tabular}
\end{table}

\subsection{Phase 4: Baselines and quality filtering}
\label{sec:appendix-baseline}

\noindent\textbf{Measuring basic task signal.}
After task generation, each task is evaluated with a small set of baselines. We first compute two cheap historical baselines: an entity-mean predictor, which uses each entity's average training label, and a last-value predictor, which uses the entity's most recent observed label when available. These baselines are inexpensive and provide a simple check for whether the task contains recoverable temporal or entity-level signal.

\noindent\textbf{Training a lightweight ML baseline.}
We also train a fixed XGBoost baseline for each task. For classification tasks we use \texttt{XGBClassifier}, and for regression tasks we use \texttt{XGBRegressor}, with \texttt{n\_estimators=100}, \texttt{max\_depth=6}, \texttt{learning\_rate=0.1}, \texttt{tree\_method="hist"}, and \texttt{n\_jobs=2}. Features are simple window aggregations over the activity table: for the 20 most populous numeric columns, we compute \texttt{count}, \texttt{mean}, \texttt{sum}, \texttt{max}, and \texttt{min} within a lookback window of length $5\times\texttt{timedelta}$ ending at the prediction time. To keep the pass scalable, training is capped at $5{,}000$ rows per task and evaluation at $2{,}000$ rows per task. This baseline is not intended to be a strong relational model and is only intended to provide a consistent and reproducible reference point for filtering tasks across domains.

\noindent\textbf{Filtering and difficulty assignment.}
The quality filter removes tasks that are unlikely to be useful for downstream evaluation, such as tasks that are nearly trivial, exhibit severe distribution drift, or fail during baseline fitting. Tasks with non-trivial baseline signal are retained, and we also retain non-trivial tasks for which XGBoost does not recover signal, since stronger relational models may still exploit structure missed by the lightweight baseline. Retained tasks are assigned a coarse difficulty level from their XGBoost score: easy, medium, or hard.

\noindent\textbf{Preparing the release format.}
After loadability validation, each task is converted into the release format. The train, validation, and test splits are concatenated in chronological order into a single time-ordered training file, while preserving the schema metadata required by downstream loaders. 




\subsection{Asset licenses, Citations, and Compliance}
\label{sec:appendix-licenses}

\noindent\textbf{Compliance practice.} Every contributing source is
either (a) a bulk archive that the upstream provider publishes for
download, or (b) a public API that the provider documents for
programmatic access. The terms of these sources differ: most are public
domain or under permissive or share-alike licenses, but some restrict
use to research, to non-commercial purposes, or to the provider's own
terms of service, and these restrictions carry over to the
corresponding databases in \theJoin{}. We do not crawl
arbitrary websites, do not bypass paywalls, do not ignore
\texttt{robots.txt}, and do not scrape sources whose terms of
service forbid automation.

\noindent\textbf{Released-corpus license.} The corpus assembly (schemas, standardization and generated tasks) is released under \textbf{CC-BY-SA 4.0}, but each database remains subject to the terms of its upstream source, which take precedence where they are more restrictive. Per-database metadata
records (i)~the upstream source identifier and URL and (ii)~the upstream
license string verbatim. Restrictions inherited from the sources include
non-commercial use only (e.g.\ ATP Tennis, Apple Podcasts, OpenSanctions, IMDb, Racing-Reference),
no derivatives (RecSys Challenge 2015),
research or academic use only (e.g.\ the CTU repository, MovieLens, Foursquare, and the McAuley Lab datasets),
and provider terms of service (e.g.\ Manifold, which prohibits commercial AI/ML training, Polymarket, StatsBomb, CoinGecko, URLhaus and OpenSubtitles).
Users of the corpus, and of models trained on it, must check and honor these per-source restrictions, particularly for commercial use.


\noindent\textbf{PII.} The corpus contains no personally-identifying
information beyond what each upstream source has already published
under its own license. Where source data uses pseudonymized
identifiers (e.g.\ Stack Exchange user IDs, Lahman player IDs), we
preserve those identifiers verbatim and do not attempt to de-anonymize or join across sources.

%
\subsection{Case Studies}
\label{app:case-studies}

\setlength{\emergencystretch}{2.5em}

\usetikzlibrary{arrows.meta}
\definecolor{cEnt}{HTML}{1F4E79}
\definecolor{cAct}{HTML}{B45309}
\definecolor{cTask}{HTML}{166534}
\definecolor{cPl}{HTML}{475569}
\tikzset{
  erTbl/.style     ={draw=cPl!50,  fill=cPl!5,   rounded corners=2.5pt, inner sep=4pt, line width=.5pt},
  erTask/.style    ={draw=cTask,   fill=cTask!8, rounded corners=2.5pt, inner sep=4pt, line width=1pt, dashed},
  erEdge/.style    ={-{Stealth[length=2.2mm,width=1.6mm]}, draw=cPl,     line width=.75pt},
  erEdgeTask/.style={-{Stealth[length=2.2mm,width=1.6mm]}, draw=cTask,   line width=.9pt, dashed},
}

To illustrate what a forecasting task in \theJoin{} represents and the diversity
of the corpus, we present three case studies from distinct real-world
domains. 
In every case, the prediction target is
a quantity monitored in the corresponding domain and instantiates a different label family among the twenty-five task patterns defined in the corpus.
\paragraph{Forecasting an account's peak balance (\texttt{join-berka})}

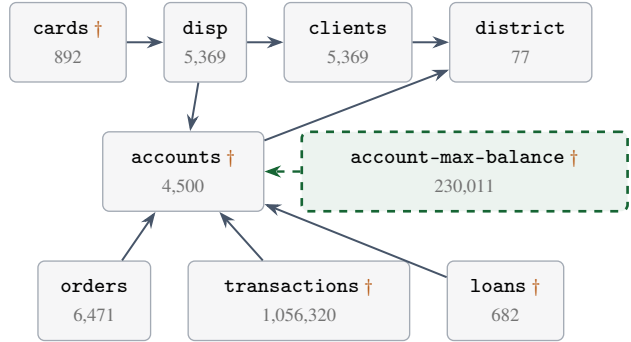
\begin{figure}[htbp]
\centering
\begin{tikzpicture}
  \node[erTbl,minimum width=21.3mm,minimum height=7.6mm] (accounts) at (27.67mm,17.10mm) {\begin{tabular}{@{}c@{}}{\ttfamily\footnotesize\bfseries accounts}\,{\scriptsize\color{cAct}$\dagger$}\\{\scriptsize\color{black!55}4,500}\end{tabular}};
  \node[erTbl,minimum width=18.8mm,minimum height=7.6mm] (district) at (72.32mm,34.20mm) {\begin{tabular}{@{}c@{}}{\ttfamily\footnotesize\bfseries district}\\{\scriptsize\color{black!55}77}\end{tabular}};
  \node[erTbl,minimum width=10.9mm,minimum height=7.6mm] (disp) at (30.58mm,34.20mm) {\begin{tabular}{@{}c@{}}{\ttfamily\footnotesize\bfseries disp}\\{\scriptsize\color{black!55}5,369}\end{tabular}};
  \node[erTbl,minimum width=14.9mm,minimum height=7.6mm] (orders) at (15.90mm,0.00mm) {\begin{tabular}{@{}c@{}}{\ttfamily\footnotesize\bfseries orders}\\{\scriptsize\color{black!55}6,471}\end{tabular}};
  \node[erTbl,minimum width=29.3mm,minimum height=7.6mm] (transactions) at (42.97mm,0.00mm) {\begin{tabular}{@{}c@{}}{\ttfamily\footnotesize\bfseries transactions}\,{\scriptsize\color{cAct}$\dagger$}\\{\scriptsize\color{black!55}1,056,320}\end{tabular}};
  \node[erTbl,minimum width=16.9mm,minimum height=7.6mm] (clients) at (49.47mm,34.20mm) {\begin{tabular}{@{}c@{}}{\ttfamily\footnotesize\bfseries clients}\\{\scriptsize\color{black!55}5,369}\end{tabular}};
  \node[erTbl,minimum width=15.4mm,minimum height=7.6mm] (loans) at (70.30mm,0.00mm) {\begin{tabular}{@{}c@{}}{\ttfamily\footnotesize\bfseries loans}\,{\scriptsize\color{cAct}$\dagger$}\\{\scriptsize\color{black!55}682}\end{tabular}};
  \node[erTbl,minimum width=15.4mm,minimum height=7.6mm] (cards) at (12.42mm,34.20mm) {\begin{tabular}{@{}c@{}}{\ttfamily\footnotesize\bfseries cards}\,{\scriptsize\color{cAct}$\dagger$}\\{\scriptsize\color{black!55}892}\end{tabular}};
  \node[erTask,minimum width=43.1mm,minimum height=7.6mm] (task) at (64.90mm,17.10mm) {\begin{tabular}{@{}c@{}}{\ttfamily\footnotesize\bfseries account-max-balance}\,{\scriptsize\color{cAct}$\dagger$}\\{\scriptsize\color{black!55}230,011}\end{tabular}};
  \draw[erEdge] (accounts) -- (district);
  \draw[erEdge] (disp) -- (clients);
  \draw[erEdge] (disp) -- (accounts);
  \draw[erEdge] (orders) -- (accounts);
  \draw[erEdge] (transactions) -- (accounts);
  \draw[erEdge] (clients) -- (district);
  \draw[erEdge] (loans) -- (accounts);
  \draw[erEdge] (cards) -- (disp);
  \draw[erEdgeTask] (task) -- (accounts);
\end{tikzpicture}
\caption{Schema of \texttt{join-berka}: every table, its row count, and the foreign key links. Arrows point from a foreign key to the table it references; $\dagger$ marks a table
carrying a time column. The dashed box is the task table derived
from the database.}
\label{fig:schema-berka}
\end{figure}

Our first example predicts a continuous quantity from a transaction stream.
The Berka dataset from the PKDD'99 Challenge is a complete relational snapshot
of a Czech retail bank, including its accounts, the clients who hold them, the
permanent orders and loans they carry, and a central ledger of just over a
million individual transactions, each
stamped with the resulting running balance.  Its eight tables form a multi-hop
schema (\Cref{fig:schema-berka}), where \texttt{accounts} are linked to their \texttt{clients} through \texttt{disp}, and both
\texttt{accounts} and \texttt{clients} are linked to the \texttt{district} in which they are registered. As peak balance is a quantity that retail banks monitor directly, we create the \texttt{account-max-balance} regression task that, for each
account and seed time, predicts the highest balance reached during the following
window of roughly $17.5$ days. 

\paragraph{Forecasting new-user retention (\texttt{join-brightkite})}

A cold-start
task evaluates an entity at its first appearance. This setting is of independent interest because it
isolates a model's ability to generalize from relational context rather than
from the entity's own past. To showcase such a task in~\theJoin{}, we consider
Brightkite, a location-based social network in which users check in at venues
and form friendships. 
The \texttt{newuser-will-return} task considers each user in the window of their
first-ever check-in and predicts whether they will check in again within the next
two windows, each roughly $6.7$ days. Because the user's own label
history is unavailable, a model cannot exploit autocorrelation and must instead
generalize from side information, including the venues visited and the user's
position in the friendship graph. 

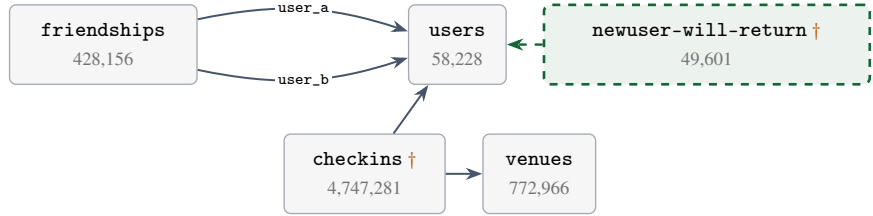
\begin{figure}[htbp]
\centering
\begin{tikzpicture}
  \node[erTbl,minimum width=21.3mm,minimum height=7.6mm] (checkins) at (46.96mm,0.00mm) {\begin{tabular}{@{}c@{}}{\ttfamily\footnotesize\bfseries checkins}\,{\scriptsize\color{cAct}$\dagger$}\\{\scriptsize\color{black!55}4,747,281}\end{tabular}};
  \node[erTbl,minimum width=12.9mm,minimum height=7.6mm] (users) at (59.23mm,17.10mm) {\begin{tabular}{@{}c@{}}{\ttfamily\footnotesize\bfseries users}\\{\scriptsize\color{black!55}58,228}\end{tabular}};
  \node[erTbl,minimum width=24.8mm,minimum height=7.6mm] (friendships) at (12.39mm,17.10mm) {\begin{tabular}{@{}c@{}}{\ttfamily\footnotesize\bfseries friendships}\\{\scriptsize\color{black!55}428,156}\end{tabular}};
  \node[erTbl,minimum width=14.9mm,minimum height=7.6mm] (venues) at (70.07mm,0.00mm) {\begin{tabular}{@{}c@{}}{\ttfamily\footnotesize\bfseries venues}\\{\scriptsize\color{black!55}772,966}\end{tabular}};
  \node[erTask,minimum width=43.1mm,minimum height=7.6mm] (task) at (92.24mm,17.10mm) {\begin{tabular}{@{}c@{}}{\ttfamily\footnotesize\bfseries newuser-will-return}\,{\scriptsize\color{cAct}$\dagger$}\\{\scriptsize\color{black!55}49,601}\end{tabular}};
  \draw[erEdge] (checkins) -- (users);
  \draw[erEdge] (checkins) -- (venues);
  \draw[erEdge] (friendships) to[bend left=15] node[midway,font=\tiny\ttfamily,fill=white,inner sep=1pt]{user\_a} (users);
  \draw[erEdge] (friendships) to[bend left=-15] node[midway,font=\tiny\ttfamily,fill=white,inner sep=1pt]{user\_b} (users);
  \draw[erEdgeTask] (task) -- (users);
\end{tikzpicture}
\caption{Schema of \texttt{join-brightkite}. Here \texttt{friendships} and \texttt{users} tables are joined by more than one foreign key, so the edges are
drawn separately and labeled with the referencing column.}
\label{fig:schema-brightkite}
\end{figure}

\paragraph{Forecasting who beats their cohort (\texttt{join-oulad})}

 We
consider the Open University Learning Analytics Dataset (OULAD), a relational
record of a large online university that links each student to their
virtual-learning-environment clickstream and assessment scores. The \texttt{student-score-beats-cohort} task predicts whether a student's mean
assessment score in the next window (of roughly $7.2$ days) will exceed the mean
score of their presentation cohort over the same window. The comparison
threshold is therefore the cohort average rather than a fixed
constant, and must be computed before the label can be assigned. Because the label is
defined relative to a peer aggregate that changes over time, a model cannot
score each student in isolation. Instead, it must compute the cohort context at each step
and track the student's trajectory across windows.

\begin{figure}[htbp]
\centering
\begin{tikzpicture}
  \node[erTbl,minimum width=45.1mm,minimum height=7.6mm] (studentregistration) at (111.38mm,0.00mm) {\begin{tabular}{@{}c@{}}{\ttfamily\footnotesize\bfseries student\_registration}\,{\scriptsize\color{cAct}$\dagger$}\\{\scriptsize\color{black!55}32,593}\end{tabular}};
  \node[erTbl,minimum width=16.9mm,minimum height=7.6mm] (courses) at (72.17mm,34.20mm) {\begin{tabular}{@{}c@{}}{\ttfamily\footnotesize\bfseries courses}\\{\scriptsize\color{black!55}22}\end{tabular}};
  \node[erTbl,minimum width=28.7mm,minimum height=7.6mm] (vleresources) at (14.37mm,17.10mm) {\begin{tabular}{@{}c@{}}{\ttfamily\footnotesize\bfseries vle\_resources}\\{\scriptsize\color{black!55}6,364}\end{tabular}};
  \node[erTbl,minimum width=18.8mm,minimum height=7.6mm] (students) at (72.94mm,17.10mm) {\begin{tabular}{@{}c@{}}{\ttfamily\footnotesize\bfseries students}\\{\scriptsize\color{black!55}32,593}\end{tabular}};
  \node[erTbl,minimum width=27.3mm,minimum height=7.6mm] (studentvle) at (24.05mm,0.00mm) {\begin{tabular}{@{}c@{}}{\ttfamily\footnotesize\bfseries student\_vle}\,{\scriptsize\color{cAct}$\dagger$}\\{\scriptsize\color{black!55}10,655,280}\end{tabular}};
  \node[erTbl,minimum width=24.8mm,minimum height=7.6mm] (assessments) at (46.13mm,17.10mm) {\begin{tabular}{@{}c@{}}{\ttfamily\footnotesize\bfseries assessments}\\{\scriptsize\color{black!55}206}\end{tabular}};
  \node[erTbl,minimum width=41.1mm,minimum height=7.6mm] (studentassessment) at (63.26mm,0.00mm) {\begin{tabular}{@{}c@{}}{\ttfamily\footnotesize\bfseries student\_assessment}\,{\scriptsize\color{cAct}$\dagger$}\\{\scriptsize\color{black!55}173,912}\end{tabular}};
  \node[erTask,minimum width=57.0mm,minimum height=7.6mm] (task) at (110.00mm,34.20mm) {\begin{tabular}{@{}c@{}}{\ttfamily\footnotesize\bfseries student-score-beats-cohort}\,{\scriptsize\color{cAct}$\dagger$}\\{\scriptsize\color{black!55}43,872}\end{tabular}};
  \draw[erEdge] (studentregistration) -- (students);
  \draw[erEdge] (vleresources) -- (courses);
  \draw[erEdge] (students) -- (courses);
  \draw[erEdge] (studentvle) -- (vleresources);
  \draw[erEdge] (studentvle) -- (students);
  \draw[erEdge] (assessments) -- (courses);
  \draw[erEdge] (studentassessment) -- (assessments);
  \draw[erEdge] (studentassessment) -- (students);
  \draw[erEdgeTask] (task) -- (students);
\end{tikzpicture}
\caption{Schema of \texttt{join-oulad}.}
\label{fig:schema-oulad}
\end{figure}

\subsection{Manual Audit of Generated Tasks}
\label{app:task-audit}

The deterministic checks of \Cref{sec:tasks} ensure that tasks are valid against the schema, but not that they are meaningful. We therefore manually reviewed the LLM task descriptions and label-generation SQL, and found the following.

\textbf{Targets are semantically named quantities.} In \texttt{join-berka}, tasks ignore the meaningless \texttt{a2}--\texttt{a16} columns of \texttt{district} and instead predict \texttt{max\_balance}, \texttt{goes\_negative}, \texttt{days\_to\_next\_txn}, etc.

\textbf{Label SQL uses domain concepts, not arbitrary aggregations.} In \texttt{join-oulad}, \texttt{student-score-beats-cohort} compares a student to their cohort's contemporaneous mean. In \texttt{join-citi-bike}, \texttt{station-time-until-next-trip} predicts station idle time, which operators need for inventory planning.

\textbf{Tasks match real practitioner decisions.} \texttt{account-max-balance} (\texttt{join-berka}) mirrors overdraft-exposure and credit-limit review; \texttt{newuser-will-return} (\texttt{join-brightkite}) is a standard onboarding-retention metric; \texttt{movie-new-tag-adoption} (\texttt{join-movielens}) tracks metadata coverage.

A small number of tasks (about $100$) did aggregate non-semantic columns, e.g., \texttt{student-feature1-} \texttt{to-feature0-ratio} in \texttt{join-act\_mooc}.

\section{RelBench Forecasting Tasks}
\label{sec:appendix-relbench-v1-tasks}


\noindent\textbf{Binary classification tasks.}
The binary classification tasks in RelBench ask whether a future event or condition will occur for a given entity. They cover churn, future engagement,
ad-click behavior, race outcomes, clinical outcomes, and event attendance
behavior. See \Cref{tab:relbench-v1-binary} for task descriptions. 

\begin{table}[t]
\centering
\scriptsize
\caption{Binary classification forecasting tasks in RelBench.}
\label{tab:relbench-v1-binary}
\begin{tabular}{@{}lll p{0.52\linewidth}@{}}
\toprule
\textbf{Dataset} & \textbf{Task} & \textbf{Entity} & \textbf{Prediction target} \\
\midrule
\texttt{rel-amazon} & \texttt{user-churn} & User &
Whether the user does not review any product in the next 3 months. \\
\texttt{rel-amazon} & \texttt{item-churn} & Item &
Whether the item receives no reviews in the next 3 months. \\
\texttt{rel-avito} & \texttt{user-visits} & User &
Whether the user visits more than one ad in the next 4 days. \\
\texttt{rel-avito} & \texttt{user-clicks} & User &
Whether the user clicks more than one ad in the next 4 days. \\
\texttt{rel-f1} & \texttt{driver-dnf} & Driver &
Whether the driver does not finish a race in the next month. \\
\texttt{rel-f1} & \texttt{driver-top3} & Driver &
Whether the driver qualifies in the top 3 for a race in the next month. \\
\texttt{rel-hm} & \texttt{user-churn} & User &
Whether the customer makes no transactions in the next week. \\
\texttt{rel-stack} & \texttt{user-engagement} & User &
Whether the user makes any votes, posts, or comments in the next 3 months. \\
\texttt{rel-stack} & \texttt{user-badge} & User &
Whether the user receives a new badge in the next 3 months. \\
\texttt{rel-trial} & \texttt{study-outcome} & Study &
Whether the trial achieves its primary outcome in the next year. \\
\texttt{rel-event} & \texttt{user-repeat} & User &
Whether a previously active user attends another event in the next 7 days. \\
\texttt{rel-event} & \texttt{user-ignore} & User &
Whether the user ignores more than two event invitations in the next 7 days. \\
\bottomrule
\end{tabular}
\end{table}

\noindent\textbf{Regression tasks.}
The regression tasks ask for a continuous or count-valued future outcome.
They cover lifetime value, item sales, click-through rate, future votes,
race position, adverse-event counts, site success, and event attendance. See \Cref{tab:relbench-v1-regression} for task descriptions.

\begin{table}[t]
\centering
\scriptsize
\caption{Regression forecasting tasks in RelBench.}
\label{tab:relbench-v1-regression}
\begin{tabular}{@{}lll p{0.52\linewidth}@{}}
\toprule
\textbf{Dataset} & \textbf{Task} & \textbf{Entity} & \textbf{Prediction target} \\
\midrule
\texttt{rel-amazon} & \texttt{user-ltv} & User &
Value of the products the user buys and reviews in the next 3 months. \\
\texttt{rel-amazon} & \texttt{item-ltv} & Item &
Value of the purchases and reviews the item receives in the next 3 months. \\
\texttt{rel-avito} & \texttt{ad-ctr} & Ad &
Click-through rate for the ad over the next 4 days. \\
\texttt{rel-f1} & \texttt{driver-position} & Driver &
Average finishing position of the driver across races in the next 2 months. \\
\texttt{rel-hm} & \texttt{item-sales} & Item &
Total sales for the article in the next week. \\
\texttt{rel-stack} & \texttt{post-votes} & Post &
Number of votes the post receives in the next 3 months. \\
\texttt{rel-trial} & \texttt{study-adverse} & Study &
Number of affected patients with severe adverse events in the next year. \\
\texttt{rel-trial} & \texttt{site-success} & Site &
Success rate of a trial site in the next year. \\
\texttt{rel-event} & \texttt{user-attendance} & User &
Number of events the user responds yes or maybe to in the next 7 days. \\
\bottomrule
\end{tabular}
\end{table}



\subsection{Hard Regimes in RelBench}
\label{app:relbench-regimes}

\theJoin{} places special emphasis on rare-event, cold-start, heavy-tailed and long-horizon tasks. RelBench is too small to serve as a pretraining corpus, but its tasks already cover these regimes (\Cref{tab:relbench-regimes}).

\begin{table}[h]
\centering
\caption{RelBench tasks in each hard regime. A task can belong to several regimes.}
\label{tab:relbench-regimes}
\begin{tabular}{@{}l>{\raggedright\arraybackslash}p{0.62\linewidth}c@{}}
\toprule
Regime & \leavevmode RelBench tasks & Count \\
\midrule
rare event & \leavevmode\texttt{rel-avito/user-clicks}, \texttt{rel-stack/user-badge}, \texttt{rel-stack/user-engagement} & 3 \\
cold start & \leavevmode\texttt{rel-trial/study-outcome}, \texttt{rel-trial/study-adverse}, \texttt{rel-f1/driver-position}, \texttt{rel-f1/driver-dnf}, \texttt{rel-f1/driver-top3}, \texttt{rel-trial/site-success}, \texttt{rel-avito/user-clicks} & 7 \\
heavy-tailed target & \leavevmode\texttt{rel-amazon/item-ltv}, \texttt{rel-trial/study-adverse}, \texttt{rel-hm/item-sales}, \texttt{rel-stack/post-votes}, \texttt{rel-amazon/user-ltv} & 5 \\
long horizon & \leavevmode\texttt{rel-trial/study-outcome}, \texttt{rel-trial/study-adverse}, \texttt{rel-trial/site-success} & 3 \\
\bottomrule
\end{tabular}
\end{table}

\textbf{Rare events.} Three RelBench tasks sit at or below $5\%$ positive: \texttt{rel-avito/user-clicks} ($3.9\%$), \texttt{rel-stack/user-badge} ($4.8\%$), and \texttt{rel-stack/user-engagement} ($5.0\%$), comparable to the rare end of \theJoin{}.

\textbf{Cold start.} \texttt{rel-trial/study-outcome} and \texttt{rel-trial/study-adverse} are $100\%$ cold start at test time: they key on \texttt{nct\_id}, each trial contributes one row, so no test trial appears in training. This is the most extreme cold-start setting in the suite, beyond the median of \theJoin{}'s cold-start tasks. \texttt{rel-f1/driver-position} ($79\%$), \texttt{rel-f1/driver-dnf} and \texttt{rel-f1/driver-top3} ($71\%$ each), \texttt{rel-trial/site-success} ($63\%$), and \texttt{rel-avito/user-clicks} ($61\%$) follow.

\textbf{Heavy tails.} \texttt{rel-amazon/item-ltv} has target skew $133$ and kurtosis $29{,}079$; \texttt{rel-trial/study-adverse} has skew $39$; \texttt{rel-hm/item-sales}, \texttt{rel-stack/post-votes}, and \texttt{rel-amazon/user-ltv} sit near skew $21$. About $140$ tasks in \theJoin{} are tail-dominated (the top $1\%$ of values hold more than $30\%$ of the mass); the heavy-tailed RelBench tasks are extreme versions of these.

\textbf{Long horizon.} The three \texttt{rel-trial} tasks forecast a full year ahead. The \texttt{rel-amazon} and \texttt{rel-stack} tasks follow with $91$-day windows, while the remaining tasks look $4$--$60$ days ahead.

\section{Relational Transformer Architecture}
\label{sec:rt_architecture}


The \rtj{} model comprises $12$
transformer blocks with hidden dimension $d=512$, $H=8$ attention heads of
dimension $d_h=64$, and feed-forward dimension $d_{\mathrm{ff}}=2048$.
Each block applies column, feature, and neighbor attention
sequentially, followed by a gated feed-forward network. The resulting
model contains $86$M trainable parameters, excluding the
frozen text encoder.

\paragraph{Cell embeddings.}
Each cell $i$ is associated with a value $v_i$, semantic type
$\tau_i$, column $c_i$, table $t_i$, and row $r_i$.
Boolean values are cast to $\{0,1\}$ and treated as numeric; numeric values are standardized per column, while
datetime values use a shared database-wide normalization.
Text values and schema names are embedded using the frozen
\texttt{all-MiniLM-L12-v2} encoder
$\mathcal{E}:\mathrm{Text}\rightarrow\mathbb{R}^{384}$.
The additional schema-based textual information for each cell is given by:
\begin{align}
    \mathbf{s}_i
    &= \mathcal{E}
       \bigl(c_i \,\Vert\, \texttt{ of } \,\Vert\, t_i\bigr),
    \label{eq:rt_schema}
\end{align}
where $\Vert$ denotes string concatenation. For example: \texttt{"age of user"}.
Let $\mathbf{u}_i$ denote the normalized scalar or text embedding
representing the cell value. The initial token embedding is
\begin{align}
    \mathbf{e}^{\mathrm{schema}}_i
    &= \operatorname{RMSNorm}_{s}
       \left(W_s\mathbf{s}_i+\mathbf{b}_s\right),
    \label{eq:rt_schema_projection}\\
    \mathbf{e}^{\mathrm{value}}_i
    &=
    \begin{cases}
        \mathbf{m}_{\tau_i},
        & \text{if cell } i \text{ is masked},\\
        \operatorname{RMSNorm}_{\tau_i}
        \left(W_{\tau_i}\mathbf{u}_i+\mathbf{b}_{\tau_i}\right),
        & \text{otherwise},
    \end{cases}
    \label{eq:rt_value_projection}\\
    \mathbf{x}^{(0)}_i
    &= \mathbf{e}^{\mathrm{schema}}_i
       + \mathbf{e}^{\mathrm{value}}_i
       \in\mathbb{R}^{512}.
    \label{eq:rt_cell_embedding}
\end{align}
Value encoders and learned mask embeddings $\mathbf{m}_{\tau_i}$
are shared across columns of the same semantic type.
Primary and foreign keys specify connectivity rather than
contributing value embeddings. No positional or learned row-ID
embeddings are used in the model.

\paragraph{Relational attention.}
Each of the $12$ transformer blocks sequentially applies column,
feature, and neighbor attention, with $8$ heads per module.
Let $\mathcal{P}(r)$ denote the parent rows referenced by the
foreign keys of row $r$. For non-padding cells, the masks are
\begin{align}
    M^{\mathrm{col}}_{ij}
    &= \mathbb{I}\!\left[c_i=c_j \land t_i=t_j\right],
    \label{eq:rt_column_mask}\\
    M^{\mathrm{feat}}_{ij}
    &= \mathbb{I}\!\left[
        r_i=r_j \lor r_j\in\mathcal{P}(r_i)
    \right],
    \label{eq:rt_feature_mask}\\
    M^{\mathrm{nbr}}_{ij}
    &= \mathbb{I}\!\left[r_i\in\mathcal{P}(r_j)\right].
    \label{eq:rt_neighbor_mask}
\end{align}
These modules respectively exchange information within columns,
within rows and their referenced parents, and with child rows
referencing the query row. Padding cells are excluded from
all attention masks. Each module receives pre-normalized inputs
$\mathbf{z}_i=\operatorname{RMSNorm}_m(\mathbf{x}_i)$, with a separate RMSNorm per module $m$.
For head $h$, bias-free projections and QK normalization~\citep{kothapalli2026plurel}
give:
\begin{align}
    \widehat{\mathbf{q}}_{ih}
    &= \operatorname{RMSNorm}_{Q}
       \left(W_{Q,h}\mathbf{z}_i\right),\\
    \widehat{\mathbf{k}}_{jh}
    &= \operatorname{RMSNorm}_{K}
       \left(W_{K,h}\mathbf{z}_j\right),\\
    \mathbf{v}_{jh}
    &= W_{V,h}\mathbf{z}_j.
\end{align}

Queries and keys are RMS-normalized along the head dimension,
with separate learned feature-wise gains shared across heads.
Normalized queries are additionally multiplied by a learned
head-specific scalar and the logarithm of the valid-key count. For a module with mask $M$, the attention logits are
\begin{align}
    k_i &= \sum_j M_{ij},\\
    \ell_{ijh}
    &=
    \begin{cases}
        \displaystyle
        \frac{\alpha_h\log\!\left(\max(k_i,1)\right)}{d_h}
        \widehat{\mathbf{q}}_{ih}^{\top}
        \widehat{\mathbf{k}}_{jh},
        & M_{ij}=1,\\
        -\infty, & M_{ij}=0,
    \end{cases}
    \label{eq:rt_attention_logits}
\end{align}
where $\alpha_h$ is a learned head-specific scale.
For queries with at least one accessible key,
\begin{align}
    a_{ijh}
    &= \operatorname{softmax}_{j}(\ell_{ijh}),\\
    \mathbf{o}_i
    &= \operatorname{Concat}_{h=1}^{8}
       \left(\sum_j a_{ijh}\mathbf{v}_{jh}\right),\\
    \mathbf{g}_i
    &= 2\sigma\!\left(W_G\mathbf{z}_i\right),\\
    \mathbf{x}_i
    &\leftarrow
    \mathbf{x}_i+
    W_O\left(\mathbf{g}_i\odot\mathbf{o}_i\right).
    \label{eq:rt_gated_attention}
\end{align}
Queries with no accessible keys receive a zero attention
contribution. Each attention module has independent parameters.
After the three attention modules, a pre-normalized residual
SwiGLU network completes the block:
\begin{align}
    \mathbf{z}_i
    &= \operatorname{RMSNorm}_{\mathrm{ffn}}(\mathbf{x}_i),\\
    \mathbf{x}_i
    &\leftarrow \mathbf{x}_i+
    W_2\left[
        \operatorname{SiLU}(W_1\mathbf{z}_i)
        \odot W_3\mathbf{z}_i
    \right],
    \label{eq:rt_ffn}
\end{align}
with intermediate dimension $2048$.
All attention and feed-forward projections are bias-free,
and all RMSNorm operations use $\epsilon=10^{-6}$.

\paragraph{Prediction heads.}
A final RMSNorm and type-specific affine decoders produce
\begin{align}
    \mathbf{h}_i
    &= \operatorname{RMSNorm}_{\mathrm{out}}
       \left(\mathbf{x}^{(12)}_i\right),\\
    \widehat{\mathbf{u}}_i
    &= D_{\tau_i}\mathbf{h}_i+
       \mathbf{b}^{\mathrm{out}}_{\tau_i}.
    \label{eq:rt_prediction}
\end{align}
Numeric and datetime heads predict normalized scalars, and the text head predicts the $384$-dimensional text embedding. All targets, including binary classification labels, are regressed with the Huber loss ($\delta=1$, averaged over output dimensions) as $\ell_{\tau_i}$. For $B$ contexts with target sets 
$\mathcal{T}_b$, the objective is
\begin{align}
    \mathcal{L}
    &=
    \frac{1}{B}\sum_{b=1}^{B}
    \frac{1}{\max(|\mathcal{T}_b|,1)}
    \sum_{i\in\mathcal{T}_b}
    \ell_{\tau_i}
    \left(\widehat{\mathbf{u}}_i,\mathbf{u}_i\right).
    \label{eq:rt_objective}
\end{align}
All prediction-head weights and biases are initialized to zero.
The complete model contains $85{,}562{,}530$ trainable
parameters, excluding the frozen text encoder.




\section{Pretraining Details}
\label{app:pretraining-details}

In all our experiments we use an $86$M-parameter RT model
with 12 layers, 512 hidden size and 8 attention heads,
incorporating the improvements of \citep{kothapalli2026plurel} (such as QK-Norm)
over the original architecture of \citep{rt}.
We sample context size $L$ from
$\mathcal{L} = \{512, 1024, 2048, 4096, 8192\}$,
the local context size $\ell$ from
$\mathcal{L}_\ell = \{128, 256, 512, 1024, 2048, 4096, 8192\}$,
the BFS width-bound $b$ from $\mathcal{B} = \{8, 16, 32, 64, 128, 256\}$,
and the recency parameter $\rho$ from $\mathcal{R} = \{0, 1\}$.
The random-walk retriever uses $W = 10{,}000$ walks of length
$K = 20$, and only the target cell is masked ($p_{\max} = 0$).

We pretrain in two phases: first on the synthetic PluRel
corpus~\citep{kothapalli2026plurel} ($86{,}211$ tasks over $1{,}900$
databases), then continuing from that checkpoint on a fixed mixture of
$13{,}243$ \theJoin{} tasks drawn from the $523$ databases whose
\emph{preprocessed} on-disk footprint stays under $5$\,GB.
Each phase is early-stopped once neither mean validation AUROC nor mean validation nMAE (of the live or SWA weights) has improved
for $10$K optimizer steps (at $18$K steps for PluRel and $26$K for \theJoin{}),
at a global batch size of
$1024$
contexts per step
using Muon~\citep{jordan2024muon} for $2$D weight matrices and AdamW for the remaining parameters,
at a constant learning rate of $5 \times 10^{-4}$
(RMS-matched for Muon as in \cite{liu2025muon})
after a linear warmup over the first $2000$ steps.
To compensate for a lack of decay phase~\citep{tian2025wsm},
and following the recommendation in \citep{ranjan2024post},
we track the stochastic weight average (SWA) performance during training
and select the best SWA checkpoint based on validation performance.
Specifically, we use an exponential moving average (EMA) for SWA,
with momentum $0.9995$, which gives an effective window of $\approx\!2000$ steps,
and find that it produces more stable curves with better results.

Each phase takes about a day or less on a single node with $8\times$A100 GPUs, with BFloat16 activations and FP32 master weights~\citep{kapfer_2025_14751899}.
The released \rtj{} checkpoint is the SWA weights after $9$K steps of
the \theJoin{} phase, selected by mean validation AUROC ($74.36$);
selection runs over the whole SWA trajectory, so the selected step lies
well before the step at which the phase stopped. The same
single checkpoint is used for every task, classification and
regression alike.

\section{Pretraining Ablations}
\label{app:pretraining-ablations}

\begin{figure}[t]
  \centering
  \begin{subfigure}[t]{0.49\linewidth}
    \centering
    \includegraphics[width=\linewidth]{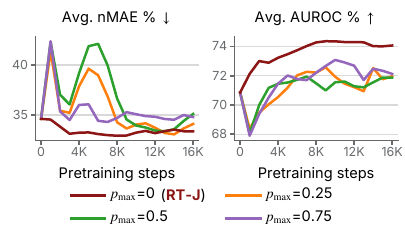}
    \caption{Multi-cell masking ablation}
    \label{fig:mask_prob}
  \end{subfigure}\hfill
  \begin{subfigure}[t]{0.49\linewidth}
    \centering
    \includegraphics[width=\linewidth]{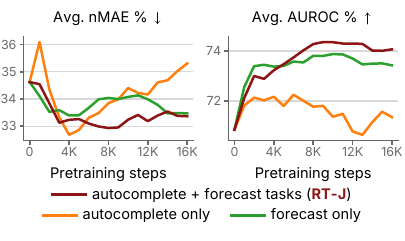}
    \caption{Pretraining task-mix ablation}
    \label{fig:task_mix}
  \end{subfigure}
  \caption{
  Pretraining ablations: multi-cell masking rate and task mix,
    as validation nMAE / AUROC over pretraining steps. Each arm differs
    from \rtj{}'s own pretraining run in one knob. (a) Masking non-target cells at any rate $p_{\max}>0$
    costs both metrics; \rtj{} masks only the target cell.
    (b) Forecasting tasks alone nearly match the mixture on
    classification but lose on regression, and autocompletion alone
    trails on both; the mixture is competitive on both.}
  \label{fig:pretraining-ablations}
\end{figure}

\subsection{Multi-cell masking rate}
\label{sec:abl-mask}

The original RT recipe~\citep{rt} masks only a single target cell 
in a given context of length $L$.
Following prior work on masked pretraining and meta-learning~\citep{devlin2019bert, dong2025r2mae, kothapalli2025cot, kothapalli2026think},
the framework also supports masking non-target
cells with a hierarchical Bernoulli scheme:
per training context $i$
a rate $p_i \sim \mathrm{Unif}\bigl([0, p_{\max}]\bigr)$ is drawn,
and each non-target, non-missing cell is masked independently with that rate
$m_{i,j} \sim \mathrm{Bernoulli}(p_i)$;
\rtj{} is pretrained with $p_{\max}=0$,
i.e., only the target cell is masked.

The hyperparameter $p_{\max}$ of the hierarchical Bernoulli masking rate
controls how many non-target cells are
masked, and thus how much supervision each context carries.
To test whether denser supervision would have been
better, we sweep $p_{\max}$ over $\{0, 0.25, 0.5, 0.75\}$, with identical data
and context-sampling settings, all arms warm-started from the same
checkpoint.

\paragraph{Observations.} \Cref{fig:mask_prob} shows that masking non-target
cells hurts both metrics. Over the common window of $10$K--$18$K pretraining
steps, single-cell masking averages $74.2$ validation AUROC and $33.3$ nMAE,
while every masked rate is $1.6$--$2.6$ AUROC points and $0.6$--$1.5$ nMAE
behind ($71.6$/$33.9$ at $p_{\max}=0.25$, $71.6$/$34.0$ at $0.5$, and
$72.5$/$34.8$ at $0.75$); the gap holds at each arm's own best step
($74.4$/$32.9$ against a best of $73.1$/$33.1$ across the masked arms).
Masked non-target cells corrupt the very context the model predicts from, and
at this model and data scale that cost is not repaid by the extra supervision
per step. Denser supervision is therefore not free: it buys gradient signal by
degrading the in-context evidence, and at a fixed step budget the trade is a
losing one.

\subsection{Pretraining task mix}
\label{sec:abl-mix}

\theJoin{} supplies two families of pretraining tasks, forecasting and
autocompletion (\Cref{sec:corpus}), and \rtj{} is pretrained on both. We
ablate the mixture by pretraining on each family alone, holding everything
else fixed.

\paragraph{Observations.} From \Cref{fig:task_mix}, the mixture is the only
setting competitive on both metrics. Over the same $10$K--$18$K window, the
mixed objective reaches $74.2$ AUROC and $33.3$ nMAE; forecasting alone is
close on classification ($73.6$) but $0.4$ nMAE behind, and autocompletion
alone trails badly on classification ($71.2$, three points down) while also
being the worst on regression ($34.8$). The single-family arms are not simply
slower: autocompletion alone posts the best regression value of any arm early
($32.7$ nMAE at $4$K steps) and then degrades steadily, which is what one
expects when the pretraining distribution stops resembling the RelBench
forecasting tasks the model is evaluated on. Mixing the two families is what
keeps both metrics improving together, so we keep it in the recipe.

\subsection{Compute scaling}
\label{app:compute-scaling}
\begin{figure}[h]
  \centering
  \includegraphics[width=0.49\linewidth]{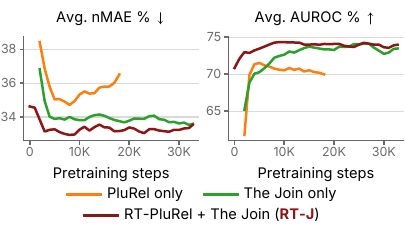}
  \caption{Validation nMAE and AUROC (SWA, $8192$-cell context) over pretraining steps. The two runs from scratch are shown from step $2$K; the PluRel-only run stops early at $18$K steps.}
  \label{fig:compute-scaling}
\end{figure}

\Cref{fig:compute-scaling} tracks validation nMAE and AUROC over pretraining for three runs with the same recipe: from scratch on synthetic PluRel data only, from scratch on \theJoin{} only, and continued pretraining of \rtp{} on \theJoin{}, which yields \rtj{}. Real data scales better than synthetic data: the PluRel-only run stops improving after about $8$K steps and degrades on regression thereafter, while the \theJoin{}-only run keeps slowly improving regression up to $32$K steps. Continued pretraining from \rtp{} reaches its best validation metrics within about $8$K steps and stays ahead of pretraining on \theJoin{} from scratch on regression throughout, and on classification at all but a few steps around $24$K, so most of its benefit arrives early.

\section{Random-Walk Context Construction}
\label{app:rw-context}

This appendix gives the full procedure for context construction summarized in
\Cref{sec:training-methodology}. For one task item, we score same-table candidate
rows by random-walk visit counts (Stage 1), then consume the candidates in
order---most recent first with visit counts breaking ties when the recency
flag $\rho = 1$, or by visit count alone when $\rho = 0$---as launch points
for the width-bounded BFS cell collector \textsc{BFSCollect} of the
original RT, which we reuse unchanged (Stage 2). Walks step uniformly at
random along PK and FK edges, restricted to neighbor rows whose timestamp
does not exceed the target's timestamp $t^\star$; this enforces temporal
validity at the walk level and avoids leaking future information into the
context. Only visits to rows other than the target in the target's table
$\tau$ contribute, so the score ranks same-table rows by their
personalized random-walk proximity to the target.

\begin{algorithm}[t]
\caption{Random-walk context construction (one task item).}
\label{alg:rw-context}
\begin{algorithmic}[1]
\Require target row $v^\star$ in table $\tau$ with timestamp $t^\star$;
  database $\mathcal{D}$; total cell budget $L$; per-seed cell budget $\ell$;
  BFS width-bound $b$; recency flag $\rho \in \{0,1\}$;
  walks $W$; walk length $K$.
\Ensure cell sequence $\mathbf{x}$ with $|\mathbf{x}| \le L$.
\Statex \emph{Stage 1: score same-table candidates by random-walk visits.}
\State $\mathrm{score}[u] \gets 0$ for every row $u$.
\For{$w = 1$ \textbf{to} $W$}
  \State $u \gets v^\star$.
  \For{$k = 1$ \textbf{to} $K$}
    \State Move $u$ to a uniformly random PK/FK neighbor whose timestamp is $\le t^\star$;
           if no such neighbor exists, end this walk.
    \If{$u \neq v^\star$ \textbf{and} $u$ is in table $\tau$}
      \State $\mathrm{score}[u] \gets \mathrm{score}[u] + 1$.
    \EndIf
  \EndFor
\EndFor
\Statex \emph{Stage 2: launch BFS from the ordered candidates.}
\State If $\rho = 1$, order same-table rows by (timestamp, $\mathrm{score}$) lexicographically descending; otherwise order by $\mathrm{score}$ descending.
\State $\mathbf{x} \gets$ empty cell sequence.
\For{each row $s$ in this order}
  \If{$|\mathbf{x}| \ge L$} \textbf{break}. \EndIf
  \State Append \textsc{BFSCollect}$(\mathcal{D}, s, \ell, b, t^\star, L - |\mathbf{x}|)$ to $\mathbf{x}$.
\EndFor
\State \Return $\mathbf{x}$.
\end{algorithmic}
\end{algorithm}

\subsection{Sensitivity to random-walk parameters}
\label{app:rw-sensitivity}
The random-walk retriever of \Cref{app:rw-context} uses $W = 10{,}000$ walks of length $K = 20$ by default. We evaluate \rtj{} with the default and three cheaper settings on the $8192$-row test subsample at $L = 8192$, over three context seeds. \Cref{tab:rw-latency} breaks down the per-example inference cost and \Cref{tab:rw-auroc,tab:rw-nmae} the accuracy. Fewer or shorter walks cut the context-sampling cost from $28.1$ to $14.3$--$17.9$ ms per example, while the forward pass stays at about $10$ ms. Classification degrades steadily as the walks shrink, from $72.9$ to $71.9$ mean AUROC, driven by \texttt{rel-trial/study-outcome} ($59.4$ to $52.7$--$55.8$), \texttt{rel-stack/user-engagement} and \texttt{rel-f1/driver-dnf}. Regression is insensitive on average ($33.3$--$33.6$ mean nMAE), although individual tasks move in both directions: \texttt{rel-hm/item-sales} improves with shorter walks ($14.1$ to $11.6$ nMAE at $(10\mathrm{K}, 10)$), while \texttt{rel-f1/driver-position} is best at the default. We keep the default.

\begin{table}[h]
\centering
\setlength{\tabcolsep}{4.5pt}
\caption{Per-example inference cost in ms at batch size $1$ and $L = 8192$, on one B200 GPU with one AMD EPYC 9565 CPU: per-task means over $100$ timed steps after $5$ warmup steps, then mean $\pm$ std.\ over the $21$ RelBench tasks.}
\label{tab:rw-latency}
\begin{tabular}{@{}lcccc@{}}
\toprule
                Stage  &  $(10\mathrm{K}, 20)$  &  $(10\mathrm{K}, 10)$  &  $(1\mathrm{K}, 20)$  &  $(1\mathrm{K}, 10)$ \\
\midrule
 Context sampler (CPU)  &     $28.06 \pm 10.39$  &      $17.87 \pm 8.14$  &     $14.31 \pm 7.99$  &     $16.00 \pm 8.49$ \\
   Forward pass (GPU)  &      $10.03 \pm 0.38$  &       $9.85 \pm 0.42$  &      $9.81 \pm 0.43$  &      $9.78 \pm 0.45$ \\
\midrule
 Total & $38.09 \pm 10.31$ & $27.72 \pm 8.12$ & $24.11 \pm 7.99$ & $25.78 \pm 8.52$ \\
\bottomrule
\end{tabular}

\end{table}

\begin{table}[h]
\centering
\setlength{\tabcolsep}{4.5pt}
\caption{AUROC (\%, higher is better) of \rtj{} for $(W, K)$ on the $8192$-row test subsample at $L = 8192$, mean $\pm$ std.\ over $3$ context seeds; the mean row averages the tasks per seed. Bold: within one standard deviation of the best per row.}
\label{tab:rw-auroc}
\begin{tabular}{@{}lcccc@{}}
\toprule
                              Task  &       $(10\mathrm{K}, 20)$  &       $(10\mathrm{K}, 10)$  &        $(1\mathrm{K}, 20)$  &        $(1\mathrm{K}, 10)$ \\
\midrule
    \texttt{rel-amazon/item-churn}  &           $76.67 \pm 0.05$  &           $76.69 \pm 0.06$  &  $\mathbf{76.79} \pm 0.12$  &  $\mathbf{76.91} \pm 0.14$ \\
    \texttt{rel-amazon/user-churn}  &           $66.88 \pm 0.10$  &  $\mathbf{67.05} \pm 0.11$  &           $66.69 \pm 0.14$  &           $66.50 \pm 0.12$ \\
    \texttt{rel-avito/user-clicks}  &  $\mathbf{52.31} \pm 0.98$  &  $\mathbf{52.83} \pm 2.65$  &  $\mathbf{53.14} \pm 3.41$  &  $\mathbf{53.87} \pm 2.69$ \\
    \texttt{rel-avito/user-visits}  &           $61.70 \pm 0.22$  &  $\mathbf{62.68} \pm 0.62$  &  $\mathbf{62.50} \pm 0.62$  &  $\mathbf{62.42} \pm 0.59$ \\
    \texttt{rel-event/user-ignore}  &  $\mathbf{79.91} \pm 1.95$  &  $\mathbf{79.55} \pm 2.20$  &  $\mathbf{78.94} \pm 2.16$  &           $77.71 \pm 0.84$ \\
    \texttt{rel-event/user-repeat}  &  $\mathbf{77.91} \pm 1.13$  &  $\mathbf{77.48} \pm 1.24$  &  $\mathbf{77.07} \pm 1.23$  &           $75.70 \pm 1.73$ \\
        \texttt{rel-f1/driver-dnf}  &  $\mathbf{82.82} \pm 0.17$  &           $82.43 \pm 0.03$  &           $81.90 \pm 0.13$  &           $81.40 \pm 0.09$ \\
       \texttt{rel-f1/driver-top3}  &  $\mathbf{90.75} \pm 0.45$  &  $\mathbf{91.18} \pm 0.40$  &  $\mathbf{90.86} \pm 0.26$  &  $\mathbf{90.95} \pm 0.27$ \\
        \texttt{rel-hm/user-churn}  &           $62.31 \pm 0.21$  &           $62.03 \pm 0.21$  &           $61.76 \pm 0.22$  &  $\mathbf{63.48} \pm 0.03$ \\
     \texttt{rel-stack/user-badge}  &  $\mathbf{79.74} \pm 0.45$  &  $\mathbf{79.55} \pm 0.57$  &  $\mathbf{79.75} \pm 0.71$  &  $\mathbf{79.43} \pm 0.58$ \\
\texttt{rel-stack/user-engagement}  &  $\mathbf{84.73} \pm 0.43$  &           $82.60 \pm 0.55$  &           $83.18 \pm 0.37$  &           $82.12 \pm 0.39$ \\
  \texttt{rel-trial/study-outcome}  &  $\mathbf{59.36} \pm 1.72$  &           $55.83 \pm 1.15$  &           $52.95 \pm 2.20$  &           $52.74 \pm 3.00$ \\
\midrule
 Mean & $\mathbf{72.92} \pm 0.27$ & $72.49 \pm 0.14$ & $72.13 \pm 0.28$ & $71.93 \pm 0.34$ \\
\bottomrule
\end{tabular}

\end{table}

\begin{table}[h]
\centering
\setlength{\tabcolsep}{4.5pt}
\caption{nMAE (\%, lower is better) of \rtj{} for $(W, K)$ on the $8192$-row test subsample at $L = 8192$, mean $\pm$ std.\ over $3$ context seeds; the mean row averages the tasks per seed. Bold: within one standard deviation of the best per row.}
\label{tab:rw-nmae}
\begin{tabular}{@{}lcccc@{}}
\toprule
                              Task  &       $(10\mathrm{K}, 20)$  &       $(10\mathrm{K}, 10)$  &        $(1\mathrm{K}, 20)$  &        $(1\mathrm{K}, 10)$ \\
\midrule
      \texttt{rel-amazon/item-ltv}  &            $8.49 \pm 0.01$  &            $8.25 \pm 0.09$  &            $8.35 \pm 0.03$  &   $\mathbf{8.07} \pm 0.03$ \\
      \texttt{rel-amazon/user-ltv}  &  $\mathbf{30.21} \pm 0.10$  &           $30.44 \pm 0.07$  &           $30.45 \pm 0.07$  &  $\mathbf{30.17} \pm 0.11$ \\
         \texttt{rel-avito/ad-ctr}  &  $\mathbf{42.17} \pm 0.55$  &  $\mathbf{42.22} \pm 0.45$  &           $43.07 \pm 0.64$  &           $43.11 \pm 0.59$ \\
\texttt{rel-event/user-attendance}  &  $\mathbf{43.46} \pm 1.66$  &  $\mathbf{43.38} \pm 1.71$  &  $\mathbf{43.45} \pm 1.70$  &  $\mathbf{43.78} \pm 1.85$ \\
   \texttt{rel-f1/driver-position}  &  $\mathbf{38.53} \pm 0.03$  &           $39.53 \pm 0.10$  &           $39.97 \pm 0.18$  &           $39.75 \pm 0.07$ \\
        \texttt{rel-hm/item-sales}  &           $14.07 \pm 0.20$  &  $\mathbf{11.61} \pm 0.09$  &           $12.56 \pm 0.07$  &           $13.38 \pm 0.06$ \\
     \texttt{rel-stack/post-votes}  &           $15.54 \pm 0.04$  &           $15.46 \pm 0.04$  &           $15.46 \pm 0.02$  &  $\mathbf{15.21} \pm 0.04$ \\
   \texttt{rel-trial/site-success}  &  $\mathbf{91.10} \pm 0.14$  &           $91.94 \pm 0.29$  &           $91.80 \pm 0.31$  &           $92.18 \pm 0.40$ \\
  \texttt{rel-trial/study-adverse}  &  $\mathbf{16.39} \pm 0.03$  &           $16.61 \pm 0.04$  &           $16.76 \pm 0.00$  &           $16.90 \pm 0.05$ \\
\midrule
 Mean & $\mathbf{33.33} \pm 0.15$ & $\mathbf{33.27} \pm 0.18$ & $33.54 \pm 0.20$ & $33.62 \pm 0.20$ \\
\bottomrule
\end{tabular}

\end{table}

\section{Additional and Per-Task Results}
\label{app:additional-results}

\subsection{Baselines with extended context}
\label{app:baselines-subsampled}

\begin{figure}[!htbp]
  \centering
  \includegraphics[width=\linewidth]{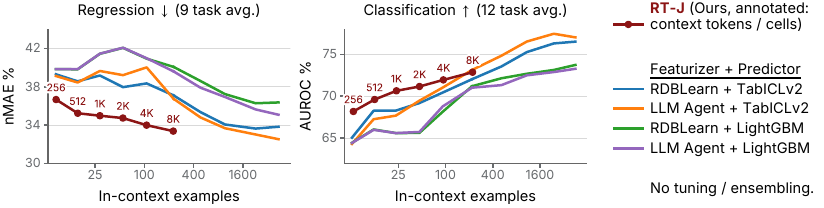}
  \caption{
  Few-shot relational prediction vs.\ in-context-learning baselines on
    a fixed $8192$-row-per-task subsample of the RelBench test set, with the
    baselines extended to $131\mathrm{K}$-cell contexts. As in
    \Cref{fig:baselines}, the $x$-axis is the mean number of in-context labeled
    task rows and the markers on the \rtj{} curve are annotated with the
    underlying context size (\rtj{} caps at its $8\mathrm{K}$-cell training
    context). With $20\times$ more labels than \rtj{} ever sees, the
    TabICLv2 pipelines lead on classification, and LLM Agent + TabICLv2
    passes \rtj{}'s regression error only from $65\mathrm{K}$ cells on.}
  \label{fig:baselines-extended}
\end{figure}

\Cref{fig:baselines-extended} repeats the aggregate baselines comparison of
\Cref{fig:baselines} on a fixed $8192$-row test subsample per task,
extending all four baselines to a $131\mathrm{K}$-cell context
(${\sim}4{,}500$ in-context examples on average, $20\times$ what \rtj{}
sees at $8\mathrm{K}$ cells)---far past \rtj{}'s $8\mathrm{K}$ training
context. The extra context pays off on classification: the TabICLv2
pipelines pass \rtj{}'s $8\mathrm{K}$-context AUROC by $3.6$--$4.2$ points at
the largest context and the LightGBM pipelines by up to about $1$ point.
On regression, only LLM Agent + TabICLv2 passes \rtj{}, and only
from $65\mathrm{K}$ cells on ($32.5$ vs.\ $33.4$ nMAE at
$131\mathrm{K}$); RDBLearn + TabICLv2
improves steadily with context but stays $0.3$--$0.5$ nMAE behind at the two
largest sizes, and the LightGBM pipelines stay $1.7$--$5.2$ nMAE behind at
every extended context size ($16\mathrm{K}$--$131\mathrm{K}$ cells).

\subsection{Per-task baseline curves}
\label{app:per-task-baselines}

\begin{figure}[!htbp]
  \centering
  \includegraphics[width=\linewidth]{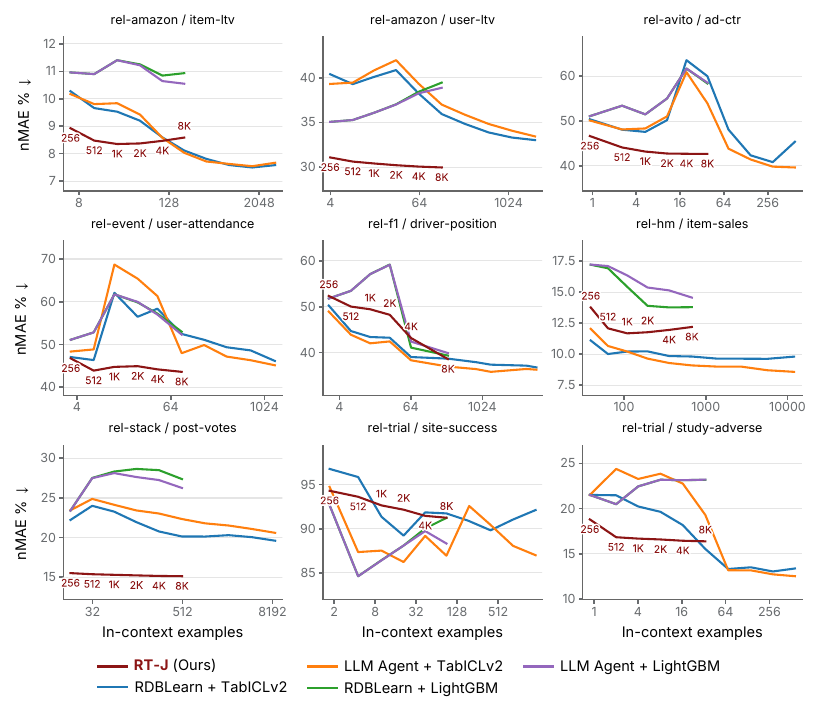}
  \caption{Per-task test nMAE versus the average number of in-context examples on RelBench regression tasks; per-task counterpart of \Cref{fig:baselines} (left).}
  \label{fig:per-task-reg}
\end{figure}

\begin{figure}[!htbp]
  \centering
  \includegraphics[width=\linewidth]{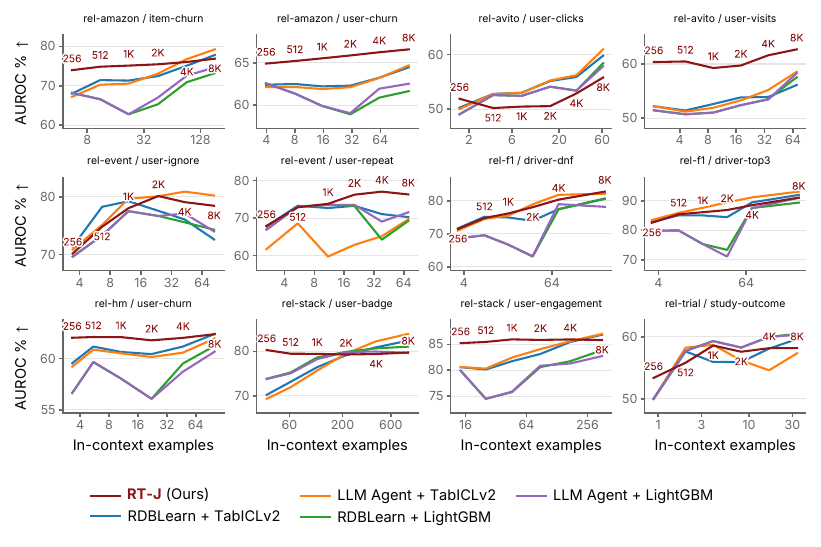}
  \caption{Per-task test AUROC versus the average number of in-context examples on RelBench classification tasks; per-task counterpart of \Cref{fig:baselines} (right).}
  \label{fig:per-task-clf}
\end{figure}

\Cref{fig:per-task-reg,fig:per-task-clf} disaggregate the aggregate
baselines comparison of \Cref{fig:baselines} into one panel per RelBench
task, plotting test nMAE for the $9$ regression tasks and test AUROC for
the $12$ classification tasks against inference-time context size. The
five methods, the context grid, and the full-test evaluation protocol
match \Cref{sec:results}; per-task $y$-axes are auto-scaled.

On regression, \rtj{}'s aggregate dominance in \Cref{fig:baselines} is
anchored by wide wins at every context size on tasks whose targets are
heavy-tailed or spread across the graph (\texttt{rel-stack/post-votes},
\texttt{rel-amazon/user-ltv}, \texttt{rel-event/user-attendance},
\texttt{rel-avito/ad-ctr}), where the flat-feature pipelines stall or get
worse with more context. Elsewhere the TabICLv2 pipelines keep improving
with context and pass \rtj{} at the largest budgets
(\texttt{rel-amazon/item-ltv}), lead it at every context size on \texttt{rel-hm/item-sales} and \texttt{rel-f1/driver-position}, as do the feature-engineering pipelines on \texttt{rel-trial/site-success}, while
\texttt{rel-trial/study-adverse} ends $0.8$ nMAE behind RDBLearn + TabICLv2.
On classification, \rtj{} leads or ties on nearly every task at small
contexts---the label-scarce regime this paper targets. At $8\mathrm{K}$
the picture is mixed: \rtj{} stays ahead on \texttt{rel-event/user-repeat}, \texttt{rel-amazon/user-churn}, \texttt{rel-f1/driver-dnf} and \texttt{rel-avito/user-visits} and ties on \texttt{rel-hm/user-churn}, while the
feature-engineering pipelines overtake it on
\texttt{rel-avito/user-clicks}, \texttt{rel-amazon/item-churn}, the two
\texttt{rel-stack} tasks, \texttt{rel-trial/study-outcome}, \texttt{rel-event/user-ignore} and \texttt{rel-f1/driver-top3}, mirroring the aggregate convergence at the largest
label budget.

\subsection{Cost of test-time compute scaling}
\label{app:ttc-cost}

\Cref{tab:ttc-cost} compares the cost of context tuning and context ensembling (\Cref{sec:test-time-compute}), counted in passes of the context sampler and model over an evaluation split, and in labels used beyond the training split from which in-context examples are drawn. See \Cref{tab:rw-latency} for forward pass latencies.

\begin{table}[h]
\centering
\caption{Passes over an evaluation split, in total and on the critical path when fully parallelized, and labels used beyond the training split.}
\label{tab:ttc-cost}
\begin{tabular}{@{}lccc@{}}
\toprule
 Setting & Total passes & Fully parallel & Extra labels \\
\midrule
 Default context & $1$ test & $1$ test & none \\
Tuning ($120$ configurations, $4$ seeds) & $144$ val.\ $+$ $1$ test & $1$ val.\ $+$ $1$ test & $\le 4096$ (val.) \\
Ensembling ($16$ seeds) & $16$ test & $1$ test & none \\
\bottomrule
\end{tabular}
\end{table}

Tuning scores $120$ configurations $(L, \ell, b, \rho)$ with $4$ context seeds each. A single pass samples one context at the largest $L$ and scores every $L \ge \ell$ on its prefixes, so the grid takes $36$ passes per seed over the shared $(\ell, b, \rho)$ settings, i.e.\ $144$ passes over a validation subsample of up to $4096$ rows, plus one test pass with the selected configuration. Ensembling needs no labels beyond the training split. Both procedures parallelize fully across configurations and seeds.

\subsection{Per-task ablation curves}
\label{app:per-task-ablations}

This appendix disaggregates each ablation in
\Cref{sec:ablations} into per-task panels, one per RelBench task,
evaluated on the $8192$-row test subsample of \Cref{sec:ablations}
across context sizes; per-panel $y$-axes are auto-scaled so the visual
ranking of methods within a task is preserved even where its absolute
scale differs. These panels disaggregate the aggregate ablation figures
(context retrieval, schema semantics) directly.

\subsubsection{Context retrieval (per task)}
\label{app:per-task-retrievers}

Per-task counterpart of \Cref{fig:retriever}. Random walk's advantage
at small $L$ is broad: at $L=256$ it is the best or tied retriever on
most regression tasks. Beyond $L=1024$ RDBLearn-feature similarity leads
on most of them (\texttt{item-ltv}, \texttt{user-ltv}, \texttt{ad-ctr},
\texttt{item-sales}, \texttt{post-votes}, \texttt{study-adverse}) and
the two are close on \texttt{user-attendance} and
\texttt{driver-position}; on \texttt{site-success} the
structure-agnostic retrievers stay flat near $94$ nMAE, $32$-width BFS is
far ahead beyond $L=1024$, random walk drifts back up past $L=1024$ and
RDBLearn-feature similarity is erratic. On classification the same two
retrievers lead nearly everywhere, RDBLearn-feature similarity by the
wider margin on the churn and driver tasks and random walk on the
\texttt{rel-event} tasks; the tasks with few labels (\texttt{user-clicks},
\texttt{user-engagement}, \texttt{study-outcome}) are noisy for every
retriever. The $256$-width BFS variant is a strong outlier at small $L$ on
several tasks; it is excluded from per-panel $y$-limit computation so
the other curves remain readable, but its line is still drawn.

\begin{figure}[!htbp]
  \centering
  \includegraphics[width=\linewidth]{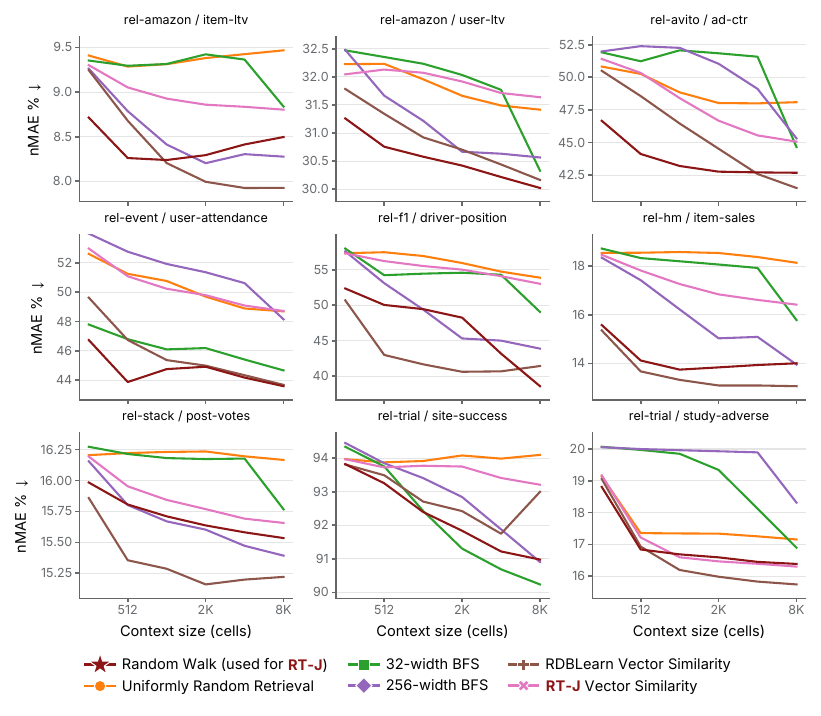}
  \caption{Per-task test nMAE for context-retrieval ablations on RelBench regression tasks; per-task counterpart of \Cref{fig:retriever} (left).}
  \label{fig:per-task-retriever-reg}
\end{figure}

\begin{figure}[!htbp]
  \centering
  \includegraphics[width=\linewidth]{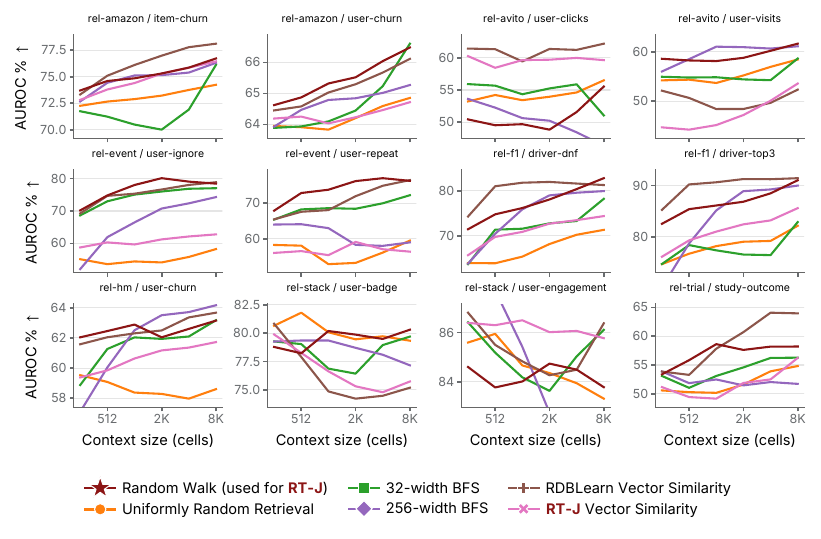}
  \caption{Per-task test AUROC for context-retrieval ablations on RelBench classification tasks; per-task counterpart of \Cref{fig:retriever} (right).}
  \label{fig:per-task-retriever-clf}
\end{figure}

\subsubsection{Schema semantics (per task)}
\label{app:per-task-semantics}

Per-task counterpart of \Cref{fig:semantics}. Schema semantics help on
most tasks. The largest classification gaps
(\texttt{rel-event/user-ignore}, \texttt{rel-event/user-repeat}) are
worth up to $5.2$ AUROC points; shuffled names win at every context size on \texttt{rel-avito/user-clicks}, \texttt{rel-f1/driver-top3} and \texttt{rel-hm/item-sales}, on \texttt{rel-amazon/user-ltv} from $L=512$, and on \texttt{rel-f1/driver-position} by up to $5.1$ nMAE at small contexts.

\begin{figure}[!htbp]
  \centering
  \includegraphics[width=0.8\linewidth]{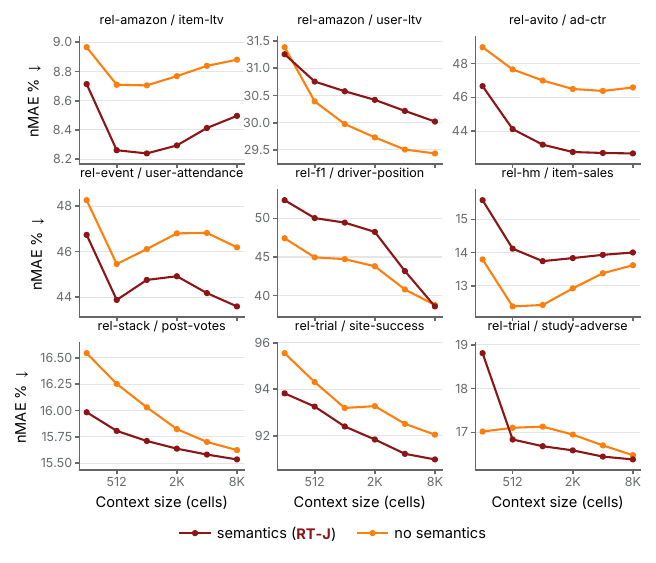}
  \caption{Per-task test nMAE for the schema-semantics ablation on RelBench regression tasks; per-task counterpart of \Cref{fig:semantics} (left).}
  \label{fig:per-task-semantics-reg}
\end{figure}

\begin{figure}[!htbp]
  \centering
  \includegraphics[width=\linewidth]{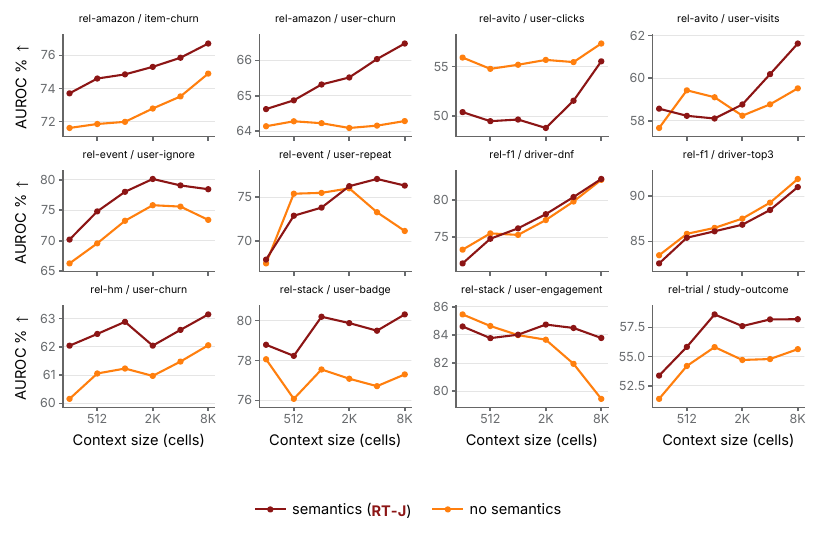}
  \caption{Per-task test AUROC for the schema-semantics ablation on RelBench classification tasks; per-task counterpart of \Cref{fig:semantics} (right).}
  \label{fig:per-task-semantics-clf}
\end{figure}

\subsubsection{Context hyperparameter selection (per task)}
\label{app:per-task-tune-table}

Each task's \emph{tuned} context is chosen by a grid search over the global context length $L \in \{512, \dots, 8192\}$, the per-seed cell budget $\ell \in \{256, \dots, 8192\}$ (with $\ell \le L$), the BFS width $b \in \{8, 32, 128\}$, and the recency flag $\rho \in \{0, 1\}$---$120$ configurations per task---selecting per task the configuration that maximizes validation performance on a $4096$-row validation subsample (four context seeds, per-row predictions averaged before scoring; nMAE for regression, AUROC for classification). \Cref{tab:val_vs_test_ctx8k} reports the resulting test metrics on the 8192-subsampled test set, comparing this tuned context against the shared \emph{default} context ($L = 8\mathrm{K}$, $\ell = 256$, $b = 32$, $\rho = 1$) at a single context seed; columns $L$, $\ell$, $b$, $\rho$ report the selected values.

\begin{table}[h]
\centering
\caption{Per-task test metrics on the 8192-subsampled test set: shared \emph{default} context ($L = 8\mathrm{K}$, $\ell = 256$, $b = 32$, $\rho = 1$) vs.\ the per-task \emph{tuned} context. Columns $L$, $\ell$, $b$, $\rho$ report the configuration selected on the validation split. Single context seed (no ensembling). All metrics shown in percent. Best per row in bold; lower is better for nMAE, higher for AUROC.}
\label{tab:val_vs_test_ctx8k}
\small
\begin{tabular}{llrrrcrr}
\toprule
DB & Task & $L$ & $\ell$ & $b$ & $\rho$ & Default & Tuned \\
\midrule
\multicolumn{8}{l}{\textit{Regression (nMAE \%, lower is better)}} \\
\midrule
\texttt{rel-amazon} & \texttt{item-ltv} & $4\mathrm{K}$ & $2048$ & $8$ & $0$ & $8.50$ & $\mathbf{7.88}$ \\
\texttt{rel-amazon} & \texttt{user-ltv} & $4\mathrm{K}$ & $2048$ & $8$ & $0$ & $30.02$ & $\mathbf{29.45}$ \\
\texttt{rel-avito} & \texttt{ad-ctr} & $8\mathrm{K}$ & $1024$ & $8$ & $0$ & $42.69$ & $\mathbf{42.47}$ \\
\texttt{rel-event} & \texttt{user-attendance} & $8\mathrm{K}$ & $256$ & $8$ & $1$ & $43.59$ & $\mathbf{42.54}$ \\
\texttt{rel-f1} & \texttt{driver-position} & $8\mathrm{K}$ & $256$ & $8$ & $1$ & $\mathbf{38.60}$ & $38.60$ \\
\texttt{rel-hm} & \texttt{item-sales} & $512$ & $256$ & $128$ & $0$ & $14.00$ & $\mathbf{11.79}$ \\
\texttt{rel-stack} & \texttt{post-votes} & $8\mathrm{K}$ & $2048$ & $8$ & $0$ & $15.51$ & $\mathbf{15.03}$ \\
\texttt{rel-trial} & \texttt{site-success} & $8\mathrm{K}$ & $4096$ & $32$ & $0$ & $90.98$ & $\mathbf{88.61}$ \\
\texttt{rel-trial} & \texttt{study-adverse} & $8\mathrm{K}$ & $256$ & $128$ & $1$ & $16.38$ & $\mathbf{16.35}$ \\
\midrule
\textbf{Mean} & & & & & & $33.36$ & $\mathbf{32.52}$ \\
\midrule
\multicolumn{8}{l}{\textit{Classification (AUROC \%, higher is better)}} \\
\midrule
\texttt{rel-amazon} & \texttt{item-churn} & $8\mathrm{K}$ & $512$ & $128$ & $1$ & $76.73$ & $\mathbf{77.79}$ \\
\texttt{rel-amazon} & \texttt{user-churn} & $8\mathrm{K}$ & $256$ & $32$ & $1$ & $\mathbf{66.48}$ & $66.48$ \\
\texttt{rel-avito} & \texttt{user-clicks} & $8\mathrm{K}$ & $2048$ & $128$ & $1$ & $\mathbf{55.54}$ & $47.63$ \\
\texttt{rel-avito} & \texttt{user-visits} & $1\mathrm{K}$ & $512$ & $128$ & $0$ & $\mathbf{61.63}$ & $61.33$ \\
\texttt{rel-event} & \texttt{user-ignore} & $4\mathrm{K}$ & $512$ & $8$ & $1$ & $78.43$ & $\mathbf{80.53}$ \\
\texttt{rel-event} & \texttt{user-repeat} & $2\mathrm{K}$ & $512$ & $8$ & $1$ & $\mathbf{76.31}$ & $76.29$ \\
\texttt{rel-f1} & \texttt{driver-dnf} & $8\mathrm{K}$ & $256$ & $128$ & $1$ & $82.85$ & $\mathbf{83.03}$ \\
\texttt{rel-f1} & \texttt{driver-top3} & $8\mathrm{K}$ & $256$ & $128$ & $1$ & $\mathbf{91.02}$ & $90.75$ \\
\texttt{rel-hm} & \texttt{user-churn} & $8\mathrm{K}$ & $512$ & $128$ & $0$ & $63.14$ & $\mathbf{65.50}$ \\
\texttt{rel-stack} & \texttt{user-badge} & $8\mathrm{K}$ & $4096$ & $32$ & $1$ & $\mathbf{80.20}$ & $79.49$ \\
\texttt{rel-stack} & \texttt{user-engagement} & $512$ & $512$ & $8$ & $1$ & $\mathbf{83.76}$ & $79.39$ \\
\texttt{rel-trial} & \texttt{study-outcome} & $8\mathrm{K}$ & $2048$ & $32$ & $0$ & $58.20$ & $\mathbf{60.52}$ \\
\midrule
\textbf{Mean} & & & & & & $\mathbf{72.86}$ & $72.39$ \\
\bottomrule
\end{tabular}

\end{table}

\subsubsection{Context ensembling (per task)}
\label{app:per-task-n-ens}

Per-task counterpart of \Cref{fig:ensemble}, for the default (blue) and
per-task-tuned (orange) context. Increasing the number of ensembled
context seeds $n_{\text{ens}}$ improves classification on most tasks ($9$ of $12$ at the default context and $10$ of $12$ at the tuned one), but degrades \texttt{rel-avito/user-clicks} by $3.0$ points at the default context, while several regression tasks
(\texttt{rel-amazon/user-ltv}, \texttt{rel-stack/post-votes}) are
insensitive, in line with the flatter aggregate regression curve. The
panels also localize the val--test mismatches behind the aggregate
classification crossover in \Cref{fig:ensemble}: on
\texttt{rel-avito/user-clicks} and \texttt{rel-stack/user-engagement} the
tuned context is clearly worse than the default at every ensemble size, and on \texttt{rel-trial/study-outcome} it starts $2.3$ AUROC ahead but ends $0.7$ behind at $n_{\text{ens}}=16$.
(On \texttt{rel-amazon/user-churn},
tuning selects the default configuration, so the two curves coincide.)

\begin{figure}[!htbp]
  \centering
  \includegraphics[width=0.8\linewidth]{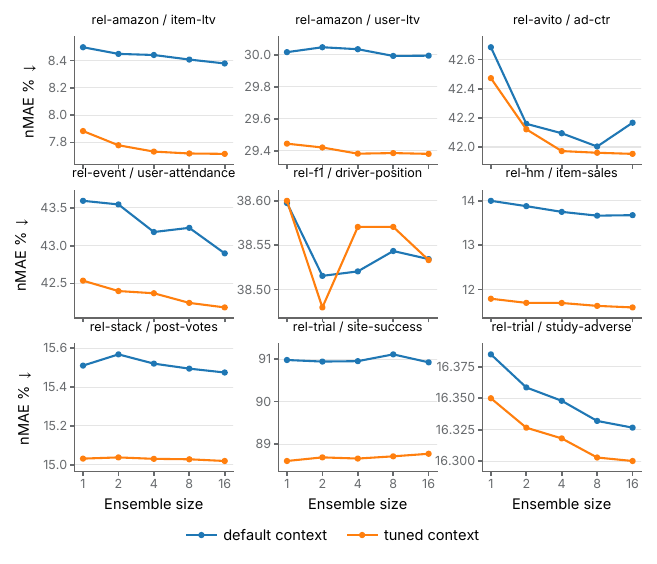}
  \caption{Per-task test nMAE for context ensembling, default vs.\
    per-task-tuned context, on RelBench regression tasks; per-task
    counterpart of \Cref{fig:ensemble} (left).}
  \label{fig:per-task-ensemble-reg}
\end{figure}

\begin{figure}[!htbp]
  \centering
  \includegraphics[width=0.8\linewidth]{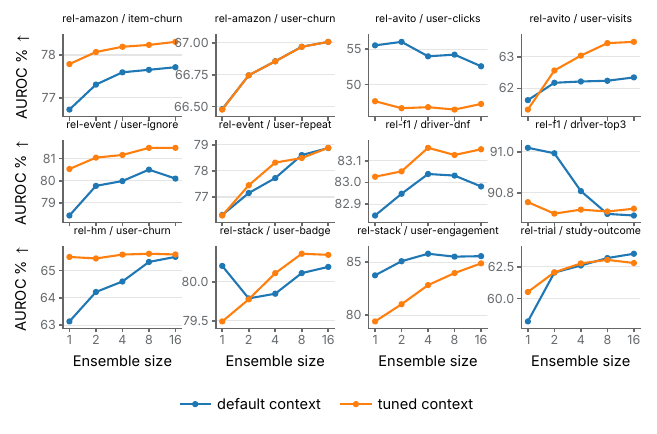}
  \caption{Per-task test AUROC for context ensembling, default vs.\
    per-task-tuned context, on RelBench classification tasks; per-task
    counterpart of \Cref{fig:ensemble} (right).}
  \label{fig:per-task-ensemble-clf}
\end{figure}

\clearpage
\section*{NeurIPS Paper Checklist}

\begin{enumerate}

\item {\bf Claims}
    \item[] Question: Do the main claims made in the abstract and introduction accurately reflect the paper's contributions and scope?
    \item[] Answer: \answerYes{} 
    \item[] Justification: All claims made in the abstract and introduction are backed by experimental results (\Cref{sec:results,sec:test-time-compute,sec:ablations}). 
    \item[] Guidelines:
    \begin{itemize}
        \item The answer \answerNA{} means that the abstract and introduction do not include the claims made in the paper.
        \item The abstract and/or introduction should clearly state the claims made, including the contributions made in the paper and important assumptions and limitations. A \answerNo{} or \answerNA{} answer to this question will not be perceived well by the reviewers. 
        \item The claims made should match theoretical and experimental results, and reflect how much the results can be expected to generalize to other settings. 
        \item It is fine to include aspirational goals as motivation as long as it is clear that these goals are not attained by the paper. 
    \end{itemize}

\item {\bf Limitations}
    \item[] Question: Does the paper discuss the limitations of the work performed by the authors?
    \item[] Answer: \answerYes{} 
    \item[] Justification: We discuss the limitations in the Limitations paragraph of \Cref{sec:discussion}. 
    \item[] Guidelines:
    \begin{itemize}
        \item The answer \answerNA{} means that the paper has no limitation while the answer \answerNo{} means that the paper has limitations, but those are not discussed in the paper. 
        \item The authors are encouraged to create a separate ``Limitations'' section in their paper.
        \item The paper should point out any strong assumptions and how robust the results are to violations of these assumptions (e.g., independence assumptions, noiseless settings, model well-specification, asymptotic approximations only holding locally). The authors should reflect on how these assumptions might be violated in practice and what the implications would be.
        \item The authors should reflect on the scope of the claims made, e.g., if the approach was only tested on a few datasets or with a few runs. In general, empirical results often depend on implicit assumptions, which should be articulated.
        \item The authors should reflect on the factors that influence the performance of the approach. For example, a facial recognition algorithm may perform poorly when image resolution is low or images are taken in low lighting. Or a speech-to-text system might not be used reliably to provide closed captions for online lectures because it fails to handle technical jargon.
        \item The authors should discuss the computational efficiency of the proposed algorithms and how they scale with dataset size.
        \item If applicable, the authors should discuss possible limitations of their approach to address problems of privacy and fairness.
        \item While the authors might fear that complete honesty about limitations might be used by reviewers as grounds for rejection, a worse outcome might be that reviewers discover limitations that aren't acknowledged in the paper. The authors should use their best judgment and recognize that individual actions in favor of transparency play an important role in developing norms that preserve the integrity of the community. Reviewers will be specifically instructed to not penalize honesty concerning limitations.
    \end{itemize}

\item {\bf Theory assumptions and proofs}
    \item[] Question: For each theoretical result, does the paper provide the full set of assumptions and a complete (and correct) proof?
    \item[] Answer: \answerNA{} 
    \item[] Justification: We do not make any theoretical claims. 
    \item[] Guidelines:
    \begin{itemize}
        \item The answer \answerNA{} means that the paper does not include theoretical results. 
        \item All the theorems, formulas, and proofs in the paper should be numbered and cross-referenced.
        \item All assumptions should be clearly stated or referenced in the statement of any theorems.
        \item The proofs can either appear in the main paper or the supplemental material, but if they appear in the supplemental material, the authors are encouraged to provide a short proof sketch to provide intuition. 
        \item Inversely, any informal proof provided in the core of the paper should be complemented by formal proofs provided in appendix or supplemental material.
        \item Theorems and Lemmas that the proof relies upon should be properly referenced. 
    \end{itemize}

    \item {\bf Experimental result reproducibility}
    \item[] Question: Does the paper fully disclose all the information needed to reproduce the main experimental results of the paper to the extent that it affects the main claims and/or conclusions of the paper (regardless of whether the code and data are provided or not)?
    \item[] Answer: \answerYes{}{} 
    \item[] Justification: All the experimental details for the models, data, and hyperparameters are provided (\Cref{sec:corpus,sec:training-methodology,sec:results} and \Cref{app:add_details_dataset,app:pretraining-details,app:leaderboards}).  
    \item[] Guidelines:
    \begin{itemize}
        \item The answer \answerNA{} means that the paper does not include experiments.
        \item If the paper includes experiments, a \answerNo{} answer to this question will not be perceived well by the reviewers: Making the paper reproducible is important, regardless of whether the code and data are provided or not.
        \item If the contribution is a dataset and\slash or model, the authors should describe the steps taken to make their results reproducible or verifiable. 
        \item Depending on the contribution, reproducibility can be accomplished in various ways. For example, if the contribution is a novel architecture, describing the architecture fully might suffice, or if the contribution is a specific model and empirical evaluation, it may be necessary to either make it possible for others to replicate the model with the same dataset, or provide access to the model. In general. releasing code and data is often one good way to accomplish this, but reproducibility can also be provided via detailed instructions for how to replicate the results, access to a hosted model (e.g., in the case of a large language model), releasing of a model checkpoint, or other means that are appropriate to the research performed.
        \item While NeurIPS does not require releasing code, the conference does require all submissions to provide some reasonable avenue for reproducibility, which may depend on the nature of the contribution. For example
        \begin{enumerate}
            \item If the contribution is primarily a new algorithm, the paper should make it clear how to reproduce that algorithm.
            \item If the contribution is primarily a new model architecture, the paper should describe the architecture clearly and fully.
            \item If the contribution is a new model (e.g., a large language model), then there should either be a way to access this model for reproducing the results or a way to reproduce the model (e.g., with an open-source dataset or instructions for how to construct the dataset).
            \item We recognize that reproducibility may be tricky in some cases, in which case authors are welcome to describe the particular way they provide for reproducibility. In the case of closed-source models, it may be that access to the model is limited in some way (e.g., to registered users), but it should be possible for other researchers to have some path to reproducing or verifying the results.
        \end{enumerate}
    \end{itemize}

\item {\bf Open access to data and code}
    \item[] Question: Does the paper provide open access to the data and code, with sufficient instructions to faithfully reproduce the main experimental results, as described in supplemental material?
    \item[] Answer: \answerYes{} 
    \item[] Justification:  All the code and datasets are open-sourced at \url{https://star-project.stanford.edu/rt-j} (\Cref{sec:discussion}). 
    \item[] Guidelines:
    \begin{itemize}
        \item The answer \answerNA{} means that paper does not include experiments requiring code.
        \item Please see the NeurIPS code and data submission guidelines (\url{https://neurips.cc/public/guides/CodeSubmissionPolicy}) for more details.
        \item While we encourage the release of code and data, we understand that this might not be possible, so \answerNo{} is an acceptable answer. Papers cannot be rejected simply for not including code, unless this is central to the contribution (e.g., for a new open-source benchmark).
        \item The instructions should contain the exact command and environment needed to run to reproduce the results. See the NeurIPS code and data submission guidelines (\url{https://neurips.cc/public/guides/CodeSubmissionPolicy}) for more details.
        \item The authors should provide instructions on data access and preparation, including how to access the raw data, preprocessed data, intermediate data, and generated data, etc.
        \item The authors should provide scripts to reproduce all experimental results for the new proposed method and baselines. If only a subset of experiments are reproducible, they should state which ones are omitted from the script and why.
        \item At submission time, to preserve anonymity, the authors should release anonymized versions (if applicable).
        \item Providing as much information as possible in supplemental material (appended to the paper) is recommended, but including URLs to data and code is permitted.
    \end{itemize}

\item {\bf Experimental setting/details}
    \item[] Question: Does the paper specify all the training and test details (e.g., data splits, hyperparameters, how they were chosen, type of optimizer) necessary to understand the results?
    \item[] Answer: \answerYes{} 
    \item[] Justification:  All the necessary experimental details are provided (\Cref{sec:results} and \Cref{app:pretraining-details,app:leaderboards}). 
    \item[] Guidelines:
    \begin{itemize}
        \item The answer \answerNA{} means that the paper does not include experiments.
        \item The experimental setting should be presented in the core of the paper to a level of detail that is necessary to appreciate the results and make sense of them.
        \item The full details can be provided either with the code, in appendix, or as supplemental material.
    \end{itemize}

\item {\bf Experiment statistical significance}
    \item[] Question: Does the paper report error bars suitably and correctly defined or other appropriate information about the statistical significance of the experiments?
    \item[] Answer: \answerNo{} 
    \item[] Justification: Error bars are not reported due to the compute cost of repeated pretraining runs. We report means over the $21$ RelBench tasks with per-task results in \Cref{app:additional-results}. 
    \item[] Guidelines:
    \begin{itemize}
        \item The answer \answerNA{} means that the paper does not include experiments.
        \item The authors should answer \answerYes{} if the results are accompanied by error bars, confidence intervals, or statistical significance tests, at least for the experiments that support the main claims of the paper.
        \item The factors of variability that the error bars are capturing should be clearly stated (for example, train/test split, initialization, random drawing of some parameter, or overall run with given experimental conditions).
        \item The method for calculating the error bars should be explained (closed form formula, call to a library function, bootstrap, etc.)
        \item The assumptions made should be given (e.g., Normally distributed errors).
        \item It should be clear whether the error bar is the standard deviation or the standard error of the mean.
        \item It is OK to report 1-sigma error bars, but one should state it. The authors should preferably report a 2-sigma error bar than state that they have a 96\% CI, if the hypothesis of Normality of errors is not verified.
        \item For asymmetric distributions, the authors should be careful not to show in tables or figures symmetric error bars that would yield results that are out of range (e.g., negative error rates).
        \item If error bars are reported in tables or plots, the authors should explain in the text how they were calculated and reference the corresponding figures or tables in the text.
    \end{itemize}

\item {\bf Experiments compute resources}
    \item[] Question: For each experiment, does the paper provide sufficient information on the computer resources (type of compute workers, memory, time of execution) needed to reproduce the experiments?
    \item[] Answer: \answerYes{} 
    \item[] Justification: All the hardware details are provided along with the training setup (\Cref{app:pretraining-details}). 
    \item[] Guidelines:
    \begin{itemize}
        \item The answer \answerNA{} means that the paper does not include experiments.
        \item The paper should indicate the type of compute workers CPU or GPU, internal cluster, or cloud provider, including relevant memory and storage.
        \item The paper should provide the amount of compute required for each of the individual experimental runs as well as estimate the total compute. 
        \item The paper should disclose whether the full research project required more compute than the experiments reported in the paper (e.g., preliminary or failed experiments that didn't make it into the paper). 
    \end{itemize}
    
\item {\bf Code of ethics}
    \item[] Question: Does the research conducted in the paper conform, in every respect, with the NeurIPS Code of Ethics \url{https://neurips.cc/public/EthicsGuidelines}?
    \item[] Answer: \answerYes{}{} 
    \item[] Justification: The research conforms with the NeurIPS Code of Ethics.  
    \item[] Guidelines:
    \begin{itemize}
        \item The answer \answerNA{} means that the authors have not reviewed the NeurIPS Code of Ethics.
        \item If the authors answer \answerNo, they should explain the special circumstances that require a deviation from the Code of Ethics.
        \item The authors should make sure to preserve anonymity (e.g., if there is a special consideration due to laws or regulations in their jurisdiction).
    \end{itemize}

\item {\bf Broader impacts}
    \item[] Question: Does the paper discuss both potential positive societal impacts and negative societal impacts of the work performed?
    \item[] Answer: \answerYes{} 
    \item[] Justification: We discuss the societal impacts of our work in \Cref{sec:discussion}. 
    \item[] Guidelines:
    \begin{itemize}
        \item The answer \answerNA{} means that there is no societal impact of the work performed.
        \item If the authors answer \answerNA{} or \answerNo, they should explain why their work has no societal impact or why the paper does not address societal impact.
        \item Examples of negative societal impacts include potential malicious or unintended uses (e.g., disinformation, generating fake profiles, surveillance), fairness considerations (e.g., deployment of technologies that could make decisions that unfairly impact specific groups), privacy considerations, and security considerations.
        \item The conference expects that many papers will be foundational research and not tied to particular applications, let alone deployments. However, if there is a direct path to any negative applications, the authors should point it out. For example, it is legitimate to point out that an improvement in the quality of generative models could be used to generate Deepfakes for disinformation. On the other hand, it is not needed to point out that a generic algorithm for optimizing neural networks could enable people to train models that generate Deepfakes faster.
        \item The authors should consider possible harms that could arise when the technology is being used as intended and functioning correctly, harms that could arise when the technology is being used as intended but gives incorrect results, and harms following from (intentional or unintentional) misuse of the technology.
        \item If there are negative societal impacts, the authors could also discuss possible mitigation strategies (e.g., gated release of models, providing defenses in addition to attacks, mechanisms for monitoring misuse, mechanisms to monitor how a system learns from feedback over time, improving the efficiency and accessibility of ML).
    \end{itemize}
    
\item {\bf Safeguards}
    \item[] Question: Does the paper describe safeguards that have been put in place for responsible release of data or models that have a high risk for misuse (e.g., pre-trained language models, image generators, or scraped datasets)?
    \item[] Answer: \answerYes{} 
    \item[] Justification: We discuss our release measures of the collected data in \Cref{sec:appendix-licenses}. 
    \item[] Guidelines:
    \begin{itemize}
        \item The answer \answerNA{} means that the paper poses no such risks.
        \item Released models that have a high risk for misuse or dual-use should be released with necessary safeguards to allow for controlled use of the model, for example by requiring that users adhere to usage guidelines or restrictions to access the model or implementing safety filters. 
        \item Datasets that have been scraped from the Internet could pose safety risks. The authors should describe how they avoided releasing unsafe images.
        \item We recognize that providing effective safeguards is challenging, and many papers do not require this, but we encourage authors to take this into account and make a best faith effort.
    \end{itemize}

\item {\bf Licenses for existing assets}
    \item[] Question: Are the creators or original owners of assets (e.g., code, data, models), used in the paper, properly credited and are the license and terms of use explicitly mentioned and properly respected?
    \item[] Answer: \answerYes{}{} 
    \item[] Justification: All the papers and datasets have been cited, with the license of each source in \Cref{tab:domains} and our compliance practice in \Cref{sec:appendix-licenses}. 
    \item[] Guidelines:
    \begin{itemize}
        \item The answer \answerNA{} means that the paper does not use existing assets.
        \item The authors should cite the original paper that produced the code package or dataset.
        \item The authors should state which version of the asset is used and, if possible, include a URL.
        \item The name of the license (e.g., CC-BY 4.0) should be included for each asset.
        \item For scraped data from a particular source (e.g., website), the copyright and terms of service of that source should be provided.
        \item If assets are released, the license, copyright information, and terms of use in the package should be provided. For popular datasets, \url{paperswithcode.com/datasets} has curated licenses for some datasets. Their licensing guide can help determine the license of a dataset.
        \item For existing datasets that are re-packaged, both the original license and the license of the derived asset (if it has changed) should be provided.
        \item If this information is not available online, the authors are encouraged to reach out to the asset's creators.
    \end{itemize}

\item {\bf New assets}
    \item[] Question: Are new assets introduced in the paper well documented and is the documentation provided alongside the assets?
    \item[] Answer: \answerYes{}{} 
    \item[] Justification: All new assets are documented (\Cref{sec:corpus} and \Cref{app:add_details_dataset}) and are open-sourced. 
    \item[] Guidelines:
    \begin{itemize}
        \item The answer \answerNA{} means that the paper does not release new assets.
        \item Researchers should communicate the details of the dataset\slash code\slash model as part of their submissions via structured templates. This includes details about training, license, limitations, etc. 
        \item The paper should discuss whether and how consent was obtained from people whose asset is used.
        \item At submission time, remember to anonymize your assets (if applicable). You can either create an anonymized URL or include an anonymized zip file.
    \end{itemize}

\item {\bf Crowdsourcing and research with human subjects}
    \item[] Question: For crowdsourcing experiments and research with human subjects, does the paper include the full text of instructions given to participants and screenshots, if applicable, as well as details about compensation (if any)? 
    \item[] Answer: \answerNA{} 
    \item[] Justification: This research does not involve crowdsourcing experiments. 
    \item[] Guidelines:
    \begin{itemize}
        \item The answer \answerNA{} means that the paper does not involve crowdsourcing nor research with human subjects.
        \item Including this information in the supplemental material is fine, but if the main contribution of the paper involves human subjects, then as much detail as possible should be included in the main paper. 
        \item According to the NeurIPS Code of Ethics, workers involved in data collection, curation, or other labor should be paid at least the minimum wage in the country of the data collector. 
    \end{itemize}

\item {\bf Institutional review board (IRB) approvals or equivalent for research with human subjects}
    \item[] Question: Does the paper describe potential risks incurred by study participants, whether such risks were disclosed to the subjects, and whether Institutional Review Board (IRB) approvals (or an equivalent approval/review based on the requirements of your country or institution) were obtained?
    \item[] Answer: \answerNA{} 
    \item[] Justification: This research does not involve human subjects. 
    \item[] Guidelines:
    \begin{itemize}
        \item The answer \answerNA{} means that the paper does not involve crowdsourcing nor research with human subjects.
        \item Depending on the country in which research is conducted, IRB approval (or equivalent) may be required for any human subjects research. If you obtained IRB approval, you should clearly state this in the paper. 
        \item We recognize that the procedures for this may vary significantly between institutions and locations, and we expect authors to adhere to the NeurIPS Code of Ethics and the guidelines for their institution. 
        \item For initial submissions, do not include any information that would break anonymity (if applicable), such as the institution conducting the review.
    \end{itemize}

\item {\bf Declaration of LLM usage}
    \item[] Question: Does the paper describe the usage of LLMs if it is an important, original, or non-standard component of the core methods in this research? Note that if the LLM is used only for writing, editing, or formatting purposes and does \emph{not} impact the core methodology, scientific rigor, or originality of the research, declaration is not required.
    \item[] Answer: \answerYes{}{} 
    \item[] Justification: We document the usage of LLMs for curating our data corpus in \Cref{sec:tasks} and \Cref{sec:appendix-agentic,sec:appendix-propose}. 
    \item[] Guidelines:
    \begin{itemize}
        \item The answer \answerNA{} means that the core method development in this research does not involve LLMs as any important, original, or non-standard components.
        \item Please refer to our LLM policy in the NeurIPS handbook for what should or should not be described.
    \end{itemize}

\end{enumerate}

\end{document}